\documentclass{article}
 \PassOptionsToPackage{numbers, compress}{natbib}

\usepackage[main, final]{neurips_2026}

\usepackage{amsmath}
\usepackage{amssymb}
\usepackage{colortbl}
\usepackage{enumitem}
\usepackage{graphicx}
\usepackage{mathtools}
\usepackage{multirow}
\usepackage{placeins}
\usepackage{needspace}
\usepackage{subcaption}
\usepackage{xcolor}
\usepackage{natbib}
\definecolor{best}{RGB}{180, 220, 200}
\definecolor{second}{RGB}{225, 250, 240}
\makeatletter
\newcommand{\oursrowshade}{\rlap{\color{gray!10}\vrule width\linewidth height\ht\@arstrutbox depth\dp\@arstrutbox}}
\makeatother
\usepackage{wrapfig}
\usepackage{bm}
\usepackage{xspace}

\usepackage[utf8]{inputenc} 
\usepackage[T1]{fontenc}    
\usepackage{hyperref}       
\hypersetup{hidelinks}
\usepackage{url}            
\usepackage{booktabs}       
\usepackage{amsfonts}       
\usepackage{nicefrac}       
\usepackage{microtype}      

\usepackage[accsupp]{axessibility}
\usepackage{orcidlink}
\usepackage{bm}

\usepackage[capitalize,noabbrev,nameinlink]{cleveref}
\crefname{section}{Section}{Sections}
\Crefname{section}{Section}{Sections}
\crefname{subsection}{Section}{Sections}
\Crefname{subsection}{Section}{Sections}
\crefname{appendix}{Appendix}{Appendices}
\Crefname{appendix}{Appendix}{Appendices}

\newcommand{\astra}[2][note]{%
  \begingroup
  \def\astramode{#1}\def\astraeditmode{edit}%
  \ifx\astramode\astraeditmode
    \ifdefined\HoloHideAstra#2\else\ifdefined\HoloReview{\color{red}#2}\else#2\fi\fi
  \else
    \ifdefined\HoloHideAstra\else\ifdefined\HoloReview{\color{red}\small\textsf{[Astra: #2]}}\fi\fi
  \fi
  \endgroup
}

\usepackage{xspace}
\newcommand{\nickname}{\textsc{Holo4D}\xspace}

\PassOptionsToPackage{pagebackref}{hyperref}

\definecolor{tealborder}{RGB}{5,170,159}

\makeatletter
\renewcommand{\paragraph}{%
  \@startsection{paragraph}{4}
  {\z@}{0em}{-0.5em}%
  {\normalfont\normalsize\bfseries}%
}
\makeatother

\title{\nickname: Holistic 4D Reconstruction as Geometric Control for Video Diffusion}

\author{%
  \textbf{Yushi Lan}$^{1*}$ \quad
  \textbf{Zeren Jiang}$^{1*}$ \quad
  \textbf{Koichi Namekata}$^{1}$ \quad \\[0.5em]
  \textbf{Xingang Pan}$^{2}$ \quad
  \textbf{Chuanxia Zheng}$^{2}$ \quad
  \textbf{Andrea Vedaldi}$^{1}$ \\[1em]
  $^{1}$University of Oxford \quad
  $^{2}$Nanyang Technological University \\[0.3em]
  \href{https://nirvanalan.github.io/projects/holo4d}{\textcolor{magenta}{\texttt{https://nirvanalan.github.io/projects/holo4d}}}
}

\begin{document}
\maketitle

\begingroup
  \renewcommand{\thefootnote}{\fnsymbol{footnote}}
  \footnotetext[1]{Equal contribution.}
\endgroup

\setlength{\textfloatsep}{10pt plus 2pt minus 2pt}
\suppressfloats[t]
\begin{figure}[h!]
\vspace{-18pt}
\centering
\includegraphics[width=\textwidth]{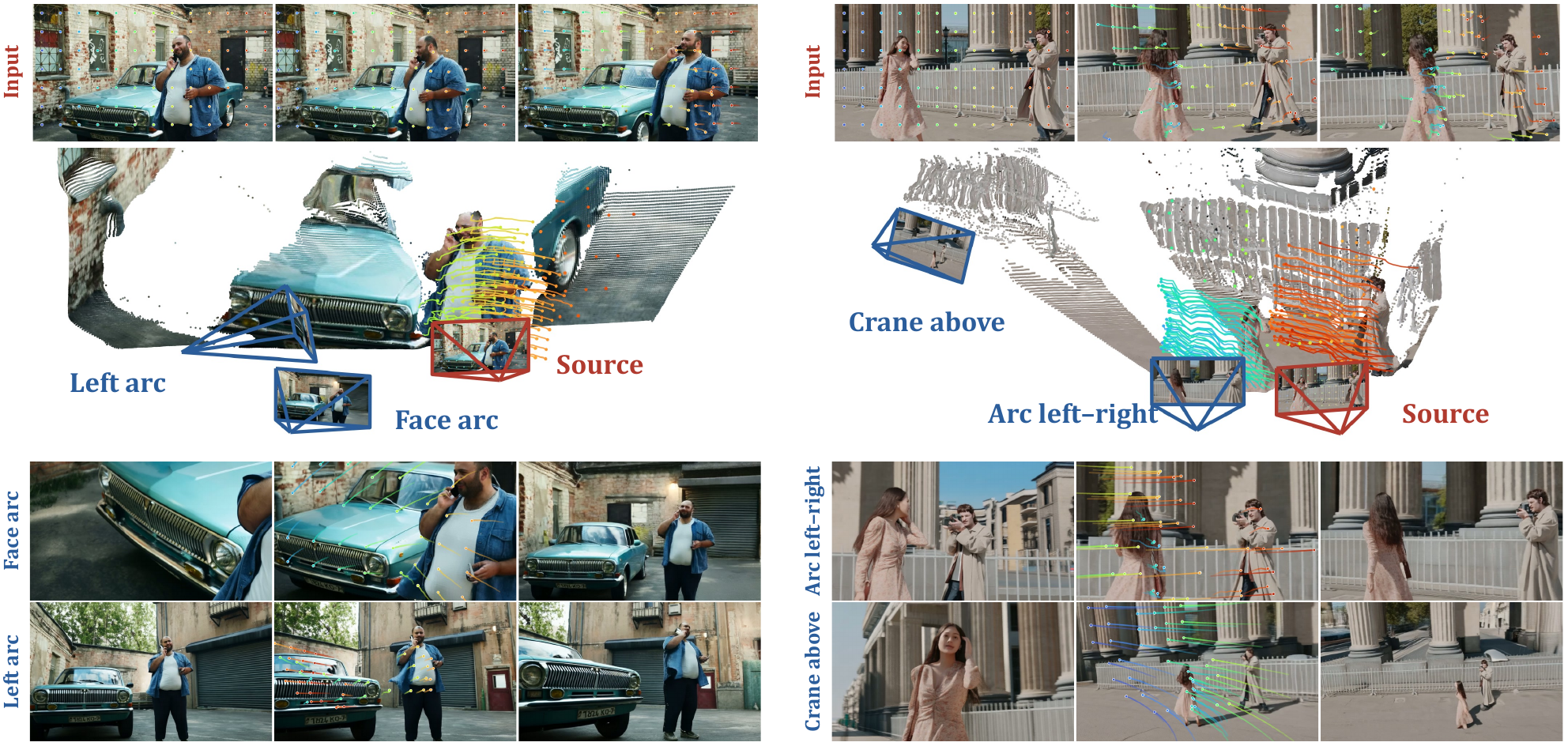}
\vspace{-4pt}
\caption{\textbf{4D reconstruction as a scaffold for video reshooting.}
Top: input video with its 3D trajectories (projected onto the image plane for visualization).
Middle: reconstruction with the source camera (red) and target cameras (blue) for camera paths from Vista4D's benchmark. Bottom: \nickname outputs for each path; the middle frame overlays the projected trajectories (subset shown).}
\label{fig:teaser}
\end{figure}
\begin{abstract}
Synthesizing a monocular video from a new camera trajectory requires preserving the observed scene geometry and motion while completing regions that become newly visible.
Feed-forward 4D reconstructors recover the structure of the observed scene, whereas video diffusion models provide a generative prior for completing unseen content.
We introduce \nickname, a video reshooting framework that uses reconstructed 3D trajectories as a holistic source-view 4D scaffold.
We project these trajectories into the requested views and encode their geometry and source appearance to guide a pretrained video diffusion model.
Together with the source video and target cameras, the scaffold grounds generation in reconstructed observations while leaving unsupported regions to the generative prior.
Experiments demonstrate the value of trajectory-based conditioning for camera control and novel-view synthesis.
The framework accommodates different 4D reconstructors without retraining the video generator, allowing it to benefit from future advances in source reconstruction.
More details can be found in our \href{https://nirvanalan.github.io/projects/holo4d}{\textcolor{magenta}{project page}}.
\end{abstract}

\section{Introduction}%
\label{sec:intro}
\begingroup
\parfillskip=0pt plus .48\linewidth

Rendering a monocular video from a new camera trajectory would allow us to change the viewpoint of a recorded scene without changing the events it depicts~\cite{bai25recammaster:}.
For example, a camera moving around a walking person should reveal a different view of the same motion, rather than generate a different action or alter the surrounding scene.
This requires both faithful reconstruction of observed content and plausible completion of surfaces that become visible only from the new viewpoint.
The task therefore goes beyond recovering the geometry of the input frames: it must turn partial observations of a dynamic scene into a coherent video along a user-specified camera path.

Recent feed-forward reconstruction models provide an increasingly useful starting point.
They estimate cameras, depth, and point maps directly from images and videos~\cite{wang25vggt:,lan26stream3r:,lin26depth:,wang26vggt-omega:}, while 4D reconstruction systems also recover the trajectories of scene points over time~\cite{luo264rc:,zhang26efficiently:,sucar26v-dpm:,zhu26motioncrafter:}.
Together, these quantities describe the observed scene in a common spatial and temporal frame: camera motion can be separated from scene motion, and observations of a moving point can be linked across frames.
However, reconstructing the observed scene does not by itself solve novel-view synthesis.
Reprojecting its points into a new camera leaves holes where surfaces have never been observed~\cite{yoon20novel:}, and inaccurate geometry can produce distortions as the viewpoint changes~\cite{lin26vista4d:,ren25gen3c:}.
The missing appearance must be inferred, rather than obtained by reprojection alone.

Video diffusion models offer the complementary ability to generate plausible appearance and motion from learned video priors~\cite{brooks24video:}.
Camera-controlled variants use these priors to synthesize new views, conditioned on camera embeddings or rendered source geometry~\cite{bai25recammaster:,yu25trajectorycrafter:,ren25gen3c:}.
Yet a realistic video need not preserve the geometry and dynamics of the particular scene being reconstructed.
Combining reconstruction with generation is therefore not simply a matter of requesting a new camera pose: the generator also needs a representation of the scene that the camera is viewing.
This motivates our central question: \emph{how should the geometry, appearance, and motion recovered from a source video be exposed to a video diffusion model?}

An important part of this question is how the condition represents a scene over time.
Lifting each frame independently with depth yields a sequence of point clouds, but does not explicitly specify which points correspond to the same moving surface in different frames.
Vista4D~\cite{lin26vista4d:} makes static scene content persistent by identifying static pixels and accumulating their observations in a common world frame.
However, moving surfaces cannot be accumulated in the same way without accounting for their motion.
Our starting point is that persistent 3D trajectories provide the required correspondence explicitly~\cite{luo264rc:,zhang26efficiently:,sucar26v-dpm:}.
Specifically, a trajectory follows one scene point through time and associates its position with its observed appearance.
A static point is the special case of a constant world-space trajectory, so static and moving points can use the same conditioning encoder.

We introduce \nickname, a video reshooting framework that converts reconstructed 3D trajectories and their source appearance into geometric features for video diffusion (\cref{fig:method_pipeline}).
We call this scaffold holistic because it represents observed geometry, appearance, and motion jointly for static and dynamic points.
Projecting these trajectories into the target cameras specifies where observed scene content should appear and how it should move.
We use these projected trajectories to guide the video diffusion model, alongside source-video conditioning~\cite{chen254dnex:} and the requested camera trajectory~\cite{sitzmann2021lfns}.
The tracks guide observed content throughout denoising, while the video prior supplies plausible appearance in newly revealed regions.

This conditioning design has two useful properties.
First, we inject track features along the channel dimension~\cite{namekata26gowiththetrack:}, providing geometric guidance without an additional rendered-video token stream~\cite{lin26vista4d:}.
Second, the scaffold uses explicit geometric quantities rather than reconstructor-specific features.
\nickname can therefore use any 4D reconstructor that provides the required geometry and trajectories~\cite{luo264rc:,sucar26v-dpm:,wang26vggt-omega:,lai26cowtracker:}.
Our experiments show that replacing the upstream reconstruction can substantially improve camera adherence without retraining the video generator.

In summary, our contributions are threefold:
(1) a holistic geometric scaffold that represents static and dynamic scene observations through persistent 3D trajectories with source appearance;
(2) a conditioning design that combines source-video context with channel-wise injection of projected 3D trajectory features to guide novel-view generation; and
(3) camera-adherence and held-out-view evaluations against strong baselines, analyses of reconstructor substitutions and trajectory perturbations, and demonstrations on robot videos.
\par
\endgroup

\section{Related Work}%
\label{sec:related}
\begingroup
\parfillskip=0pt plus .48\linewidth

\paragraph{3D and 4D reconstruction.}
Dynamic scenes can be reconstructed by fitting deformable radiance fields~\cite{park21nerfies:,park21hypernerf:,pumarola21d-nerf:,li23dynibar:} or time-varying Gaussian representations~\cite{yang24deformable:,wu244d-gaussian:,wang25shape:}.
ViDAR~\cite{nazarczuk25vidar:} additionally uses a personalized diffusion prior to supervise per-scene Gaussian reconstruction.
We instead condition a shared video generator on reconstructed geometry, without per-scene optimization.

Learned models recover point maps, cameras, and depth from images~\cite{wang24dust3r:,wang25vggt:,lin26depth:,wang26pi:,keetha26mapanything:,wang26vggt-omega:} or accumulate observations over video~\cite{wang25continuous:,lan26stream3r:}.
Video depth and dynamic-scene estimators provide geometry from moving-camera videos~\cite{chen25video:,hu25depthcrafter:,xu25geometrycrafter:,li25megasam:,zhang24monst3r:,yao25uni4d:,jiang25geo4d:}.
Recent 4D methods estimate dynamic point maps, motion, or persistent trajectories~\cite{feng25st4rtrack:,sucar25dynamic:,sucar26v-dpm:,liu26trace:,karhade26any4d:,luo264rc:,zhang26efficiently:,zhu26motioncrafter:}, while 4DGT~\cite{xu254dgt:} predicts 4D Gaussians from posed videos.
We use compatible geometric outputs as a scaffold for video reshooting.

\paragraph{Geometry-guided view synthesis.}
Geometry-guided rendering aggregates source-view information to synthesize novel views~\cite{wang21ibrnet:,chen24mvsplat360:,szymanowicz25flash3d,ha25geometryguided:}.
Ha et al.~\cite{ha25geometryguided:} also use temporal guidance and inpainting for multiview video.
Our setting uses reconstructed 3D trajectories to condition video diffusion for reshooting a monocular video under large viewpoint changes.

\paragraph{Camera-controlled video generation.}
Camera-conditioned diffusion models encode target views through poses or rays~\cite{liu23zero-1-to-3:,zheng24free3d,voleti2024sv3d,wang24motionctrl:,he25cameractrl:,he25cameractrl2:,bahmani24vd3d:}, and have been adapted to video reshooting~\cite{bai25recammaster:,van-hoorick24generative:}.
Explicit conditions instead ground generation in rendered source geometry~\cite{yu25trajectorycrafter:,ren25gen3c:,hu25ex-4d:} or warped observations~\cite{you25nvs-solver:}.
Related methods generate multiview videos before Gaussian reconstruction~\cite{wu25cat4d:}, or render Gaussian features for sparse-view synthesis~\cite{chen24mvsplat360:}.

Vista4D~\cite{lin26vista4d:} conditions on source video and rendered point clouds with temporally persistent static pixels.
LaVR~\cite{xie26lavr:} uses latent reconstruction states, whereas our interface is defined by explicit geometric quantities rather than reconstructor-specific hidden features.
Go-with-the-Track~\cite{namekata26gowiththetrack:} conditions generation on reference images and reference-anchored 2D tracks for compositing and motion control, and also supports camera retargeting.
Edit-by-Track~\cite{lee2026generative} edits camera and object motion in a video by conditioning on paired source and target 3D point tracks.
For video reshooting, \nickname encodes 3D trajectory positions, source color, validity, and target-camera depth into geometric features, alongside the synchronized source video.

\begin{figure*}[t]
\centering
\includegraphics[width=\textwidth]{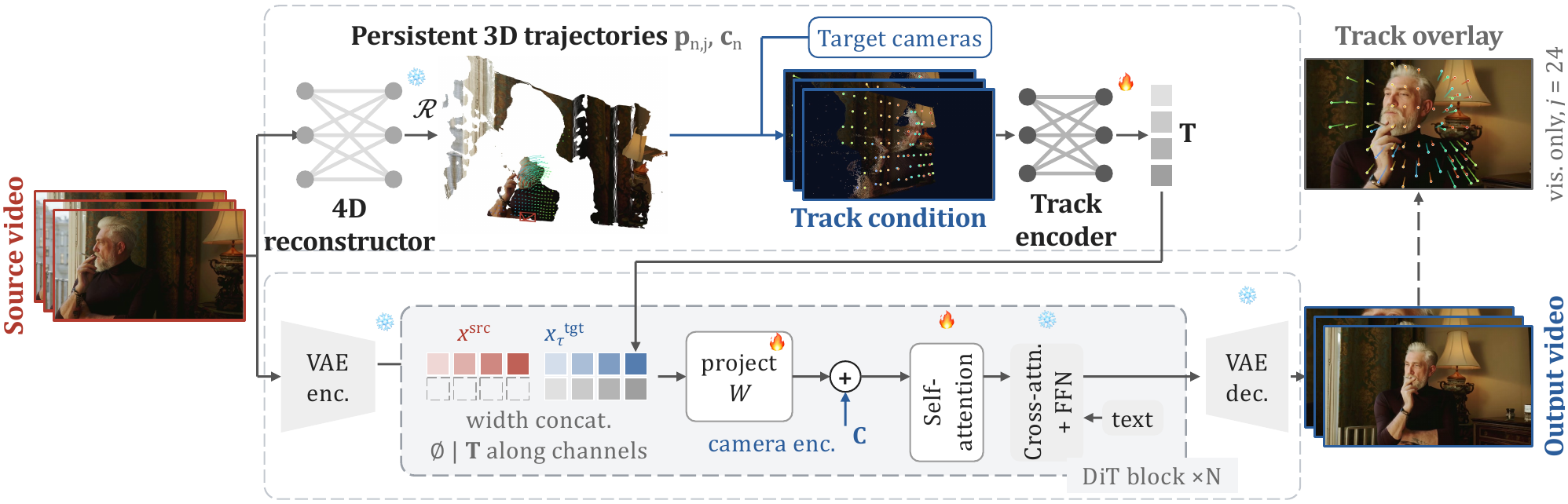}
\caption{\textbf{\nickname overview.}
Top (4D scaffold): a 4D reconstructor $\mathcal R$ recovers persistent 3D trajectories, paired with sampled source colors, which are projected into the target cameras and encoded into track features $\mathbf T$.
Bottom (video diffusion): source and noisy target latents are concatenated along width; each DiT block concatenates $\mathbf T$ with its hidden features along channels (null on the source half) and adds camera features $\mathbf C$ encoded from Pl\"ucker rays.
Flames and snowflakes mark trained and frozen modules (\cref{app:impl}).
Trajectory panels show a subset of the dense tracks; the top-right overlay is for visualization only. The output is obtained by iterative denoising; \cref{fig:track_condition_detail} details the DiT block.}
\label{fig:method_pipeline}
\end{figure*}
\paragraph{Diffusion models for geometry.}
Diffusion priors also support geometry prediction, from monocular depth~\cite{ke24repurposing:,xu25pixel-perfect:} to video geometry and motion~\cite{jiang25geo4d:,chen254dnex:,zhu26motioncrafter:}.
Gen3R~\cite{huang26gen3r:} jointly generates appearance and geometric latents.
VIST3A~\cite{go26text-to-3d:} couples a video generator to a reconstruction decoder, while Plan4D~\cite{zhang2026plan4d} further generates plannable 4D worlds by stitching video generators and reconstruction decoders.
\nickname instead takes reconstructed geometry as input and generates target-view RGB.

\paragraph{Video models for robots.}
Video-generation backbones increasingly serve robot learning, e.g., as world action models~\cite{yuan26fast-wam:}; \nickname instead reshoots recorded robot videos from new viewpoints (\cref{sec:robot}).
\par
\endgroup

\section{Method}%
\label{sec:method}

\label{sec:problem}
Given a source video $V^{\mathrm{src}}=(I_0^{\mathrm{src}},\dots,I_{N-1}^{\mathrm{src}})$ and a target camera trajectory, \nickname generates a video $V^{\mathrm{tgt}}=(I_0^{\mathrm{tgt}},\dots,I_{N-1}^{\mathrm{tgt}})$ that shows the same scene and motion from the requested cameras.
Source and target frames with the same index refer to the same scene time.
In \cref{fig:method_pipeline}, the top branch reconstructs persistent 3D trajectories from the source video, forming a \emph{4D scaffold}, projects them into the target cameras, and encodes them into target-aligned track features (\cref{sec:track_condition}).
The bottom branch, a pretrained video diffusion model, generates the target video from the source video, the target cameras, and these features (\cref{sec:camera_injection}).

\subsection{Constructing a 4D scaffold}
\label{sec:track_condition}

Reshooting a dynamic scene requires specifying where observed content should appear in each target frame and how its position changes over time.
A video diffusion model conditioned only on the source video and target cameras~\cite{bai25recammaster:} must infer this layout implicitly from appearance.
Without an explicit scene representation, it struggles to preserve source-scene geometry and follow the requested camera under large viewpoint changes~\cite{xie26lavr:,lin26vista4d:}.
To provide explicit geometric guidance, existing methods lift source frames into point clouds using estimated depth and render them from the target cameras~\cite{yu25trajectorycrafter:,ren25gen3c:}.
However, even when static points are accumulated across frames~\cite{lin26vista4d:}, independently reconstructed dynamic points do not explicitly establish correspondence between moving surfaces across frames.
We therefore represent the source scene by persistent 3D trajectories~\cite{xiao24spatialtracker:}, each carrying the appearance of a source observation.

To construct this scaffold, we use a 4D reconstructor $\mathcal R$ to recover camera poses and point trajectories from the source video~\cite{luo264rc:,zhang26efficiently:,sucar26v-dpm:,wang26vggt-omega:,lai26cowtracker:}.
Each trajectory follows a scene point $n$ through its positions $\mathbf p_{n,j}$ at successive frames $j$ and carries a color $\mathbf c_n$, sampled from the source frame $q_n$ where the point was queried.
We express these positions and the target cameras in the coordinate frame of the first source camera~\cite{wang24dust3r:,sucar26v-dpm:}, where static points have constant positions and moving points trace out their motion.
Any reconstructor providing these trajectories and cameras can supply the scaffold.

To express the scaffold in the target views, we project each $\mathbf p_{n,j}$ with the target intrinsics $K_j^{\mathrm{tgt}}$ and pose $\pi_j^{\mathrm{tgt}}$ of frame $j$.
This determines the point's target-view pixel location and its target-camera depth $d_{n,j}$, both computed from the reconstructed 3D position.
A validity flag $v_{n,j}$ marks observations that are valid according to the upstream reconstructor and project inside the target image; it does not account for occlusion by other points.
Each observation carries its frame index, position $\mathbf p_{n,j}$, depth $d_{n,j}$, validity $v_{n,j}$, and color $\mathbf c_n$.
The color stays fixed along the trajectory, linking source appearance to the point's reconstructed motion.

Finally, we convert the projected observations into a feature volume aligned with the target video.
A sinusoidal encoding~\cite{tancik2020fourier:} followed by a shared MLP maps the attributes of each observation to a feature vector, which is scattered to its projected location on a target-view grid.
Features of observations that fall into the same cell are combined by a channel-wise maximum, and an occupancy channel records which cells received any observation.
A causal 3D convolutional adapter then brings this volume to the DiT token grid, producing the track features $\mathbf T$; target tokens with no projected observations receive a learned null embedding.
For a fixed target-video resolution and duration, the resulting feature volume has a size independent of the number of trajectories.
Normalization, masking, and dataset-specific depth handling are given in \cref{app:impl}.

\Needspace{6\baselineskip}
\subsection{Conditioning video diffusion}
\label{sec:camera_injection}

The generator is a pretrained video diffusion transformer that combines source-video context, target-camera features, and the track features $\mathbf T$.
The VAE encodes the source video into latents $x^{\mathrm{src}}$, concatenated with the noisy target latents $x_\tau^{\mathrm{tgt}}$ along the width~\cite{chen254dnex:}.
This keeps synchronized source and target frames at the same temporal index, whereas temporal concatenation~\cite{lin26vista4d:,bai25recammaster:} places them at different indices.
Learned domain embeddings distinguish the source and target halves.
The text prompt enters through the backbone's pretrained cross-attention.

The track features enter the transformer along the channel dimension, following Go-with-the-Track~\cite{namekata26gowiththetrack:}, rather than as an additional sequence of tokens.
We place them on the target half of the sequence and extend $\mathbf T$ on the source half with the same learned null embedding used for unoccupied target tokens.
Channel-wise injection leaves the self-attention sequence length unchanged, and our model achieves a $1.34\times$ denoising speedup over Vista4D (\cref{tab:vista4d_runtime}).

The projected trajectories already convey the target viewpoint through their image locations.
To provide auxiliary camera information in regions without valid projected observations, we also encode the target cameras as Pl\"ucker rays~\cite{sitzmann2021lfns} into camera features $\mathbf C$.
Source cameras establish the common reconstruction frame and are not separately embedded in the generator.

In the implementation, we inject the track and camera features before self-attention in every DiT block.
Writing the hidden features of block $\ell$ after normalization and timestep modulation as $\mathbf h^{(\ell)}$, we compute the self-attention input as
\begin{equation}
\widehat{\mathbf h}^{(\ell)}=W^{(\ell)}[\mathbf h^{(\ell)};\mathbf T]+\mathbf C,
\label{eq:track_injection}
\end{equation}
where $[\cdot;\cdot]$ concatenates channels and $W^{(\ell)}$ is a learned projection back to the hidden width.
Following the zero-initialized conditioning principle of ControlNet~\cite{zhang2023adding}, the projection initially preserves the hidden features and assigns zero weight to the track channels.
The complete block layout is shown in \cref{fig:track_condition_detail}, with implementation details in \cref{app:impl}.

\subsection{Training and inference}
\label{sec:training_inference}

A training example consists of a source video with 3D trajectory annotations, target cameras, and the target video seen from those cameras.
The trajectory annotations are required to construct the conditioning features $\mathbf T$ and are obtained from an upstream reconstructor or, where available, dataset ground truth.
We encode the target video into latents $x^{\mathrm{tgt}}$ and corrupt them with Gaussian noise $\epsilon$ as $x_\tau^{\mathrm{tgt}}=(1-\tau)\epsilon+\tau x^{\mathrm{tgt}}$, where $\tau\in[0,1]$ is the diffusion time, distinct from the video frame index; the source latents stay clean.
The generator $u_{\bm\theta}$ is trained to predict the velocity of this interpolation with the rectified-flow objective~\cite{liu23flow:}
\begin{equation}
\mathcal L_{\mathrm{FM}}=\mathbb E\left[\left\|u_{\bm\theta}(x_\tau^{\mathrm{tgt}},\tau; x^{\mathrm{src}},\mathbf T,\mathbf C)-(x^{\mathrm{tgt}}-\epsilon)\right\|_2^2\right],
\label{eq:conditional_fm_simple}
\end{equation}
where the expectation is over training examples, noise, and diffusion time; auxiliary conditions are detailed in \cref{app:impl}.
Only the target latents are supervised; no geometry is predicted.

At test time, we reconstruct the source video, build $\mathbf T$ and $\mathbf C$ for the requested cameras, and integrate the learned velocity from noise to target latents, which the VAE decodes into the output video.
Videos longer than the training window are generated window by window (\cref{sec:robot} and \cref{app:impl}).

\Needspace{6\baselineskip}
\subsection{Training details and datasets}
\label{sec:training_details}

We initialize from Wan2.1-T2V-14B~\cite{wan25wan:} and train on $49$-frame videos for $15$k steps at $144{\times}256$, followed by $4.5$k steps at $384{\times}672$.
AdamW~\cite{loshchilov19decoupled:} uses learning rates $1{\times}10^{-5}$ and $2{\times}10^{-5}$ and effective batch sizes $8$ and $32$ for the two stages, respectively.
We train conditioning modules and DiT self-attention, keeping pretrained cross-attention and feed-forward layers frozen~\cite{fu26plenoptic:}.

\paragraph{Datasets.}
We train on a mixture of synthetic multiview videos of dynamic scenes~\cite{jiang26syn4d:,greff22kubric:,bai25recammaster:}, real videos of static scenes~\cite{ling24dl3dv-10k:,zhou18stereo:}, and synthetic static-scene sequences~\cite{huang18deepmvs:,wang20tartanair:}.
For dynamic scenes, source--target pairs come from synchronized multiview videos.
For static scenes, we form source--target pairs from different chunks of the same monocular video, similar to 4Real-Video-V2~\cite{wang26fused:}; temporal synchronization is unnecessary since the scene is static.
This mixture covers both scene motion and camera motion, with estimated or ground-truth 3D trajectory annotations; dataset names, sampling weights, and annotation sources are given in \cref{app:training_data}.

To expose the generator to imperfect and incomplete conditions, the second training stage varies trajectory density and query-frame sampling and applies mild geometric perturbations.
Module dimensions and augmentation settings are specified in \cref{app:impl}.

\section{Experiments}%
\label{sec:experiments}

We evaluate camera control and novel-view synthesis (\cref{sec:quantitative,sec:qualitative}), analyze geometric conditioning (\cref{sec:ablation}), and demonstrate robot-video applications (\cref{sec:robot}).

\begin{figure*}[t]
\centering
\includegraphics[width=\textwidth]{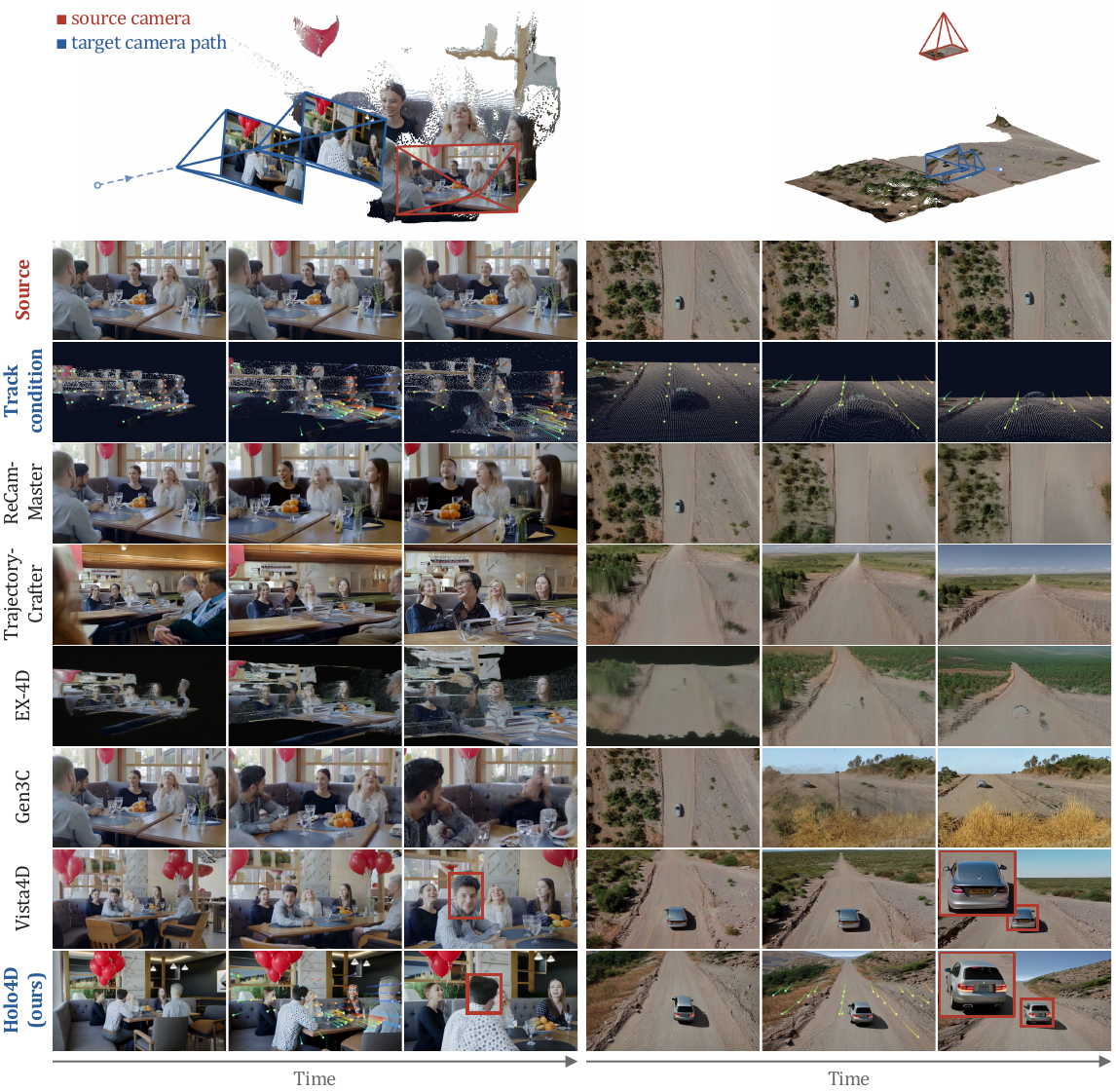}
\caption{\textbf{Video reshooting on Vista4D's benchmark trajectories.} Top: dense source geometry, source camera (red), and requested camera path (blue; circle: $j=0$, frusta: $j=24,48$). The track condition projects source trajectories into the target view; dark regions have no support. The \nickname frame at $j=24$ overlays a subset of the tracks. Red boxes (with magnified insets) mark Vista4D's errors: the diner is turned toward the camera, and the car loses its rear window. \nickname uses $50$ steps with the two-stage backend (VGGT-$\Omega$ + CoWTracker; left) or 4RC (right); baseline inputs and released settings are described in \cref{app:vista4d_native}.}
\label{fig:qualitative}
\end{figure*}

\begin{figure*}[t]
\centering
\includegraphics[width=\textwidth]{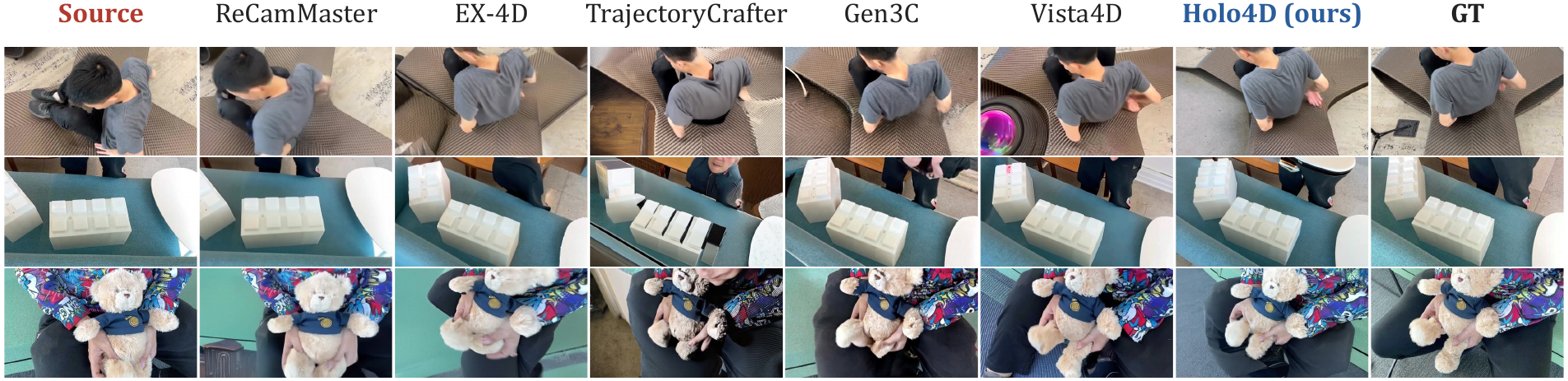}
\caption{\textbf{Novel-view synthesis on iPhone.} Rows show spin, block, and teddy at $j=24$, with ground-truth target views (GT) on the right. \nickname uses the two-stage backend and $50$ steps.}
\label{fig:iphone_nvs}
\end{figure*}

\subsection{Experimental setup}

\paragraph{Baselines.}
We compare \nickname with several representative state-of-the-art methods, including ReCamMaster~\cite{bai25recammaster:}, TrajectoryCrafter~\cite{yu25trajectorycrafter:}, EX-4D~\cite{hu25ex-4d:}, Gen3C~\cite{ren25gen3c:}, and the concurrent Vista4D~\cite{lin26vista4d:}.
ReCamMaster is a \emph{camera-only} baseline: it conditions on the source video and the target cameras.
The others are \emph{rendered-geometry} baselines: they also render source geometry from target cameras; Vista4D keeps static geometry persistent across frames.

\paragraph{Evaluation datasets.}
For camera control, we select $50$ DAVIS~\cite{pont-tuset17davis:} clips and randomly sample two target camera trajectories per clip, and also evaluate on Vista4D's released $110$-trajectory benchmark.
Five iPhone/DyCheck scenes~\cite{gao22monocular:} provide synchronized held-out views, while held-out Syn4D~\cite{jiang26syn4d:} scenes provide exact target RGB and geometry.
For each metric, all methods are evaluated at the same resolution; detailed protocols are in \cref{app:davis_eval,app:iphone_eval,app:robustness}.

\paragraph{4D scaffold reconstruction.}
Our default source reconstructor $\mathcal R$ is 4RC~\cite{luo264rc:}.
We also test two-stage 4D reconstruction from 3D geometry and 2D tracking, as used for annotation in MolmoMotion~\cite{zhang26molmomotion:}, by combining VGGT-$\Omega$~\cite{wang26vggt-omega:} with CoWTracker~\cite{lai26cowtracker:}.
For this two-stage backend, we optionally use GMOS~\cite{xie26gmos:} to hold background points fixed in 3D while preserving foreground motion.

\paragraph{Inference.}
For \nickname, we query points in the first source frame and use $50$ denoising steps on DAVIS and iPhone; Syn4D settings are in \cref{app:robustness}. Baselines use their released sampling settings.

\subsection{Quantitative comparisons}%
\label{sec:quantitative}

\Needspace{11\baselineskip}
\paragraph{Novel-view video synthesis.}
\label{sec:iphone_nvs}
\begin{wraptable}[11]{r}{0.57\textwidth}
\vspace{-1.15\baselineskip}
\centering
\captionsetup{font=footnotesize,skip=3pt}
\caption{\textbf{Novel-view synthesis on iPhone~\cite{gao22monocular:}.} $\alpha$: 4RC; $\beta$: two-stage backend; same generator, no GMOS. ``m'': masked metrics. Bold/underline: best/second best.}%
\label{tab:iphone_nvs}
\fontsize{7.5}{9}\selectfont
\renewcommand{\arraystretch}{1.0}
\setlength{\tabcolsep}{1pt}
\begin{tabular*}{\linewidth}{@{\extracolsep{\fill}}lcccccc@{}}
\toprule
\textbf{Method}
& \textbf{mPSNR}$\uparrow$
& \textbf{mSSIM}$\uparrow$
& \textbf{mLPIPS}$\downarrow$
& \textbf{PSNR}$\uparrow$
& \textbf{SSIM}$\uparrow$
& \textbf{LPIPS}$\downarrow$ \\
\midrule
ReCamMaster
& 10.34
& 0.478
& 0.628
& 10.56
& 0.326
& 0.689 \\
TrajectoryCrafter
& 11.56
& 0.438
& 0.569
& 11.14
& 0.288
& 0.631 \\
EX-4D
& \underline{13.31}
& \textbf{0.520}
& 0.464
& \underline{12.91}
& \textbf{0.366}
& 0.532 \\
Gen3C
& 12.42
& 0.469
& 0.489
& 12.20
& 0.312
& 0.550 \\
Vista4D
& 12.84
& 0.487
& 0.484
& 12.46
& 0.328
& 0.535 \\
\midrule
\rowcolor{gray!10}[0pt][0pt]\oursrowshade \nickname-14B\smash{$^{\alpha}$}
& 13.22
& 0.484
& \underline{0.435}
& 12.68
& 0.326
& \underline{0.509} \\
\rowcolor{gray!10}[0pt][0pt]\oursrowshade \nickname-14B\smash{$^{\beta}$}
& \textbf{14.03}
& \underline{0.507}
& \textbf{0.379}
& \textbf{13.20}
& \underline{0.348}
& \textbf{0.465} \\
\bottomrule
\end{tabular*}
\vspace{-0.5\baselineskip}
\end{wraptable}

\Cref{tab:iphone_nvs} compares generated views with synchronized iPhone recordings. PSNR, SSIM, and LPIPS score full images; masked (``m'') variants score only co-visible pixels.

With the two-stage backend, \nickname obtains the best PSNR ($13.20$\,dB) and LPIPS in both forms, ahead of EX-4D and Vista4D; EX-4D leads in SSIM, and our 4RC variant is second in LPIPS.
Both variants use the same generator without retraining, illustrating the effect of the source geometry and its preprocessing.
\Cref{fig:iphone_nvs} shows visual comparisons; \cref{tab:vidar_comparison} compares with per-scene ViDAR.

\paragraph{Camera control accuracy.}
\Cref{tab:your_result} compares camera control on $100$ DAVIS video--trajectory pairs using VGGT-$\Omega$ camera recovery~\cite{wang26vggt-omega:}. ATE measures trajectory error; RPE-T and RPE-R measure adjacent-frame translation and rotation errors. Translation metrics use $54$ pairs; rotation uses all $100$.
\nickname has the lowest trajectory and translation errors of all methods and lower rotation error than all rendered-geometry baselines.
With both methods using 4RC source reconstruction on DAVIS, \nickname reduces ATE by $54\%$ and RPE-T by $33\%$ compared with the concurrent Vista4D.
ReCamMaster attains a slightly lower rotation error ($0.099^\circ$ versus $0.102^\circ$) but the largest trajectory error of all methods, consistent with outputs that stay close to the source viewpoint.
\paragraph{Video quality.}
\Cref{tab:video_quality} reports five VBench~\cite{huang24vbench:} dimensions on the same DAVIS videos.
\nickname scores highest on imaging quality and, among rendered-geometry baselines, on subject and background consistency; it also exceeds Vista4D on aesthetic quality, with a marginally lower temporal-style score.
ReCamMaster's high consistency accompanies the largest trajectory error: consistency alone does not establish camera fidelity.
FVD computed against source videos can reward copying the input despite failed camera control.
We therefore report FVD only on Syn4D, using ground-truth target-view videos as references; \nickname achieves the lowest FVD with ground-truth source geometry (\cref{tab:syn4d_gt_comparison}).

\begingroup
\addtolength{\parskip}{0pt plus 2pt}

\begin{table}[tb!]
\centering
\captionsetup{font=footnotesize}
\caption{\textbf{Camera control and video quality on DAVIS.} \nickname outperforms rendered-geometry baselines in camera control, imaging quality, and subject and background consistency. ATE: $\times10^2$; RPE-T: $\times10^3$; RPE-R: degrees. Bold and underline mark \textbf{best} and \underline{second-best} results.}
\label{tab:your_result}
\label{tab:video_quality}
\fontsize{8}{10}\selectfont
\renewcommand{\arraystretch}{1.05}
\setlength{\tabcolsep}{2pt}
\begin{tabular*}{\linewidth}{@{\extracolsep{\fill}}llcccccccc@{}}
\toprule
& & \multicolumn{3}{c}{\textbf{Camera control}} & \multicolumn{5}{c}{\textbf{VBench}~\cite{huang24vbench:}} \\
\cmidrule(lr){3-5}\cmidrule(l){6-10}
\textbf{Method} & \textbf{Control} & \textbf{ATE} $\downarrow$ & \textbf{RPE-T} $\downarrow$ & \textbf{RPE-R} $\downarrow$ & \textbf{Aest.} $\uparrow$ & \textbf{Imag.} $\uparrow$ & \textbf{Subj.} $\uparrow$ & \textbf{Backg.} $\uparrow$ & \textbf{Temp.} $\uparrow$ \\
\midrule
ReCamMaster & Extrinsics & 9.547 & 15.37 & \textbf{0.099} & \underline{0.493} & 0.503 & \textbf{0.922} & \textbf{0.942} & \textbf{0.256} \\
\midrule
TrajectoryCrafter & Point cloud & 3.533 & 4.817 & 0.207 & \textbf{0.500} & \underline{0.548} & 0.894 & 0.923 & 0.253 \\
EX-4D & Point cloud & \underline{2.903} & \underline{3.560} & 0.587 & 0.442 & 0.377 & 0.827 & 0.899 & 0.238 \\
Gen3C & Point cloud & 5.168 & 7.581 & 0.386 & 0.454 & 0.450 & 0.856 & 0.911 & 0.248 \\
Vista4D & Point cloud & 3.576 & 3.608 & 0.132 & 0.478 & 0.534 & 0.904 & 0.926 & \underline{0.256} \\
\midrule
\rowcolor{gray!10}[0pt][0pt]\oursrowshade \nickname-14B & 3D tracks & \textbf{1.657} & \textbf{2.415} & \underline{0.102} & 0.493 & \textbf{0.601} & \underline{0.910} & \underline{0.931} & 0.253 \\
\bottomrule
\end{tabular*}
\end{table}
\paragraph{Novel-view synthesis on synthetic scenes.}
\label{sec:robustness}
Syn4D provides exact target views and source geometry, removing reconstruction error from the comparison.
With ground-truth source depth and cameras for rendered-geometry baselines and 3D trajectories for \nickname, \nickname leads on all camera and image metrics (\cref{tab:syn4d_gt_comparison}).
With 4RC estimates on a separate $32$-pair set, \nickname leads on camera errors and LPIPS, while camera-only ReCamMaster leads on PSNR and SSIM (\cref{tab:syn4d_4rc_nvs}).
Test scenes are held out, but Syn4D is in our training mixture and the baseline checkpoints are not adapted to it; real-video evaluations test beyond our training datasets.

\paragraph{Depth consistency of generated views.}
We reconstruct generated-video depth and compare it with exact target depth on Syn4D (\cref{tab:independent_depth}).
\nickname has the lowest AbsRel and highest $\delta_1$ under both output reconstructors, VGGT-$\Omega$ and 4RC; VGGT-$\Omega$ AbsRel falls from Vista4D's $0.628$ to $0.423$.
On iPhone, with reconstructed target-video depth as a pseudo-reference, our two-stage variant also has the lowest AbsRel under both reconstructors, supporting improved generated-view geometry.

\paragraph{Comparison with per-scene optimization.}
ViDAR~\cite{nazarczuk25vidar:} takes about $5$ h $18$ min per iPhone scene on an A6000 for optimization; \nickname needs no scene-specific optimization.
Given the same first $49$ source frames, our two-stage variant has higher PSNR and lower LPIPS, including masked variants, while ViDAR has higher SSIM (\cref{tab:vidar_comparison}); ViDAR natively uses the full sequence.

\paragraph{Sampling cost.}
Because the track condition enters along channels, \nickname adds no tokens to self-attention, whereas Vista4D appends a rendered-video token sequence.
In a matched comparison at $384\times672\times49$ with $50$ denoising steps on one H200, \nickname achieves a $1.34\times$ denoising speedup ($7.39$ versus $9.89$\,s) at similar peak memory (\cref{tab:vista4d_runtime}); end-to-end time per video is nonetheless similar ($570.9$ versus $581.9$\,s), since loading, conditioning, decoding, and writing offset the gain.
Source reconstruction is paid once per input video and reused across target trajectories ($8.8$\,s for 4RC and $11.8$\,s for the two-stage backend in warm A6000 runs, \cref{tab:backend_cost}); only projection and encoding are redone per trajectory.

\begin{figure}[t]
\centering
\includegraphics[width=\textwidth]{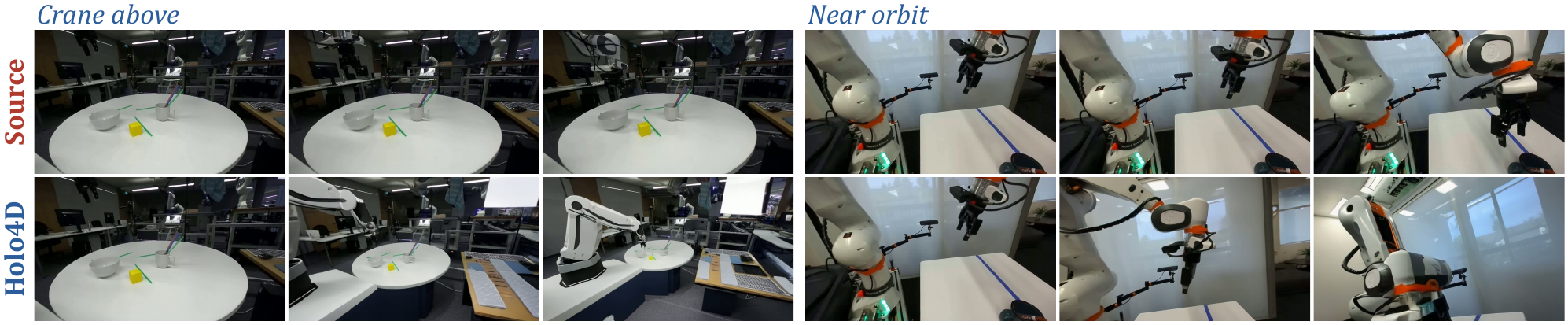}
\caption{\textbf{Zero-shot robot reshooting.} Two DROID~\cite{khazatsky24droid:} episodes (top: source) reshot along a crane-above path and a near orbit (bottom), at $j=0,24,48$. \nickname-14B uses no robot training data; inference uses 4RC and $50$ steps. More examples and track conditions are in \cref{fig:robot}.}
\label{fig:robot_main}
\end{figure}
\begin{table}[t]
\centering
\captionsetup{font=footnotesize}
\caption{\textbf{Novel-view synthesis on Syn4D.} Ground-truth source geometry at inference removes upstream reconstruction error. ATE: $\times10^2$; RPE-T: $\times10^3$; RPE-R: degrees. Bold and underline mark \textbf{best} and \underline{second-best} results.}%
\label{tab:syn4d_gt_comparison}
\renewcommand{\arraystretch}{1.0}
\fontsize{8}{10}\selectfont
\setlength{\tabcolsep}{2pt}
\begin{tabular*}{\linewidth}{@{\extracolsep{\fill}}lccccccc@{}}
\toprule
\textbf{Method}
& \textbf{ATE} $\downarrow$
& \textbf{RPE-T} $\downarrow$
& \textbf{RPE-R} $\downarrow$
& \textbf{PSNR} $\uparrow$
& \textbf{SSIM} $\uparrow$
& \textbf{LPIPS} $\downarrow$
& \textbf{FVD} $\downarrow$ \\
\midrule
ReCamMaster
& 12.68
& 62.20
& 0.749
& 13.66
& 0.118
& 0.236
& 1849 \\
TrajectoryCrafter
& \underline{5.018}
& \underline{20.61}
& \underline{0.225}
& 13.64
& 0.256
& 0.178
& 1878 \\
EX-4D
& 8.163
& 57.89
& 0.563
& 12.69
& 0.273
& 0.271
& 1916 \\
Gen3C
& 5.724
& 40.03
& 0.469
& 13.10
& 0.103
& 0.211
& 1633 \\
Vista4D
& 5.953
& 25.19
& 0.255
& \underline{14.28}
& \underline{0.285}
& \underline{0.151}
& \underline{1214} \\
\midrule
\rowcolor{gray!10}[0pt][0pt]\oursrowshade \nickname-14B
& \textbf{1.065}
& \textbf{6.154}
& \textbf{0.073}
& \textbf{16.52}
& \textbf{0.441}
& \textbf{0.119}
& \textbf{973} \\
\bottomrule
\end{tabular*}
\vspace{-2mm}
\end{table}

\subsection{Qualitative comparisons}
\label{sec:qualitative}

\Cref{fig:qualitative} shows a restaurant side dolly and a car-following shot from Vista4D's benchmark.
ReCamMaster stays close to the source framing; TrajectoryCrafter and Gen3C change or duplicate people and, like EX-4D, lose the car. EX-4D also leaves large holes.
Vista4D turns the foreground diner toward the new camera and omits the car's rear window (red boxes and insets).
\nickname preserves the diner's orientation, rear window, and lights while following the requested path; \cref{tab:your_result,tab:vista4d_native110} quantify camera accuracy, and \cref{fig:baseline_comparison} shows two further trajectories.
On iPhone (\cref{fig:iphone_nvs}), \nickname preserves the person and scene objects at the target viewpoint.

\par\endgroup
\Needspace*{14\baselineskip}
\subsection{Analysis of the geometric scaffold}
\label{sec:ablation}
\begin{figure}[t]
\centering
\includegraphics[width=\textwidth]{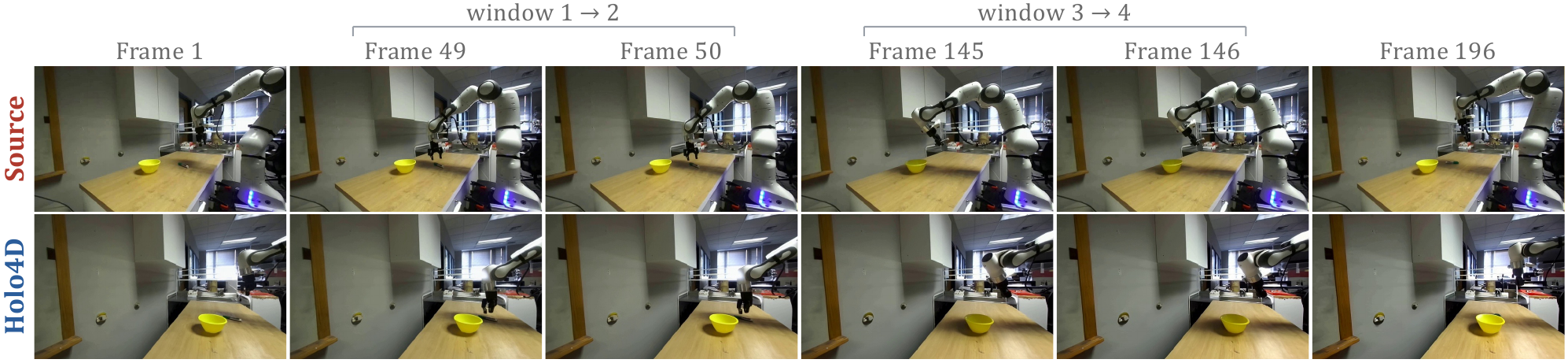}
\caption{\textbf{Long-video rollout.} A $196$-frame DROID~\cite{khazatsky24droid:} episode reshot to a static target view in five $49$-frame windows (4RC, $50$ steps): \nickname-14B for window~1, the I2V continuation model for windows 2--5 (\cref{app:impl}). Brackets mark the window boundaries shown.}
\label{fig:long_rollout}
\end{figure}

\noindent
\begin{wraptable}[8]{r}{0.54\textwidth}
\vspace{-\intextsep}
\centering
\captionsetup{font=footnotesize,skip=3pt}
\caption{\textbf{Stage-1 conditioning design on DAVIS.} All variants condition on trajectories. ATE: $\times10^2$; RPE-T: $\times10^3$; RPE-R: degrees.}%
\label{tab:stage1_design}
\renewcommand{\arraystretch}{1.0}
\fontsize{7.5}{9}\selectfont
\setlength{\tabcolsep}{3pt}
\begin{tabular*}{\linewidth}{@{\extracolsep{\fill}}lccc@{}}
\toprule
\textbf{Condition} & \textbf{ATE} $\downarrow$ & \textbf{RPE-T} $\downarrow$ & \textbf{RPE-R} $\downarrow$ \\
\midrule
w/o source color & 2.327 & \underline{2.125} & 0.094 \\
w/o target depth & 1.882 & 2.135 & \underline{0.068} \\
+ rendered RGB/masks & \underline{1.83}\phantom{0} & 2.48\phantom{0} & 0.098 \\
\midrule
\rowcolor{gray!10}[0pt][0pt]\oursrowshade Full (\nickname) & \textbf{1.621} & \textbf{2.077} & \textbf{0.065} \\
\bottomrule
\end{tabular*}
\vspace{-0.5\baselineskip}
\end{wraptable}
\textbf{Conditioning design.}
\Cref{tab:stage1_design} compares conditioning variants trained with the same stage-1 recipe ($15$k steps at $144\times256$).
Removing the per-track color raises ATE by $44\%$ even though the source video remains available, so track-attached appearance is not redundant with the video.
Removing target depth has a smaller effect, likely because depth follows from the 3D positions and target cameras.
Adding rendered RGB and mask features to the track features improves none of the camera metrics, favoring trajectory-only conditioning for camera control at this training budget.

\FloatBarrier
\begin{wraptable}[14]{r}{0.54\textwidth}
\vspace{-\intextsep}
\centering
\captionsetup{font=footnotesize,skip=3pt}
\caption{\textbf{Reconstruction backends and track noise on Syn4D.} Top: backends ($^*$VGGT-$\Omega$ depth and cameras); bottom: GT tracks with relative noise $\alpha$. ATE $\times10^2$; RPE-R in degrees. No GMOS. Bold/underline: best/second best per block.}
\label{tab:scaffold_analysis_main}
\fontsize{7.5}{9}\selectfont
\renewcommand{\arraystretch}{1.06}
\setlength{\tabcolsep}{1pt}
\begin{tabular*}{\linewidth}{@{\extracolsep{\fill}}lccccc@{}}
\toprule
\textbf{Source geometry} & \textbf{ATE} $\downarrow$ & \textbf{RPE-R} $\downarrow$ & \textbf{PSNR} $\uparrow$ & \textbf{SSIM} $\uparrow$ & \textbf{LPIPS} $\downarrow$ \\
\midrule
Open-D4RT$^*$ & 1.858 & 0.151 & 13.10 & 0.158 & 0.169 \\
V-DPM & 1.660 & \textbf{0.079} & 14.96 & 0.268 & 0.151 \\
4RC & \underline{1.353} & \underline{0.087} & \underline{14.98} & \underline{0.278} & \textbf{0.131} \\
CoWTracker$^*$ & \textbf{1.205} & 0.107 & \textbf{15.09} & \textbf{0.294} & \underline{0.137} \\
\midrule
GT geometry & \textbf{1.065} & \textbf{0.073} & \textbf{16.52} & \textbf{0.441} & \textbf{0.119} \\
$\alpha=0.1$ & 3.033 & \underline{0.221} & 13.44 & \underline{0.171} & \underline{0.172} \\
$\alpha=0.5$ & 2.568 & 0.540 & \underline{13.49} & 0.111 & 0.238 \\
$\alpha=1.0$ & 2.228 & 0.574 & 13.37 & 0.100 & 0.241 \\
$\alpha=4.0$ & \underline{2.163} & 0.604 & 13.26 & 0.093 & 0.242 \\
\bottomrule
\end{tabular*}
\vspace{-0.5\baselineskip}
\end{wraptable}

\noindent\textbf{Reconstruction backends.}
The same generator accepts 4RC, V-DPM, and VGGT-$\Omega$ paired with Open-D4RT or CoWTracker as $\mathcal R$ without retraining (\cref{tab:scaffold_analysis_main}), and the backends trade off differently: the CoWTracker variant gives the best ATE, PSNR, and SSIM, V-DPM the best rotation error, and 4RC the best LPIPS.
Ground-truth geometry improves all five metrics, leaving room for better reconstructors.

\noindent\textbf{Trajectory accuracy.}
To isolate the effect of geometric error, we add Gaussian noise to ground-truth tracks, with standard deviation $\alpha$ times that of the source tracks along each axis, keeping all other inputs fixed.
Even mild noise ($\alpha=0.1$) triples rotation error ($0.073^\circ$ to $0.221^\circ$) and degrades SSIM and LPIPS although the clean source video remains available, showing that clean video context does not fully compensate for geometric error.
ATE does not grow beyond $\alpha=0.1$; independent alignment of each recovered camera trajectory can absorb part of the error.
Full results and protocols are in \cref{app:conditioning,app:robustness}.

\paragraph{Rotation control with accurate geometry.}
With ground-truth source geometry on nine held-out Syn4D clips, \nickname has lower error on all five camera metrics than Vista4D, including source-aligned rotation ($0.242^\circ$ versus $0.372^\circ$; \cref{tab:rotation_attribution}).
Replacing GT with 4RC estimates on the same clips and checkpoint raises this error to $0.931^\circ$ but barely changes relative rotation error ($0.051^\circ$ to $0.055^\circ$), consistent with an orientation offset despite similar local motion. Accurate relative motion therefore does not guarantee the requested viewpoint.

\Needspace{11\baselineskip}
\begin{wraptable}[9]{r}{0.54\textwidth}
\vspace{-\intextsep}
\centering
\captionsetup{font=footnotesize,skip=3pt}
\caption{\textbf{Camera control on Vista4D's $110$ trajectories.} $\alpha$: 4RC; $\beta$: two-stage. ATE $\times10^2$; RPE-T $\times10^3$; rotations in degrees.}%
\label{tab:vista4d_native110}
\renewcommand{\arraystretch}{1.0}
\setlength{\tabcolsep}{1pt}
\fontsize{7.5}{9}\selectfont
\resizebox{\linewidth}{!}{%
\begin{tabular}{@{}lccccc@{}}
\toprule
\multirow{2}{*}{\textbf{Method}}
& \multicolumn{3}{c}{\textbf{VGGT-$\Omega$}}
& \multicolumn{2}{c}{\textbf{Pi3}} \\
\cmidrule(lr){2-4}\cmidrule(lr){5-6}
& \textbf{ATE} $\downarrow$ & \textbf{RPE-T} $\downarrow$ & \textbf{RPE-R} $\downarrow$
& \textbf{TransErr$^2$} $\downarrow$ & \textbf{RotErr} $\downarrow$ \\
\midrule
Vista4D
& \underline{46.04}
& 36.81
& \textbf{0.255}
& 4.573
& \textbf{4.169} \\
\rowcolor{gray!10} \nickname-14B\smash{$^{\alpha}$}
& 46.31
& \underline{21.33}
& 1.142
& \textbf{2.164}
& 14.98 \\
\rowcolor{gray!10} \nickname-14B\smash{$^{\beta}$}
& \textbf{13.76}
& \textbf{9.064}
& \underline{0.301}
& \underline{2.367}
& \underline{6.098} \\
\bottomrule
\end{tabular}}
\vspace{-0.5\baselineskip}
\end{wraptable}

\noindent\textbf{Camera control on Vista4D trajectories.}
On Vista4D's $110$ trajectories (\cref{tab:vista4d_native110}), Vista4D uses its released Pi3 geometry and reprojections, while \nickname uses 4RC or the two-stage backend (protocols in \cref{app:vista4d_native}).
With scalar-depth-aligned 4RC geometry, \nickname has similar ATE, lower RPE-T and Pi3 translation error, but higher rotation errors than Vista4D.
Substituting the two-stage backend cuts rotation error by $74\%$ under VGGT-$\Omega$ and $59\%$ under Pi3 ($14.98^\circ$ to $6.098^\circ$), improves ATE and RPE-T, and slightly raises Pi3 translation error. Pi3 rotation remains above Vista4D's $4.169^\circ$.
The comparison with Vista4D does not isolate the generator.

\subsection{Robot videos}
\label{sec:robot}
\begin{table}[t]
\centering
\captionsetup{font=footnotesize}
\caption{\textbf{Policy learning with Fast-WAM.} Success (\%) after $20$k LIBERO steps. Control and \nickname differ only in video self-attention initialization; the official release is a separate reference. Bold/underline: best/second best.}%
\label{tab:fastwam_main}
\footnotesize
\setlength{\tabcolsep}{3pt}
\begin{tabular*}{\linewidth}{@{\extracolsep{\fill}}lccccccccc@{}}
\toprule
& \multicolumn{2}{c}{\textbf{Overall}} & \multicolumn{7}{c}{\textbf{LIBERO-plus perturbation category}} \\
\cmidrule(lr){2-3}\cmidrule(l){4-10}
\textbf{Initialization} & \textbf{LIBERO} & \textbf{LIB.-plus} & \textbf{Background} & \textbf{Camera} & \textbf{Language} & \textbf{Light} & \textbf{Layout} & \textbf{Robot} & \textbf{Noise} \\
\midrule
Official release & \textbf{96.80} & \underline{50.91} & 50.09 & \textbf{16.82} & 69.94 & 81.79 & 61.97 & 46.00 & \textbf{39.41} \\
Control (Wan2.2) & 95.60 & 50.81 & \underline{56.32} & \underline{9.01} & \underline{70.79} & \underline{85.20} & \textbf{66.62} & \underline{49.10} & 31.73 \\
\rowcolor{gray!10}[0pt][0pt]\oursrowshade \nickname init. & \underline{96.45} & \textbf{52.10} & \textbf{63.38} & 8.01 & \textbf{71.57} & \textbf{89.14} & \underline{66.16} & \textbf{49.16} & \underline{32.92} \\
\bottomrule
\end{tabular*}
\end{table}

\paragraph{Zero-shot reshooting.}
\nickname-14B is trained without robot videos, yet reshoots DROID manipulation episodes and humanoid-robot clips without fine-tuning (\cref{fig:robot_main,fig:robot}).

\paragraph{Long-video rollout.}
We generate overlapping $49$-frame windows: the first with \nickname-14B, then with an image-to-video continuation model conditioned on the last generated frame (\cref{app:impl}).
The continuation model trains on OmniWorld-DROID, so later windows are not zero-shot.
\Cref{fig:long_rollout} qualitatively shows consistent appearance across two window boundaries in a $196$-frame episode.

\paragraph{Transfer to robot policy learning.}
We initialize Fast-WAM's~\cite{yuan26fast-wam:} $30$ video self-attention blocks from a 5B \nickname variant (\cref{app:fastwam}) and train on LIBERO~\cite{liu23libero:} alongside a matched Wan2.2-initialized control (\cref{tab:fastwam_main}).
Results are mixed: LIBERO has few tasks and is near saturation ($95.60\%$ vs.\ $96.45\%$), leaving little room to distinguish the two. On LIBERO-plus~\cite{fei25libero-plus:}, \nickname gains $1.29$ percentage points overall, mainly under background-texture ($+7.1$) and lighting ($+3.9$) perturbations, but loses on the spatial suite. Camera-viewpoint perturbations remain hard for all checkpoints (single seed). Whether novel-view training helps embodied policies remains open and merits further study.

\section{Conclusion}%
\label{sec:conclusion}

\nickname guides video diffusion with persistent 3D trajectories that carry source appearance, leaving regions beyond the scaffold to the generative prior.
It achieves the lowest trajectory and translation errors on DAVIS and the best PSNR and LPIPS on iPhone with the two-stage backend, and works with different reconstructors without retraining; it remains sensitive to geometric error, incomplete source coverage, and difficult camera trajectories.

\paragraph{Future work.}
Since video generators are general-purpose vision learners~\cite{wang26video:}, one model could jointly perform 4D reconstruction and novel-view synthesis so that both improve together. Querying trajectories in every source frame, not just the first (e.g., UniQueR~\cite{peng26uniquer:}), would cover later-appearing content; stronger base models such as MiniMax H3~\cite{minimax26h3:} should improve visual quality.

\begin{ack}

This research was supported by the Sony Research Award Programme 2024 and ERC CoG UNION (101001212). We thank Isambard-AI and Dawn AIRR supercomputers for supporting this project. The authors declare no competing interests.
\end{ack}

\FloatBarrier

{
\small
\setlength{\bibsep}{2pt plus 0.3ex}
\bibliographystyle{unsrt}
\bibliography{vedaldi_general,vedaldi_specific,yslan_general,yslan_specific}

@STRING{ai      = {Artificial Intelligence} }

@STRING{arxiv   = {arXiv} }

@STRING{cvpr    = {CVPR} }

@STRING{eccv    = {ECCV} }

@STRING{iccv    = {ICCV} }

@STRING{iclr    = {ICLR} }

@STRING{icml    = {ICML} }

@STRING{iros    = {IROS} }

@STRING{nips    = {NeurIPS} }

@STRING{rss     = {RSS} }

@STRING{siggraph= {SIGGRAPH} }

@STRING{threedv = {3DV} }

@STRING{tog     = {ACM TOG} }

@Article{chen254dnex:,
  author       = {Chen, Zhaoxi and Liu, Tianqi and Zhuo, Long and Ren, Jiawei
               and Tao, Zeng and Zhu, He and Hong, Fangzhou and Pan, Liang
               and Liu, Ziwei},
  title        = {{4DNeX:} Feed-Forward {4D} Generative Modeling Made Easy},
  journal      = {arXiv},
  year         = {2025},
  volume       = {2508.13154}
}

@InProceedings{go26text-to-3d:,
  author       = {Hyojun Go and Dominik Narnhofer and Goutam Bhat and Prune
               Truong and Federico Tombari and Konrad Schindler},
  title        = {Text-to-{3D} by Stitching a Multi-view Reconstruction Network
               to a Video Generator},
  booktitle    = {ICLR},
  year         = {2026}
}

@InProceedings{he25cameractrl:,
  author    = {Hao He and Yinghao Xu and Yuwei Guo and Gordon Wetzstein and
              Bo Dai and Hongsheng Li and Ceyuan Yang},
  title     = {{CameraCtrl:} Enabling Camera Control for Text-to-Video
              Generation},
  booktitle = {ICLR},
  year      = {2025}
}

@InProceedings{he25cameractrl2:,
  author    = {He, Hao and Yang, Ceyuan and Lin, Shanchuan and Xu, Yinghao
              and Wei, Meng and Gui, Liangke and Zhao, Qi and Wetzstein,
              Gordon and Jiang, Lu and Li, Hongsheng},
  title     = {{CameraCtrl II}: Dynamic Scene Exploration via
              Camera-controlled Video Diffusion Models},
  booktitle = {ICCV},
  year      = {2025}
}

@InProceedings{van-hoorick24generative:,
  author    = {Van Hoorick, Basile and Wu, Rundi and Ozguroglu, Ege and
              Sargent, Kyle and Liu, Ruoshi and Tokmakov, Pavel and Dave,
              Achal and Zheng, Changxi and Vondrick, Carl},
  title     = {Generative Camera Dolly: Extreme Monocular Dynamic Novel View
              Synthesis},
  booktitle = {ECCV},
  year      = {2024}
}

@InProceedings{chen25video:,
  author    = {Chen, Sili and Guo, Hengkai and Zhu, Shengnan and Zhang,
              Feihu and Huang, Zilong and Feng, Jiashi and Kang, Bingyi},
  title     = {{Video Depth Anything}: Consistent Depth Estimation for
              Super-Long Videos},
  booktitle = {CVPR},
  year      = {2025}
}

@InProceedings{wang26pi:,
  author    = {Yifan Wang and Jianjun Zhou and Haoyi Zhu and Wenzheng Chang
              and Yang Zhou and Zizun Li and Junyi Chen and Jiangmiao Pang
              and Chunhua Shen and Tong He},
  title     = {$\pi^3$: Permutation-Equivariant Visual Geometry Learning},
  booktitle = {ICLR},
  year      = {2026}
}

@InProceedings{wang25shape:,
  author       = {Wang, Qianqian and Ye, Vickie and Gao, Hang and Zeng, Weijia
               and Austin, Jake and Li, Zhengqi and Kanazawa, Angjoo},
  title        = {Shape of motion: {4D} reconstruction from a single video},
  booktitle    = {ICCV},
  year         = {2025}
}

@InProceedings{xu25pixel-perfect:,
  author       = {Gangwei Xu and Haotong Lin and Hongcheng Luo and Xianqi Wang and Jingfeng Yao and Lianghui Zhu and Yuechuan Pu and Cheng Chi and Haiyang Sun and Bing Wang and Guang Chen and Hangjun Ye and Sida Peng and Xin Yang},
  title        = {Pixel-perfect depth with semantics-prompted diffusion
               transformers},
  year         = {2025},
  booktitle    = {NeurIPS}
}

@InProceedings{zhu26motioncrafter:,
  author       = {Zhu, Ruijie and Lu, Jiahao and Hu, Wenbo and Han, Xiaoguang
               and Cai, Jianfei and Shan, Ying and Zheng, Chuanxia},
  title        = {{MotionCrafter}: Dense Geometry and Motion Reconstruction with
               a {4D} {VAE}},
  booktitle    = {CVPR},
  year         = {2026}
}

@InProceedings{lin26vista4d:,
  author    = {Lin, {Kuan Heng} and Liu, Zhizheng and Salamanca, Pablo and
              Kant, Yash and Burgert, Ryan and Namekata, Koichi and Zhao,
              Yiwei and Zhou, Bolei and Goldblum, Micah and Debevec, Paul
              and Yu, Ning},
  title     = {{Vista4D}: Video Reshooting with {4D} Point Clouds},
  booktitle = {CVPR},
  year      = {2026}
}

@InProceedings{you25nvs-solver:,
  author    = {Meng You and Zhiyu Zhu and Hui Liu and Junhui Hou},
  title     = {{NVS-Solver:} Video Diffusion Model as Zero-Shot Novel View
              Synthesizer},
  booktitle = {ICLR},
  year      = {2025}
}

@Article{hu25ex-4d:,
  author  = {Tao Hu and Haoyang Peng and Xiao Liu and Yuewen Ma},
  title   = {{EX-4D:} {EXtreme} Viewpoint {4D} Video Synthesis via Depth
            Watertight Mesh},
  journal = arxiv,
  year    = {2025},
  volume  = {2506.05554}
}

@InProceedings{yu25trajectorycrafter:,
  author    = {Mark Yu and Wenbo Hu and Jinbo Xing and Ying Shan},
  title     = {{TrajectoryCrafter:} Redirecting Camera Trajectory for
              Monocular Videos via Diffusion Models},
  booktitle = {ICCV},
  year      = {2025}
}

@InProceedings{zhou26omniworld:,
  author    = {Yang Zhou and Yifan Wang and Jianjun Zhou and Wenzheng Chang
              and Haoyu Guo and Zizun Li and Kaijing Ma and Xinyue Li and
              Yating Wang and Haoyi Zhu and Mingyu Liu and Dingning Liu and
              Jiange Yang and Zhoujie Fu and Junyi Chen and Chunhua Shen and
              Jiangmiao Pang and Kaipeng Zhang and Tong He},
  title     = {{OmniWorld}: A multi-domain and multi-modal dataset for {4D}
              world modeling},
  booktitle = iclr,
  year      = {2026}
}

@InProceedings{huang26gen3r:,
  author    = {Jiaxin Huang and Yuanbo Yang and Bangbang Yang and Lin Ma and
              Yuewen Ma and Yiyi Liao},
  title     = {{Gen3R}: {3D} Scene Generation Meets Feed-Forward
              Reconstruction},
  booktitle = {CVPR},
  year      = {2026}
}

@InProceedings{luo264rc:,
  author    = {Yihang Luo and Shangchen Zhou and Yushi Lan and Xingang Pan
              and Chen Change Loy},
  title     = {{4RC}: {4D} Reconstruction via Conditional Querying Anytime
              and Anywhere},
  booktitle = {ICML},
  year      = {2026}
}

@InProceedings{karhade26any4d:,
  author    = {Jay Karhade and Nikhil Keetha and Yuchen Zhang and
              Tanisha Gupta and Akash Sharma and Sebastian Scherer and Deva
              Ramanan},
  title     = {{Any4D}: Unified Feed-Forward Metric {4D}
              Reconstruction},
  booktitle = {CVPR},
  year      = {2026}
}

@InProceedings{lan26stream3r:,
  author    = {Lan, Yushi and Luo, Yihang and Hong, Fangzhou and Zhou,
              Shangchen and Chen, Honghua and Lyu, Zhaoyang and Yang, Shuai
              and Dai, Bo and Loy, Chen Change and Pan, Xingang},
  title     = {{STream3R}: Scalable Sequential {3D} Reconstruction with
              Causal Transformer},
  booktitle = {ICLR},
  year      = {2026}
}

@InProceedings{zhang26efficiently:,
  author    = {Chuhan Zhang and Guillaume Le Moing and Skanda Koppula and
              Ignacio Rocco and Liliane Momeni and Junyu Xie and Shuyang Sun
              and Rahul Sukthankar and Jo{\"e}lle K. Barral and Raia Hadsell
              and Zoubin Ghahramani and Andrew Zisserman and Junlin Zhang
              and Mehdi S. M. Sajjadi},
  title     = {Efficiently Reconstructing Dynamic Scenes One {D4RT} at a
              Time},
  booktitle = {CVPR},
  year      = {2026}
}

@InProceedings{lin26depth:,
  author    = {Haotong Lin and Sili Chen and Jun Hao Liew and Donny Y. Chen and Zhenyu Li and Yang Zhao and Sida Peng and Hengkai Guo and Xiaowei Zhou and Guang Shi and Jiashi Feng and Bingyi Kang},
  title     = {Depth Anything 3: Recovering the Visual Space from Any
              Views},
  booktitle = {ICLR},
  year      = {2026},
  url       = {https://openreview.net/forum?id=yirunib8l8}
}

@InProceedings{wu25cat4d:,
  author    = {Rundi Wu and Ruiqi Gao and Ben Poole and Alex Trevithick and
              Changxi Zheng and Jonathan T. Barron and Aleksander Holynski},
  title     = {{CAT4D:} Create Anything in {4D} with Multi-View Video
              Diffusion Models},
  booktitle = {CVPR},
  year      = {2025}
}

@InProceedings{bai25recammaster:,
  author    = {Jianhong Bai and Menghan Xia and Xiao Fu and Xintao Wang and
              Lianrui Mu and Jinwen Cao and Zuozhu Liu and Haoji Hu and
              Xiang Bai and Pengfei Wan and Di Zhang},
  title     = {{ReCamMaster:} Camera-Controlled Generative Rendering from A
              Single Video},
  booktitle = {ICCV},
  year      = {2025}
}

@InProceedings{feng25st4rtrack:,
  author    = {Haiwen Feng and Junyi Zhang and Qianqian Wang and Yufei Ye
              and Pengcheng Yu and Michael J. Black and Trevor Darrell and
              Angjoo Kanazawa},
  title     = {{St4RTrack:} Simultaneous {4D} Reconstruction and Tracking in
              the World},
  booktitle = {ICCV},
  year      = {2025}
}

@InProceedings{loshchilov19decoupled:,
  author    = {Loshchilov, Ilya and Hutter, Frank},
  title     = {Decoupled weight decay regularization},
  booktitle = {ICLR},
  year      = {2019}
}

@Article{zhou18stereo:,
  author       = {Zhou, Tinghui and Tucker, Richard and Flynn, John and Fyffe,
               Graham and Snavely, Noah},
  title        = {Stereo magnification: Learning view synthesis using
               multiplane images},
  journal      = {ACM TOG},
  year         = {2018},
  volume       = {37}
}

@InProceedings{wang20tartanair:,
  author    = {Wang, Wenshan and Zhu, Delong and Wang, Xiangwei and Hu,
              Yaoyu and Qiu, Yuheng and Wang, Chen and Hu, Yafei and Kapoor,
              Ashish and Scherer, Sebastian},
  title     = {{TartanAir:} A Dataset to Push the Limits of Visual {SLAM}},
  booktitle = {IROS},
  year      = {2020}
}

@InProceedings{huang18deepmvs:,
  author    = {Huang, Po-Han and Matzen, Kevin and Kopf, Johannes and Ahuja,
              Narendra and Huang, Jia-Bin},
  title     = {{DeepMVS}: Learning Multi-View Stereopsis},
  booktitle = {CVPR},
  year      = {2018}
}

@InProceedings{li25megasam:,
  author    = {Zhengqi Li and Richard Tucker and Forrester Cole and Qianqian
              Wang and Linyi Jin and Vickie Ye and Angjoo Kanazawa and
              Aleksander Holynski and Noah Snavely},
  title     = {{MegaSaM:} Accurate, Fast and Robust Structure and Motion
              from Casual Dynamic Videos},
  booktitle = {CVPR},
  year      = {2025}
}

@InProceedings{liu26trace:,
  author    = {Xinhang Liu and Yuxi Xiao and Donny Y. Chen and Jiashi Feng
              and Yu-Wing Tai and Chi-Keung Tang and Bingyi Kang},
  title     = {Trace Anything: Representing Any Video in {4D} via Trajectory
              Fields},
  booktitle = {ICLR},
  year      = {2026}
}

@InProceedings{keetha26mapanything:,
  author    = {Nikhil Keetha and Norman M{\"u}ller and Johannes
              Sch{\"o}nberger and Lorenzo Porzi and Yuchen Zhang and Tobias
              Fischer and Arno Knapitsch and Duncan Zauss and Ethan Weber
              and Nelson Antunes and Jonathon Luiten and Manuel
              Lopez-Antequera and Samuel Rota Bul{\`o} and Christian
              Richardt and Deva Ramanan and Sebastian Scherer and Peter
              Kontschieder},
  title     = {{MapAnything:} Universal Feed-Forward Metric {3D}
              Reconstruction},
  booktitle = {3DV},
  year      = {2026}
}

@InProceedings{xu254dgt:,
  author       = {Zhen Xu and Zhengqin Li and Zhao Dong and Xiaowei Zhou and
               Richard Newcombe and Zhaoyang Lv},
  title        = {{4DGT}: Learning a {4D} {Gaussian} Transformer Using Real-World
               Monocular Videos},
  year         = {2025},
  booktitle    = {NeurIPS}
}

@InProceedings{khazatsky24droid:,
  author    = {Alexander Khazatsky and Karl Pertsch and Suraj Nair and
              Ashwin Balakrishna and Sudeep Dasari and Siddharth Karamcheti
              and Soroush Nasiriany and Mohan Kumar Srirama and Lawrence
              Yunliang Chen and Kirsty Ellis and Peter David Fagan and Joey
              Hejna and Masha Itkina and Marion Lepert and Yecheng Jason Ma
              and Patrick Tree Miller and Jimmy Wu and Suneel Belkhale and
              Shivin Dass and Huy Ha and Arhan Jain and Abraham Lee and
              Youngwoon Lee and Marius Memmel and Sungjae Park and Ilija
              Radosavovic and Kaiyuan Wang and Albert Zhan and Kevin Black
              and Cheng Chi and Kyle Beltran Hatch and Shan Lin and Jingpei
              Lu and Jean Mercat and Abdul Rehman and Pannag R Sanketi and
              Archit Sharma and Cody Simpson and Quan Vuong and Homer Rich
              Walke and Blake Wulfe and Ted Xiao and Jonathan Heewon Yang
              and Arefeh Yavary and Tony Z. Zhao and Christopher Agia and
              Rohan Baijal and Mateo Guaman Castro and Daphne Chen and Qiuyu
              Chen and Trinity Chung and Jaimyn Drake and Ethan Paul Foster
              and Jensen Gao and Vitor Guizilini and David Antonio Herrera
              and Minho Heo and Kyle Hsu and Jiaheng Hu and Muhammad Zubair
              Irshad and Donovon Jackson and Charlotte Le and Yunshuang Li
              and Kevin Lin and Roy Lin and Zehan Ma and Abhiram Maddukuri
              and Suvir Mirchandani and Daniel Morton and Tony Nguyen and
              Abigail O'Neill and Rosario Scalise and Derick Seale and
              Victor Son and Stephen Tian and Emi Tran and Andrew E. Wang
              and Yilin Wu and Annie Xie and Jingyun Yang and Patrick Yin
              and Yunchu Zhang and Osbert Bastani and Glen Berseth and
              Jeannette Bohg and Ken Goldberg and Abhinav Gupta and Abhishek
              Gupta and Dinesh Jayaraman and Joseph J Lim and Jitendra Malik
              and Roberto Mart{\'\i}n-Mart{\'\i}n and Subramanian
              Ramamoorthy and Dorsa Sadigh and Shuran Song and Jiajun Wu and
              Michael C. Yip and Yuke Zhu and Thomas Kollar and Sergey
              Levine and Chelsea Finn},
  title     = {{DROID:} A Large-Scale In-The-Wild Robot Manipulation
              Dataset},
  booktitle = {RSS},
  year      = {2024}
}

@InProceedings{liu23libero:,
  author    = {Liu, Bo and Zhu, Yifeng and Gao, Chongkai and Feng, Yihao and
              Liu, Qiang and Zhu, Yuke and Stone, Peter},
  title     = {{LIBERO:} Benchmarking Knowledge Transfer for Lifelong Robot
              Learning},
  booktitle = nips,
  year      = {2023}
}

@InProceedings{wang25continuous:,
  author    = {Qianqian Wang and Yifei Zhang and Aleksander Holynski and
              Alexei A. Efros and Angjoo Kanazawa},
  title     = {Continuous {3D} Perception Model with Persistent State},
  booktitle = {CVPR},
  year      = {2025}
}

@InProceedings{yao25uni4d:,
  author    = {Yao, David Yifan and Zhai, Albert J. and Wang, Shenlong},
  title     = {{Uni4D}: Unifying Visual Foundation Models for {4D} Modeling
              from a Single Video},
  booktitle = {CVPR},
  year      = {2025}
}

@InProceedings{xu25geometrycrafter:,
  author       = {Tian-Xing Xu and Xiangjun Gao and Wenbo Hu and Xiaoyu Li and
               Song-Hai Zhang and Ying Shan},
  title        = {{GeometryCrafter:} Consistent Geometry Estimation for
               Open-world Videos with Diffusion Priors},
  year         = {2025},
  booktitle    = {ICCV}
}

@InProceedings{ren25gen3c:,
  author    = {Xuanchi Ren and Tianchang Shen and Jiahui Huang and Huan Ling
              and Yifan Lu and Merlin Nimier-David and Thomas M{\"u}ller and
              Alexander Keller and Sanja Fidler and Jun Gao},
  title     = {{GEN3C:} {3D}-Informed World-Consistent Video Generation with
              Precise Camera Control},
  booktitle = {CVPR},
  year      = {2025}
}

@InProceedings{hu25depthcrafter:,
  author    = {Wenbo Hu and Xiangjun Gao and Xiaoyu Li and Sijie Zhao and
              Xiaodong Cun and Yong Zhang and Long Quan and Ying Shan},
  title     = {{DepthCrafter:} Generating Consistent Long Depth Sequences
              for Open-world Videos},
  booktitle = {CVPR},
  year      = {2025}
}

@InProceedings{ke24repurposing:,
  author    = {Bingxin Ke and Anton Obukhov and Shengyu Huang and Nando
              Metzger and Rodrigo Caye Daudt and Konrad Schindler},
  title     = {Repurposing Diffusion-Based Image Generators for Monocular
              Depth Estimation},
  booktitle = {CVPR},
  year      = {2024}
}

@InProceedings{wang24motionctrl:,
  author       = {Zhouxia Wang and Ziyang Yuan and Xintao Wang and Yaowei Li
               and Tianshui Chen and Menghan Xia and Ping Luo and Ying Shan},
  title        = {{MotionCtrl}: {A} Unified and Flexible Motion Controller for
               Video Generation},
  booktitle    = {SIGGRAPH},
  year         = {2024}
}

@InProceedings{zhang24monst3r:,
  author       = {Junyi Zhang and Charles Herrmann and Junhwa Hur and Varun
               Jampani and Trevor Darrell and Forrester Cole and Deqing Sun
               and Ming-Hsuan Yang},
  title        = {{MonST3R:} A Simple Approach for Estimating Geometry in the
               Presence of Motion},
  year         = {2025},
  booktitle    = {ICLR}
}

@TechReport{brooks24video:,
  author      = {Tim Brooks and Bill Peebles and Connor Holmes and Will DePue
                and Yufei Guo and Li Jing and David Schnurr and Joe Taylor and
                Troy Luhman and Eric Luhman and Clarence Ng and Ricky Wang and
                Aditya Ramesh},
  title       = {Video generation models as world simulators},
  institution = {OpenAI},
  year        = {2024}
}

@InProceedings{yang24deformable:,
  author       = {Ziyi Yang and Xinyu Gao and Wen Zhou and Shaohui Jiao and
               Yuqing Zhang and Xiaogang Jin},
  title        = {Deformable {3D} {Gaussians} for High-Fidelity Monocular Dynamic
               Scene Reconstruction},
  booktitle    = {CVPR},
  year         = {2024}
}

@InProceedings{bahmani24vd3d:,
  author       = {Bahmani, Sherwin and Skorokhodov, Ivan and Siarohin,
               Aliaksandr and Menapace, Willi and Qian, Guocheng and
               Vasilkovsky, Michael and Lee, Hsin-Ying and Wang, Chaoyang and
               Zou, Jiaxu and Tagliasacchi, Andrea and Lindell, David B. and
               Tulyakov, Sergey},
  title        = {{VD3D:} Taming Large Video Diffusion Transformers for {3D}
               Camera Control},
  year         = {2025},
  booktitle    = {ICLR}
}

@InProceedings{huang24vbench:,
  author    = {Ziqi Huang and Yinan He and Jiashuo Yu and Fan Zhang and
              Chenyang Si and Yuming Jiang and Yuanhan Zhang and Tianxing Wu
              and Qingyang Jin and Nattapol Chanpaisit and Yaohui Wang and
              Xinyuan Chen and Limin Wang and Dahua Lin and Yu Qiao and
              Ziwei Liu},
  title     = {{VBench:} Comprehensive Benchmark Suite for Video Generative
              Models},
  booktitle = {CVPR},
  year      = {2024}
}

@InProceedings{liu23flow:,
  author    = {Xingchao Liu and Chengyue Gong and Qiang Liu},
  title     = {Flow Straight and Fast: Learning to Generate and Transfer
              Data with Rectified Flow},
  booktitle = {ICLR},
  year      = {2023}
}

@InProceedings{xiao24spatialtracker:,
  author  = {Yuxi Xiao and Qianqian Wang and Shangzhan Zhang and Nan Xue
            and Sida Peng and Yujun Shen and Xiaowei Zhou},
  title   = {{SpatialTracker:} Tracking Any {2D} Pixels in {3D} Space},
  booktitle = {CVPR},
  year    = {2024}
}

@InProceedings{ling24dl3dv-10k:,
  author       = {Lu Ling and Yichen Sheng and Zhi Tu and Wentian Zhao and
               Cheng Xin and Kun Wan and Lantao Yu and Qianyu Guo and Zixun
               Yu and Yawen Lu and Xuanmao Li and Xingpeng Sun and Rohan
               Ashok and Aniruddha Mukherjee and Hao Kang and Xiangrui Kong
               and Gang Hua and Tianyi Zhang and Bedrich Benes and Aniket
               Bera},
  title        = {{DL3DV-10K:} A Large-Scale Scene Dataset for Deep
               Learning-based {3D} Vision},
  booktitle    = {CVPR},
  year         = {2024}
}

@InProceedings{wang24dust3r:,
  author    = {Shuzhe Wang and Vincent Leroy and Yohann Cabon and Boris
              Chidlovskii and Jerome Revaud},
  title     = {{DUSt3R}: Geometric {3D} Vision Made Easy},
  booktitle = {CVPR},
  year      = {2024}
}

@InProceedings{wu244d-gaussian:,
  author    = {Guanjun Wu and Taoran Yi and Jiemin Fang and Lingxi Xie and
              Xiaopeng Zhang and Wei Wei and Wenyu Liu and Qi Tian and
              Xinggang Wang},
  title     = {{4D} {Gaussian} Splatting for Real-Time Dynamic Scene
              Rendering},
  booktitle = {CVPR},
  year      = {2024}
}

@InProceedings{greff22kubric:,
  author    = {Klaus Greff and Francois Belletti and Lucas Beyer and Carl
              Doersch and Yilun Du and Daniel Duckworth and David J Fleet
              and Dan Gnanapragasam and Florian Golemo and Charles Herrmann
              and Thomas Kipf and Abhijit Kundu and Dmitry Lagun and Issam
              Laradji and Hsueh-Ti (Derek) Liu and Henning Meyer and Yishu
              Miao and Derek Nowrouzezahrai and Cengiz Oztireli and Etienne
              Pot and Noha Radwan and Daniel Rebain and Sara Sabour and
              Mehdi S. M. Sajjadi and Matan Sela and Vincent Sitzmann and
              Austin Stone and Deqing Sun and Suhani Vora and Ziyu Wang and
              Tianhao Wu and Kwang Moo Yi and Fangcheng Zhong and Andrea
              Tagliasacchi},
  title     = {Kubric: a scalable dataset generator},
  booktitle = {CVPR},
  year      = {2022}
}

@InProceedings{liu23zero-1-to-3:,
  author    = {Ruoshi Liu and Rundi Wu and Basile Van Hoorick and Pavel
              Tokmakov and Sergey Zakharov and Carl Vondrick},
  title     = {Zero-1-to-3: Zero-shot One Image to {3D} Object},
  booktitle = iccv,
  year      = {2023}
}

@Article{park21hypernerf:,
  author       = {Keunhong Park and Utkarsh Sinha and Peter Hedman and Jonathan
               T. Barron and Sofien Bouaziz and Dan B. Goldman and Ricardo
               Martin{-}Brualla and Steven M. Seitz},
  title        = {{HyperNeRF}: a higher-dimensional representation for
               topologically varying neural radiance fields},
  journal      = {ACM TOG},
  year         = {2021},
  volume       = {40},
  number       = {6}
}

@InProceedings{gao22monocular:,
  author    = {Hang Gao and Ruilong Li and Shubham Tulsiani and Bryan
              Russell and Angjoo Kanazawa},
  title     = {Monocular Dynamic View Synthesis: {A} Reality Check},
  booktitle = {NeurIPS},
  year      = {2022}
}

@InProceedings{li23dynibar:,
  author    = {Zhengqi Li and Qianqian Wang and Forrester Cole and Richard
              Tucker and Noah Snavely},
  title     = {{DynIBaR}: Neural Dynamic Image-Based Rendering},
  booktitle = {CVPR},
  year      = {2023}
}

@InProceedings{wang21ibrnet:,
  author       = {Qianqian Wang and Zhicheng Wang and Kyle Genova and Pratul P.
               Srinivasan and Howard Zhou and Jonathan T. Barron and Ricardo
               Martin{-}Brualla and Noah Snavely and Thomas A. Funkhouser},
  title        = {{IBRNet}: Learning Multi-View Image-Based Rendering},
  booktitle    = {CVPR},
  year         = {2021}
}

@InProceedings{park21nerfies:,
  author    = {Keunhong Park and Utkarsh Sinha and Jonathan T. Barron and
              Sofien Bouaziz and Dan B. Goldman and Steven M. Seitz and
              Ricardo Martin{-}Brualla},
  title     = {Nerfies: Deformable Neural Radiance Fields},
  booktitle = {ICCV},
  year      = {2021}
}

@InProceedings{yoon20novel:,
  author    = {Jae Shin Yoon and Kihwan Kim and Orazio Gallo and Hyun Soo
              Park and Jan Kautz},
  title     = {Novel View Synthesis of Dynamic Scenes With Globally Coherent
              Depths From a Monocular Camera},
  booktitle = {CVPR},
  year      = {2020}
}

@InProceedings{pumarola21d-nerf:,
  author    = {Albert Pumarola and Enric Corona and Gerard Pons{-}Moll and
              Francesc Moreno{-}Noguer},
  title     = {{D-NeRF}: Neural Radiance Fields for Dynamic Scenes},
  booktitle = {CVPR},
  year      = {2021}
}

@InProceedings{namekata26gowiththetrack:,
  author    = {Namekata, Koichi and Kant, Yash and Liu, Zhizheng and Burgert, Ryan D. and Xu, Yuancheng and Lin, Kuan Heng and Steven, Emmett and Philip, Julien and Ma, Li and Vedaldi, Andrea and Debevec, Paul and Yu, Ning},
  title     = {Go-with-the-Track: Video Compositing and Motion Control with Point Tracking},
  booktitle = {SIGGRAPH},
  year      = {2026}
}

@InProceedings{jiang26syn4d:,
  author    = {Zeren Jiang and Yushi Lan and Yihang Luo and Yufan Deng and
              Zihang Lai and Edgar Sucar and Christian Rupprecht and Iro
              Laina and Diane Larlus and Chuanxia Zheng and Andrea Vedaldi},
  title     = {{Syn4D}: A Multiview Synthetic {4D} Dataset},
  booktitle = {ECCV},
  year      = {2026}
}

@InProceedings{wang26vggt-omega:,
  author    = {Jianyuan Wang and Minghao Chen and Shangzhan Zhang and Nikita
              Karaev and Johannes Sch{\"o}nberger and Patrick Labatut and
              Piotr Bojanowski and David Novotny and Andrea Vedaldi and
              Christian Rupprecht},
  title     = {{VGGT-$\Omega$}},
  booktitle = {CVPR},
  year      = {2026}
}

@InProceedings{lai26cowtracker:,
  author    = {Zihang Lai and Eldar Insafutdinov and Edgar Sucar and Andrea
              Vedaldi},
  title     = {{CoWTracker}: Tracking by Warping instead of Correlation},
  booktitle = {CVPR},
  year      = {2026}
}

@InProceedings{sucar26v-dpm:,
  author    = {Edgar Sucar and Eldar Insafutdinov and Zihang Lai and Andrea
              Vedaldi},
  title     = {{V-DPM}: Video Reconstruction with Dynamic Point Maps},
  booktitle = {CVPR},
  year      = {2026}
}

@InProceedings{jiang25geo4d:,
  author    = {Zeren Jiang and Chuanxia Zheng and Iro Laina and Diane Larlus
              and Andrea Vedaldi},
  title     = {{Geo4D}: Leveraging Video Generators for Geometric {4D} Scene
              Reconstruction},
  booktitle = {ICCV},
  year      = {2025}
}

@InProceedings{sucar25dynamic:,
  author    = {Edgar Sucar and Zihang Lai and Eldar Insafutdinov and Andrea
              Vedaldi},
  title     = {{Dynamic Point Maps}: A Versatile Representation for Dynamic
              {3D} Reconstruction},
  booktitle = {ICCV},
  year      = {2025}
}

@InProceedings{wang25vggt:,
  author    = {Jianyuan Wang and Minghao Chen and Nikita Karaev and Andrea
              Vedaldi and Christian Rupprecht and David Novotny},
  title     = {{VGGT}: Visual Geometry Grounded Transformer},
  booktitle = {CVPR},
  year      = {2025}
}

@InProceedings{chen24mvsplat360:,
  author    = {Yuedong Chen and Chuanxia Zheng and Haofei Xu and Bohan
              Zhuang and Andrea Vedaldi and Tat-Jen Cham and Jianfei Cai},
  title     = {{MVSplat360}: Feed-Forward 360 Scene Synthesis from Sparse Views},
  booktitle = {NeurIPS},
  year      = {2024}
}

@InProceedings{szymanowicz25flash3d,
  author    = {Stanislaw Szymanowicz and Eldar Insafutdinov and Chuanxia
              Zheng and Dylan Campbell and Jo{\~a}o F. Henriques and
              Christian Rupprecht and Andrea Vedaldi},
  title     = {{Flash3D}: Feed-Forward Generalisable {3D} Scene
              Reconstruction from a Single Image},
  booktitle = threedv,
  year      = {2025}
}

@InProceedings{zheng24free3d,
  author    = {Chuanxia Zheng and Andrea Vedaldi},
  title     = {{Free3D}: Consistent Novel View Synthesis without {3D}
              Representation},
  booktitle = cvpr,
  year      = {2024}
}

@InProceedings{peng26uniquer:,
  author        = {Chensheng Peng and Quentin Herau and Jiezhi Yang and Yichen Xie and Yihan Hu and Wenzhao Zheng and Matthew Strong and Masayoshi Tomizuka and Wei Zhan},
  title         = {{UniQueR}: Unified Query-based Feedforward {3D} Reconstruction},
  booktitle     = {ECCV},
  year          = {2026},
  archivePrefix = {arXiv},
  eprint        = {2603.22851},
  url           = {https://uniquer3d.github.io/}
}

@String(CVPR  = {CVPR})

@String(ICCV  = {ICCV})

@String(ECCV  = {ECCV})

@String(NIPS  = {NeurIPS})

@String(TOG   = {ACM TOG})

@String(ICLR  = {ICLR})

@InProceedings{tancik2020fourier:,
  author    = {Matthew Tancik and Pratul P. Srinivasan and Ben Mildenhall and Sara Fridovich-Keil and Nithin Raghavan and Utkarsh Singhal and Ravi Ramamoorthi and Jonathan T. Barron and Ren Ng},
  title     = {Fourier Features Let Networks Learn High Frequency Functions in Low Dimensional Domains},
  booktitle = {NeurIPS},
  year      = {2020}
}

@InProceedings{zhang2023adding,
  author    = {Lvmin Zhang and Anyi Rao and Maneesh Agrawala},
  title     = {Adding Conditional Control to Text-to-Image Diffusion Models},
  booktitle = {ICCV},
  year      = {2023}
}

@InProceedings{sitzmann2021lfns,
  author    = {Sitzmann, Vincent
              and Rezchikov, Semon
              and Freeman, William T.
              and Tenenbaum, Joshua B.
              and Durand, Fredo},
  title     = {Light Field Networks: Neural Scene Representations
              with Single-Evaluation Rendering},
  booktitle = {NeurIPS},
  year      = {2021}
}

@Article{wan25wan:,
  author  = {Team Wan and Ang Wang and Baole Ai and Bin Wen and Chaojie Mao and Chen-Wei Xie and Di Chen and Feiwu Yu and Haiming Zhao and Jianxiao Yang and Jianyuan Zeng and Jiayu Wang and Jingfeng Zhang and Jingren Zhou and Jinkai Wang and Jixuan Chen and Kai Zhu and Kang Zhao and Keyu Yan and Lianghua Huang and Mengyang Feng and Ningyi Zhang and Pandeng Li and Pingyu Wu and Ruihang Chu and Ruili Feng and Shiwei Zhang and Siyang Sun and Tao Fang and Tianxing Wang and Tianyi Gui and Tingyu Weng and Tong Shen and Wei Lin and Wei Wang and Wei Wang and Wenmeng Zhou and Wente Wang and Wenting Shen and Wenyuan Yu and Xianzhong Shi and Xiaoming Huang and Xin Xu and Yan Kou and Yangyu Lv and Yifei Li and Yijing Liu and Yiming Wang and Yingya Zhang and Yitong Huang and Yong Li and You Wu and Yu Liu and Yulin Pan and Yun Zheng and Yuntao Hong and Yupeng Shi and Yutong Feng and Zeyinzi Jiang and Zhen Han and Zhi-Fan Wu and Ziyu Liu},
  title   = {Wan: Open and Advanced Large-Scale Video Generative Models},
  journal = {arXiv},
  volume = {2503.20314},
  year    = {2025}
}

@InProceedings{chung23unimax:,
  author       = {Chung, Hyung Won and Constant, Noah and Garcia, Xavier and Roberts, Adam and Tay, Yi and Narang, Sharan and Firat, Orhan},
  title        = {{UniMax}: Fairer and more effective language sampling for large-scale multilingual pretraining},
  year         = {2023},
  booktitle    = {ICLR}
}

@article{pont-tuset17davis:,
  author       = {Jordi Pont-Tuset and Federico Perazzi and Sergi Caelles and Pablo Arbel\'aez and Alexander Sorkine-Hornung and Luc {Van Gool}},
  title        = {The 2017 {DAVIS} Challenge on Video Object Segmentation},
  journal      = {arXiv},
  volume = {1704.00675},
  year         = {2017}
}

@InProceedings{xie26lavr:,
  author    = {Mingyang Xie and Numair Khan and Tianfu Wang and Naina Dhingra and Seonghyeon Nam and Haitao Yang and Zhuo Hui and Christopher Metzler and Andrea Vedaldi and Hamed Pirsiavash and Lei Luo},
  title     = {{LaVR}: Scene Latent Conditioned Generative Video Trajectory Re-Rendering using Large {4D} Reconstruction Models},
  booktitle = {CVPR},
  year      = {2026}
}

@Article{zhao23fsdp:,
  author       = {Yanli Zhao and Andrew Gu and Rohan Varma and Liang Luo and Chien-Chin Huang and Min Xu and Less Wright and Hamid Shojanazeri and Myle Ott and Sam Shleifer and Alban Desmaison and Can Balioglu and Pritam Damania and Bernard Nguyen and Geeta Chauhan and Yuchen Hao and Ajit Mathews and Shen Li},
  title        = {{PyTorch} {FSDP}: Experiences on Scaling Fully Sharded Data Parallel},
  year         = {2023},
  journal      = {PVLDB},
  volume       = {16},
  number       = {12},
  pages        = {3848--3860}
}

@inproceedings{wang26fused:,
  author       = {Chaoyang Wang and Ashkan Mirzaei and Vidit Goel and Willi Menapace and Aliaksandr Siarohin and Michael Vasilkovsky and Ivan Skorokhodov and Vladislav Shakhrai and Sergei Korolev and Sergey Tulyakov and Peter Wonka},
  title        = {Fused View-Time Attention and Feedforward Reconstruction for {4D} Scene Generation},
  booktitle    = {NeurIPS},
  year         = {2026},
  url          = {https://openreview.net/forum?id=BDZpNjIC1y}
}

@InProceedings{nazarczuk25vidar:,
  author    = {Nazarczuk, Michal and Catley-Chandar, Sibi and Tanay, Thomas and Zhang, Zhensong and Slabaugh, Gregory and Pérez-Pellitero, Eduardo},
  title     = {{ViDAR: Video Diffusion-Aware 4D Reconstruction From Monocular Inputs}},
  booktitle = {NeurIPS},
  year      = {2025}
}

@InProceedings{ha25geometryguided:,
  author    = {Ha, Hyunho and Xiao, Lei and Richardt, Christian and
              Nguyen-Phuoc, Thu and Kim, Changil and Kim, Min H. and
              Lanman, Douglas and Khan, Numair},
  title     = {Geometry-guided Online {3D} Video Synthesis with Multi-View
              Temporal Consistency},
  booktitle = {CVPR},
  year      = {2025}
}

@Article{zhang26molmomotion:,
  author    = {Jianing Zhang and Chenhao Zheng and Yajun Yang and Max Argus and Rustin Soraki and Winson Han and Taira Anderson and Chun-Liang Li and Shuo Liu and Jiafei Duan and Zhongzheng Ren and Jieyu Zhang and Ranjay Krishna},
  title     = {{MolmoMotion}: Forecasting Point Trajectories in {3D} with Language Instruction},
  journal   = {arXiv},
  volume = {2606.18558},
  year      = {2026},
  url       = {https://arxiv.org/abs/2606.18558}
}

@InProceedings{fu26plenoptic:,
  author    = {Fu, Xiao and Tang, Shitao and Shi, Min and Liu, Xian and Gu, Jinwei and Liu, Ming-Yu and Lin, Dahua and Lin, Chen-Hsuan},
  title     = {Plenoptic Video Generation},
  booktitle = {CVPR},
  year      = {2026}
}

@Article{yuan26fast-wam:,
  author  = {Tianyuan Yuan and Zibin Dong and Yicheng Liu and Hang Zhao},
  title   = {{Fast-WAM}: Do World Action Models Need Test-time Future Imagination?},
  journal = {arXiv},
  volume  = {2603.16666},
  year    = {2026}
}

@Article{fei25libero-plus:,
  author  = {Senyu Fei and Siyin Wang and Junhao Shi and Zihao Dai and Jikun Cai and Pengfang Qian and Li Ji and Xinzhe He and Shiduo Zhang and Zhaoye Fei and Jinlan Fu and Jingjing Gong and Xipeng Qiu},
  title   = {{LIBERO-Plus}: In-depth Robustness Analysis of Vision-Language-Action Models},
  journal = {arXiv},
  volume  = {2510.13626},
  year    = {2025}
}

@Article{carion25sam3:,
  author  = {Nicolas Carion and Laura Gustafson and Yuan-Ting Hu and Shoubhik Debnath and Ronghang Hu and Didac Suris and Chaitanya Ryali and Kalyan Vasudev Alwala and Haitham Khedr and Andrew Huang and Jie Lei and Tengyu Ma and Baishan Guo and Arpit Kalla and Markus Marks and Joseph Greer and Meng Wang and Peize Sun and Roman R{\"a}dle and Triantafyllos Afouras and Effrosyni Mavroudi and Katherine Xu and Tsung-Han Wu and Yu Zhou and Liliane Momeni and Rishi Hazra and Shuangrui Ding and Sagar Vaze and Francois Porcher and Feng Li and Siyuan Li and Aishwarya Kamath and Ho Kei Cheng and Piotr Doll{\'a}r and Nikhila Ravi and Kate Saenko and Pengchuan Zhang and Christoph Feichtenhofer},
  title   = {{SAM 3}: Segment Anything with Concepts},
  journal = {arXiv},
  volume  = {2511.16719},
  year    = {2025},
  url     = {https://arxiv.org/abs/2511.16719}
}

@InProceedings{wang26video:,
  author    = {Letian Wang and Chuhan Zhang and Rishabh Kabra and Jasper Uijlings and Steven Waslander and Andrew Zisserman and Joao Carreira and Kaiming He and Misha Andriluka and Eduard Gabriel Bazavan and Andrei Zanfir and Cristian Sminchisescu},
  title     = {Video Generation Models are General-Purpose Vision Learners},
  booktitle = {ECCV},
  year      = {2026}
}

@Misc{minimax26h3:,
  author       = {{MiniMax-AI}},
  title        = {{MiniMax H3}},
  howpublished = {Official GitHub repository},
  year         = {2026}
}

@inproceedings{voleti2024sv3d,
  author    = {Voleti, Vikram and Yao, Chun-Han and Boss, Mark and Letts, Adam and Pankratz, David and Tochilkin,  Dmitrii and Laforte, Christian and Rombach, Robin and Jampani, Varun},
  title     = {{SV3D}: Novel Multi-view Synthesis and {3D} Generation from a Single Image using Latent Video Diffusion},
  booktitle = {ECCV},
  year      = {2024},
}

@inproceedings{lee2026generative,
  title={Generative video motion editing with 3d point tracks},
  author={Lee, Yao-Chih and Zhang, Zhoutong and Huang, Jiahui and Wang, Jui-Hsien and Lee, Joon-Young and Huang, Jia-Bin and Shechtman, Eli and Li, Zhengqi},
  booktitle={CVPR},
  pages={18306--18318},
  year={2026}
}

@inproceedings{zhang2026plan4d,
  title={Plan4D: Generative Plannable 4D Worlds},
  author={Zhang, Hangwei and Dang, Rujing and Lin, Ruimin and Wu, Jiahong and Chu, Xiangxiang and Zheng, Chuanxia},
  booktitle={NeurIPS},
  year={2026}
}

@InProceedings{xie26gmos:,
  author    = {Junyu Xie and Tengda Han and Weidi Xie and Andrew Zisserman},
  title     = {{GMOS}: Grounding Moving Object Segmentation in {3D} Space and Time},
  booktitle = {NeurIPS},
  year      = {2026}
}
}

\appendix
\raggedbottom
\makeatletter
\setlength{\@fptop}{0pt}
\setlength{\@fpsep}{12pt}
\makeatother
\crefalias{section}{appendix}
\crefalias{subsection}{appendix}

\clearpage
\section{Changes from the Submitted Version}
\label{app:changes}
\begin{figure}[!b]
\centering
\includegraphics[width=\textwidth]{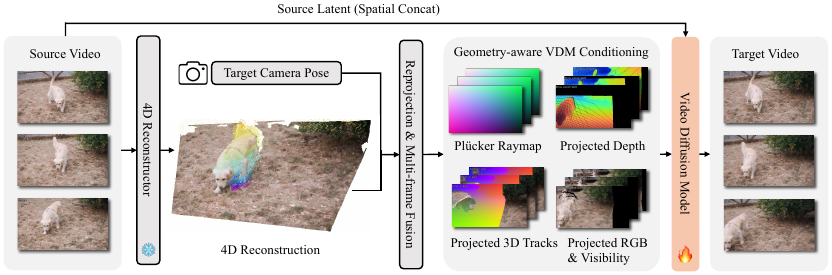}
\caption{\textbf{Submitted pipeline} (reproduced from the submitted version). Static and dynamic source pixels were rendered into a hybrid target-view RGB-D cache (projected RGB, depth, and visibility), which conditioned the video diffusion model together with Pl\"ucker rays and projected 3D tracks. This version removes the rendered RGB-D branch and keeps the track branch (\cref{fig:method_pipeline}).}
\label{fig:sub_pipeline}
\vspace{8pt}
\includegraphics[width=\textwidth]{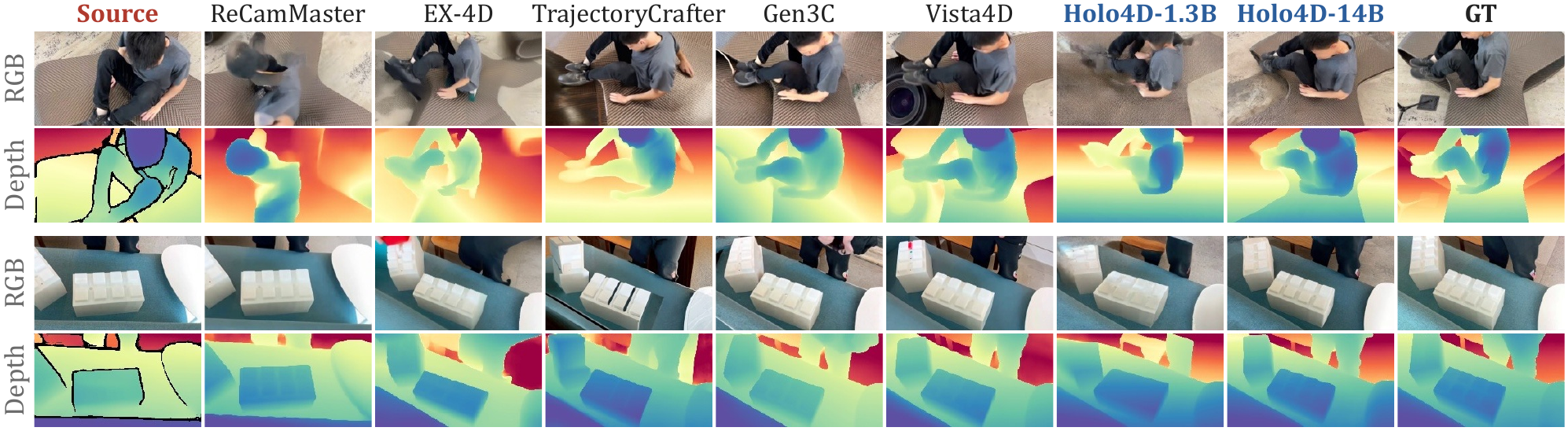}
\caption{\textbf{Submitted novel-view RGB and depth on iPhone} (frames from the submitted version). Two scenes (spin, block), each shown as a target-view RGB row over a depth row; GT is the held-out view, and its depth is a 4RC pseudo-reference. \nickname-1.3B and \nickname-14B are the submitted checkpoints. Final-model results are in \cref{fig:iphone_nvs}.}
\label{fig:sub_iphone}
\end{figure}

\begin{figure}[p]
\centering
\includegraphics[width=\textwidth]{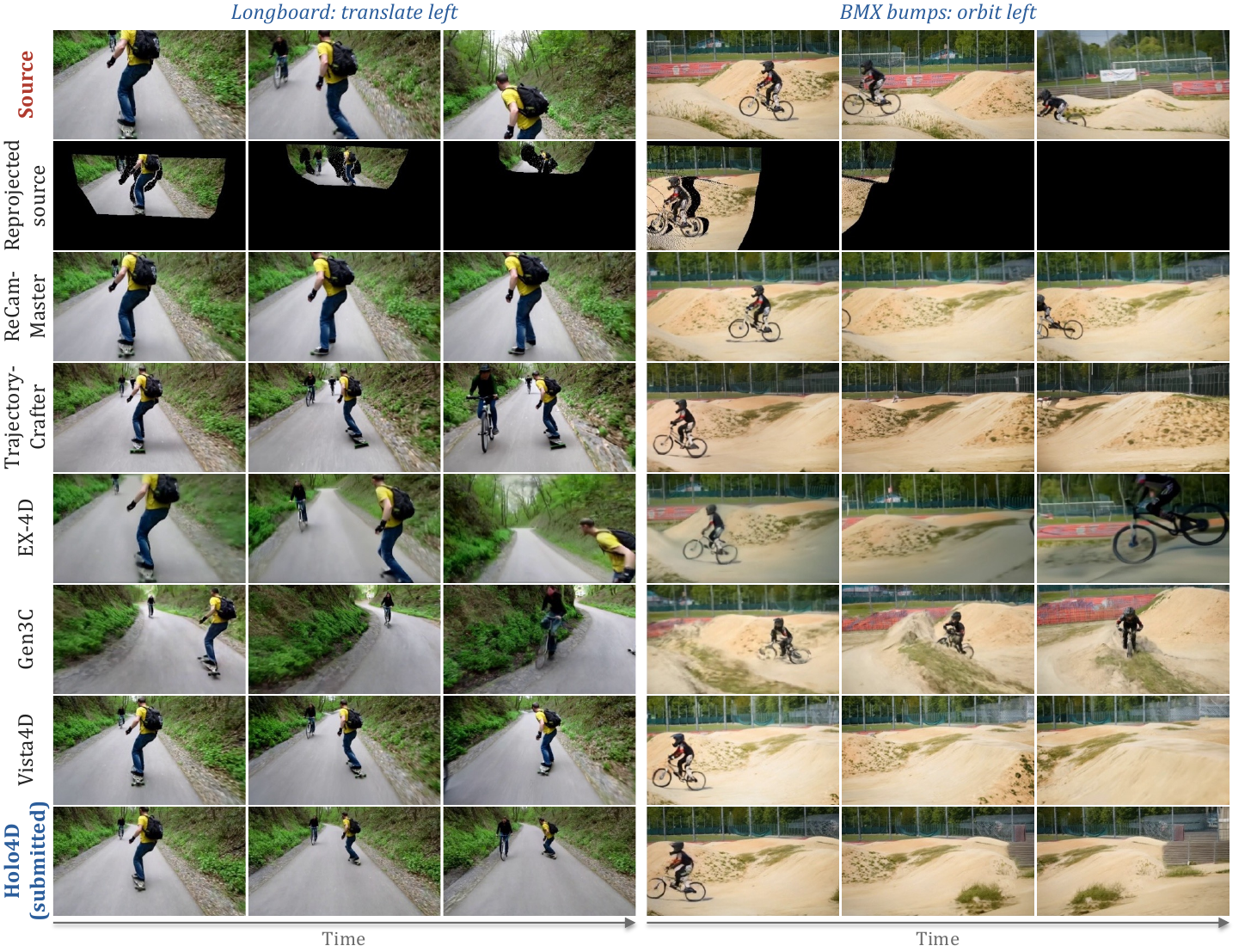}
\caption{\textbf{Submitted qualitative comparison on DAVIS} (frames from the submitted version). ``Holo4D (submitted)'' is the submitted 14B checkpoint with rendered RGB-D and track conditioning; the second row shows source pixels reprojected into the target view (labelled GT in the submission). See \cref{fig:qualitative,fig:baseline_comparison} for the final model.}
\label{fig:sub_davis}
\vspace{8pt}
\includegraphics[width=\textwidth]{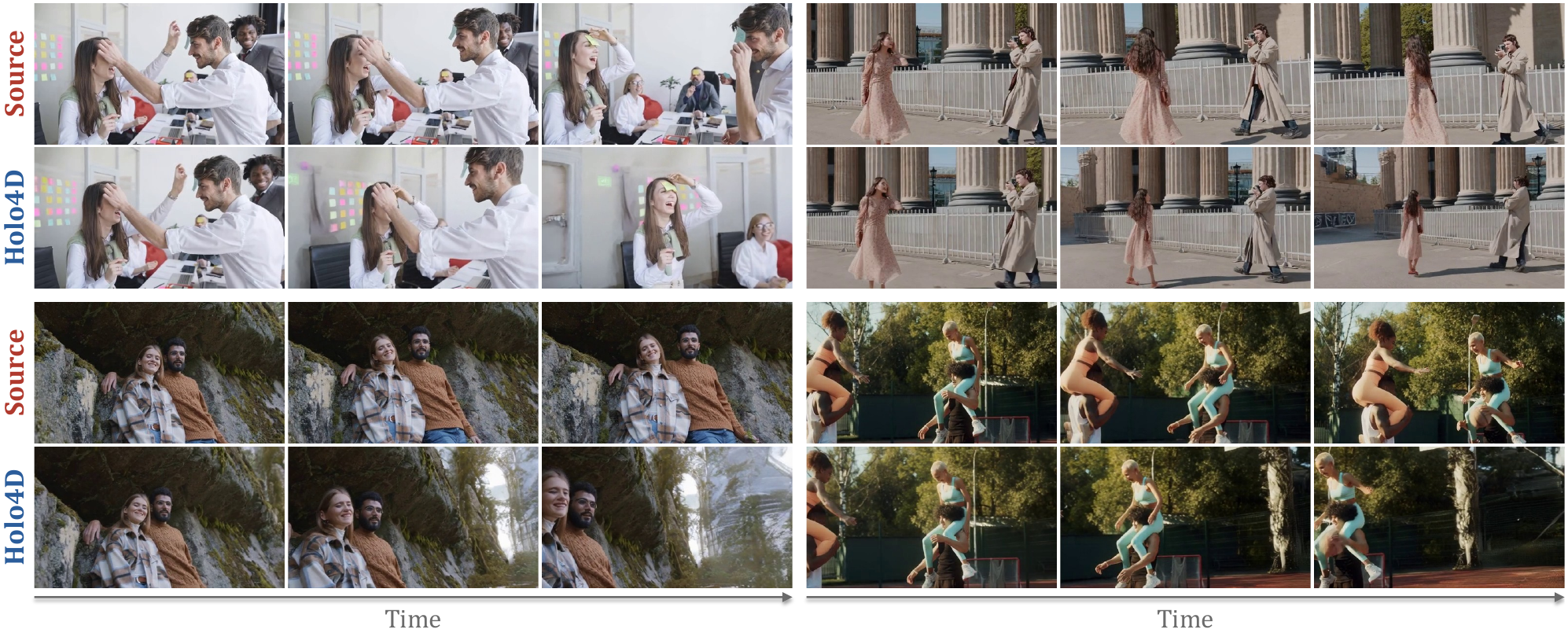}
\caption{\textbf{Submitted qualitative results on real-world videos} (frames from the submitted version). For each of four videos, source frames are shown above the target-view frames generated by the submitted model; three of the five frames shown in the submission are kept. Final-model results are in \cref{fig:supp_vis}.}
\label{fig:sub_realworld}
\end{figure}

The task, title, and central idea are unchanged: \nickname exposes source-view 4D reconstruction to a pretrained video diffusion model to reshoot a monocular video along a new camera path.
After review, we simplified conditioning, retrained the model, and expanded evaluation. Changed checkpoints, resolutions, and camera evaluators preclude direct comparison with submitted numbers.
\Cref{tab:changes} summarizes the differences; the main one is that the rendered RGB-D (depth) branch is removed and only the 3D-track branch is kept.
\Cref{fig:sub_pipeline} reproduces the submitted pipeline, and \cref{fig:sub_davis,fig:sub_iphone,fig:sub_realworld} reproduce its qualitative results.

\begin{table}[h]
\centering
\captionsetup{font=footnotesize}
\caption{\textbf{Submitted version versus this version.} The submitted column summarizes the described design; the rendering used for the submitted results is distinguished below.}
\label{tab:changes}
\footnotesize
\setlength{\tabcolsep}{4pt}
\begin{tabular}{@{}>{\raggedright\arraybackslash}p{0.17\linewidth}>{\raggedright\arraybackslash}p{0.385\linewidth}>{\raggedright\arraybackslash}p{0.385\linewidth}@{}}
\toprule
 & Submitted version & This version \\
\midrule
Source video & VAE latents, width-concatenated with target latents & Unchanged \\
RGB-D branch & Target-view RGB, normalized depth, and validity rendered from a static/dynamic point cache; Conv3d geometry tokens & Removed \\
Track branch & Tracked positions; z-buffer aggregation in the described design; fused with RGB-D tokens & Positions, source color, target-camera depth, and validity; channel-wise max with occupancy; injected along channels in every DiT block \\
Target cameras & Pl\"ucker rays & Pl\"ucker rays, encoded into camera features \\
Backbone & Wan2.1-I2V-14B; Wan2.1-T2V-1.3B & Wan2.1-T2V-14B \\
Resolution & $144\times256$, then $288\times512$ & $144\times256$, then $384\times672$ \\
Depth & Separate readout head on VAE decoder features & No readout; depth reconstructed from generated videos \\
Camera evaluator & ViPE & VGGT-$\Omega$ and Pi3 \\
Video quality & FVD and CLIP on DAVIS & VBench on DAVIS; FVD on Syn4D \\
Robot videos & -- & Reshooting, long videos, policy transfer \\
\bottomrule
\end{tabular}
\end{table}

\paragraph{Submitted conditioning.}
The submitted paper described four conditioning signals: the source video, a hybrid RGB-D cache, target-camera Pl\"ucker rays, and 3D-track tokens.
A 3D-motion threshold separated static and dynamic pixels. Static pixels were fused from $K=7$ keyframes selected by farthest-point sampling over camera positions and viewing directions; dynamic pixels came from the synchronized source frame. The described design merged their renders by z-buffer selection into target-view RGB, normalized depth, and validity.
Track tokens encoded 3D trajectory positions and were fused with the RGB-D tokens.
The actual submitted 14B DAVIS and iPhone results used multiframe softsplat RGB-D conditioning; the 1.3B DAVIS results used single-frame z-buffer conditioning. Both included 3D tracks.

\paragraph{Final conditioning.}
This version removes the rendered RGB-D branch, which supplied target-view RGB, depth, and validity, and keeps only the 3D-track branch, together with the source video and target-camera rays (\cref{sec:method}, \cref{fig:method_pipeline} versus \cref{fig:sub_pipeline}).
Observations encode source color and target-camera depth (\cref{eq:track_payload}); validity combines upstream validity with in-image projection, without the occlusion test of z-buffer visibility.
Since a static point is a trajectory with a constant position, the static--dynamic threshold and keyframe selection are no longer needed.
In a controlled stage-1 comparison, adding rendered RGB and masks to the trajectory features did not improve camera control (\cref{tab:stage1_design}); \cref{tab:conditioning_1p3b,tab:ablation_progressive,tab:ablation_nvs} report further comparisons with rendered conditioning from earlier checkpoints.
Track features enter every DiT block along channels and add no attention tokens (\cref{app:efficiency}).

\paragraph{Model and training.}
The submission reported a 14B model built on Wan2.1-I2V-14B and a 1.3B model built on Wan2.1-T2V-1.3B, both trained at $144\times256$ and $288\times512$.
The main results in this version use a 14B model initialized from Wan2.1-T2V-14B and trained at $144\times256$ and then $384\times672$, with a revised data mixture (\cref{app:impl,app:training_data}).
Early 1.3B checkpoints appear only in the conditioning study of \cref{tab:conditioning_1p3b}, not as main results.
The submission also probed target-view depth with a separately trained head on intermediate VAE decoder features; this version omits that probe and evaluates depth reconstructed from generated RGB videos (\cref{tab:independent_depth}).

\paragraph{Evaluation.}
Cameras of generated videos are now recovered with VGGT-$\Omega$ instead of ViPE (\cref{app:davis_eval}).
Video quality on DAVIS is measured with VBench; FVD is reported on Syn4D, where target-view references exist.
New evaluations cover Vista4D's $110$-trajectory benchmark (\cref{app:vista4d_native}), Syn4D with exact target RGB and geometry (\cref{app:robustness}), per-scene optimization with ViDAR (\cref{tab:vidar_comparison}), alternative source reconstructors used without retraining, noise injected into the trajectories (\cref{tab:scaffold_analysis_main,tab:track_corruption}), and runtime (\cref{app:efficiency}).
Several of these were requested by reviewers and first reported in the author response.

\paragraph{Robot videos.}
\Cref{sec:robot} is new.
It shows zero-shot robot-video reshooting with \nickname-14B, long-video generation using a continuation model trained partly on robot videos (\cref{app:impl}), and Fast-WAM policy learning initialized from a 5B \nickname variant (\cref{app:fastwam}).

\paragraph{Presentation.}
The abstract, introduction, and method are rewritten for trajectory-based conditioning, and all figures are redrawn.

\paragraph{Submitted visual results.}
For reference, \cref{fig:sub_davis,fig:sub_iphone,fig:sub_realworld} reproduce the qualitative results of the submitted version: the frames are unchanged and are laid out in the format of the current figures.
They come from the submitted checkpoints with rendered RGB-D and track conditioning; the 14B DAVIS and iPhone results used multiframe softsplat rendering.
The corresponding results of the final model are in \cref{fig:qualitative,fig:iphone_nvs,fig:supp_vis,fig:baseline_comparison}.

\section{Implementation Details}
\label{app:impl}
\begin{figure*}[t]
\centering
\includegraphics[width=\textwidth]{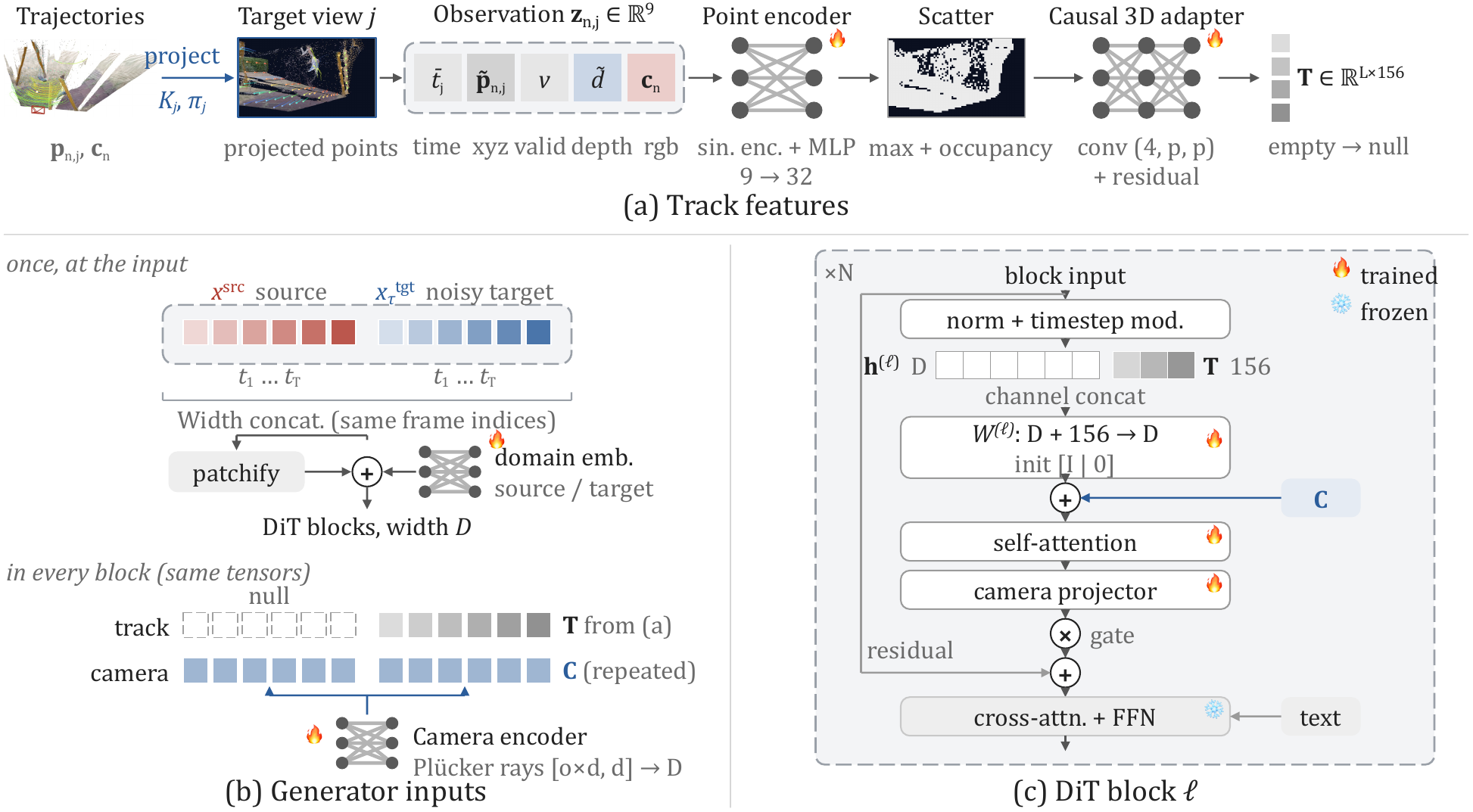}
\caption{\textbf{Trajectory conditioning.} (a) Each trajectory is projected into target frame $j$; its observation $\mathbf z_{n,j}$ (frame time, normalized position, validity, target-camera depth, and source color) is encoded, scattered to the target grid with a channel-wise maximum and an occupancy channel, and brought to the DiT token grid by a causal 3D adapter, giving $\mathbf T$. (b) Source and noisy target latents are concatenated along width once, at the input; in every block, $\mathbf T$ occupies the target half (null on the source half) and the camera features $\mathbf C$ from one shared encoder are repeated on both halves. (c) Each of the $40$ DiT blocks concatenates $\mathbf T$ with the normalized hidden features, projects back to width $D$ with $W^{(\ell)}$ initialized as $[I\,|\,0]$, and adds $\mathbf C$ before self-attention; the camera projector is identity-initialized and gated. Flames and snowflakes mark trained and frozen modules (network icons: dark = trained, light = frozen); token shading runs from the first to the last frame. Standard cross-attention/FFN residuals and the auxiliary query-frame input are omitted.}
\label{fig:track_condition_detail}
\end{figure*}

\paragraph{GMOS preprocessing.}
For the DAVIS two-stage backend, GMOS~\cite{xie26gmos:} separates query-frame background and foreground identities. Background points are fixed at their query-frame 3D positions and kept valid; foreground points retain their depth-lifted trajectories and validity. Both enter the same track encoder, without additional mask channels or a separate foreground/background branch.

\paragraph{Source--target alignment.}
Following the dynamic point-map notation of V-DPM~\cite{sucar26v-dpm:}, $P_i^{\mathrm{src}}(t_j,\pi_0^{\mathrm{src}})(u)$ denotes the position at video time $t_j$ of the point observed at pixel $u$ in source frame $i$, expressed in the first source-camera frame. For point $n$, queried at pixel $u_n$ in source frame $q_n$, the main text uses $\mathbf p_{n,j}=P_{q_n}^{\mathrm{src}}(t_j,\pi_0^{\mathrm{src}})(u_n)$ and $\mathbf c_n=I_{q_n}^{\mathrm{src}}(u_n)$.
Source and target latents are concatenated along width, so the two views retain the same temporal indices. The source half receives null track features; projected track features guide the target half. Source cameras define the common geometric frame, while target cameras determine projection and the Pl\"ucker-ray condition. No separate source-camera rays are encoded.

\paragraph{Backbone.}
We initialize from Wan2.1-T2V-14B~\cite{wan25wan:}, which has $40$ DiT blocks and a $16$-channel video VAE with spatial compression $8$ and temporal compression $4$. We fine-tune self-attention and the new conditioning modules, while keeping cross-attention and feed-forward layers frozen. The new modules comprise the source--target domain embedding, query-frame indicator embedding, track encoder and injectors, and camera adapter and projectors.

\paragraph{Track encoding.}
For point $n$ in video frame $j$, the input attributes are
\begin{equation}
\mathbf z_{n,j}=\left[\bar t_j;\;\widetilde{\mathbf p}_{n,j};\;v_{n,j};\;\widetilde d_{n,j};\;\mathbf c_n\right]\in\mathbb R^9.
\label{eq:track_payload}
\end{equation}
The frame index is normalized as $\bar t_j=j/\max(N-1,1)$. Tildes denote normalized position and depth: each position coordinate is scaled using its per-sample minimum and maximum over all point identities and frames, then zeroed for invalid observations. Depth is scaled by its per-sample maximum.
Validity combines the upstream track mask with valid projection into the target image; it is not an exact target-view occlusion test. Depth is the point's target-camera $z$ coordinate, rather than a separate depth-image condition. Source color is sampled at the query frame and held fixed along the track. We discard identities invalid in every target frame, then subsample or pad to $M=HW$ identities, with padded rows inactive. At $384\times672$, this gives $M=258{,}048$.

A sinusoidal encoding and MLP map observations to $32$-dimensional features. Target-view projection determines where to scatter these features, with channel-wise maximum aggregation and an occupancy indicator. Temporal patchification repeats the first frame to respect the VAE's causal layout, then applies a 3D convolution with kernel and stride $(4,p,p)$, where $p$ is the spatial downsampling factor, followed by one causal residual block and projection to $156$ channels. The resulting features match the DiT token grid. Empty tokens use a learned, zero-initialized embedding. The source half of the width-concatenated latent sequence receives null track tokens.

\paragraph{Camera conditioning.}
One shared camera encoder produces the target-camera features used in every DiT block. These features are repeated along width to match both source and target tokens. This repetition does not introduce source-camera rays; explicit source-camera rays would be compatible with this design but are untested. Each block has its own track concat--projection and camera output projector.
Camera rays use the Pl\"ucker convention $[\mathbf{o}\times\mathbf{d},\mathbf{d}]$. Pixel unshuffle and a 2D convolution patchify the ray map; a zero-initialized pointwise 3D convolution maps its six channels to the DiT width. Within each block, track concatenation and projection follow normalization and timestep modulation. The camera feature is then added before self-attention, whose output passes through an identity-initialized camera projector and the gated residual connection. Cross-attention and the feed-forward sublayer follow.

\paragraph{Query-frame indicator.}
A query-frame indicator identifies the source frames used to initialize trajectories. A binary sequence is one at the selected query frame and zero elsewhere; each value is broadcast over the image and repeated across three channels for encoding with the video VAE. The resulting control latents occupy the source half of the width-concatenated input, with zeros on the target half. A separate, zero-initialized Conv3D patch embedding maps these control latents to the hidden width, and its output is added to the video patch embeddings before the DiT blocks. This input addition uses neither the per-block track concat--projection nor a concatenation with the RGB latent channels.

\paragraph{Training.}
Stage~1 uses $49$ frames at $144\times256$ for $15{,}000$ steps, with effective batch size $8$ and learning rate $1{\times}10^{-5}$. Stage~2 continues the same checkpoint to step $19{,}500$ at $384\times672$, with effective batch size $32$ and learning rate $2\times10^{-5}$ on $32$ GPUs using FSDP~\cite{zhao23fsdp:}. Both stages run on NVIDIA GH200 GPUs for about $24$ hours each. Both stages optimize RGB flow matching. No geometric-conditioning dropout is used; text-conditioning dropout is $0.1$.

\paragraph{Long-video rollout.}
We extend generation beyond the $49$-frame training window by following Vista4D's window-by-window inference~\cite{lin26vista4d:}.
(i)~Split the source into $49$-frame windows with one-frame overlap: $k$ windows cover $49+48(k-1)$ frames. Pad the last window by repeating its final source frame.
(ii)~Reconstruct each window independently using the backends specified below, querying trajectories at its first frame; identities are not shared across windows. Register each reconstruction to the first window through the shared boundary frame, holding its source camera fixed while fitting a scale to the shared point map. Use the corresponding segment of the requested target-camera path.
(iii)~Generate the first window with \nickname-14B (step $19{,}500$), conditioned on the source video, tracks, and cameras.
(iv)~Generate later windows with an image-to-video continuation model using the same conditions plus the previous window's last generated frame. This shared boundary image is never a source or ground-truth frame. Remove duplicate boundary frames and padding when stitching.

The continuation model uses a Wan2.1-I2V-14B~\cite{wan25wan:} backbone and the stage-1 \nickname-14B track and camera modules. Its image condition enters through the first-frame VAE latent and CLIP image feature. During training, this is the first ground-truth target frame, dropped with probability $0.3$ and noised at level $0.05$. Training proceeds at $144\times256$ from step $15{,}000$ to $19{,}000$, then at $384\times672$ to step $22{,}500$, with effective batch size $32$ and learning rate $2\times10^{-5}$. The mixture is MultiCamVideo ($22.5\%$), Syn4D ($30\%$), DL3DV ($15\%$), MVS-Synth ($7.5\%$), and OmniWorld-DROID~\cite{zhou26omniworld:,khazatsky24droid:} ($25\%$); unlike \nickname-14B, this model has seen robot videos.
Window~1 uses 4RC for DROID and the two-stage backend for humanoid clips; later windows use 4RC without GMOS. Each 4RC reconstruction uses all $49$ source frames. All windows use $50$ denoising steps, CFG $5$, shift $5$, and seed $42$ (\cref{fig:long_rollout,fig:long_rollout_more}).

\paragraph{Condition augmentation.}
Stage~2 perturbs geometric conditions with probability $0.25$. The configured track-offset and temporal-jitter standard deviations are $0.005$ and $0.002$; image-coordinate noise ratio is $0.10$. Frame-depth jitter, depth offset, and depth-scale perturbation use $0.20$, $0.002$, and $0.003$, respectively. RGB perturbations use $0.003/0.006$, and mask augmentation probability is $0.10$. Sparse-track sampling is enabled with probability $0.5$, with strides $2$, $4$, or $8$; multi-query sampling uses two or three query frames with probability $0.5$.

\FloatBarrier
\section{Training Data}
\label{app:training_data}
\Cref{tab:training_data} reports sampling weights for both training stages. The first stage uses a broad mixture of synthetic and real, static and dynamic scenes; the second concentrates on four datasets at higher resolution. Geometry supplies conditions for RGB generation, not auxiliary prediction targets. Syn4D's \texttt{flying\_group} and MVS-Synth sequence \texttt{0000} are excluded from both stages.
Weights are rounded to two decimal places; category totals are computed before rounding.

\begin{table}[t]
\centering
\captionsetup{font=small}
\caption{\textbf{Training data.} Sampling weights (\%) per stage, not proportions of distinct videos. Static scenes need no tracker: backprojected source points stay fixed in 3D. Dynamic pairs come from synchronized multiview videos; static pairs from different chunks of the same video. DA3: DepthAnything~3~\cite{lin26depth:}.}%
\label{tab:training_data}
\small
\renewcommand{\arraystretch}{1.05}
\setlength{\tabcolsep}{3pt}
\begin{tabular*}{\linewidth}{@{\extracolsep{\fill}}c|lccll|rr@{}}
\toprule
\textbf{Index} & \textbf{Dataset} & \textbf{Real / Syn.} & \textbf{Dyn. / Static} & \textbf{Depth, cameras} & \textbf{3D tracks} & \textbf{Stage 1} & \textbf{Stage 2} \\
\midrule
1 & Syn4D~\cite{jiang26syn4d:} & Synthetic & Dynamic & Ground truth & Ground truth & 31.65 & 40.00 \\
2 & MultiCamVideo~\cite{bai25recammaster:} & Synthetic & Dynamic & VGGT-$\Omega$ & CoWTracker & 0.00 & 30.00 \\
3 & DL3DV~\cite{ling24dl3dv-10k:} & Real & Static & DA3 + dataset & Fixed points & 32.91 & 20.00 \\
4 & MVS-Synth~\cite{huang18deepmvs:} & Synthetic & Static & Ground truth & Fixed points & 6.33 & 10.00 \\
5 & Kubric~\cite{greff22kubric:} & Synthetic & Dynamic & Ground truth & Box motion & 12.66 & 0.00 \\
6 & TartanAir~\cite{wang20tartanair:} & Synthetic & Static & Ground truth & Fixed points & 12.66 & 0.00 \\
7 & RE10K~\cite{zhou18stereo:} & Real & Static & DA3 + dataset & Fixed points & 3.80 & 0.00 \\
\bottomrule
\end{tabular*}
\end{table}

Stage~1 combines Syn4D and two equally weighted Kubric subsets for dynamic scenes, together with real static scenes from DL3DV and RE10K and synthetic static scenes from MVS-Synth and TartanAir. Each Kubric subset contributes approximately $6.33\%$ of the stage-1 sampling weight. Stage~2 uses Syn4D and MultiCamVideo for dynamic scenes and DL3DV and MVS-Synth for static scenes, increasing the dynamic share from $44.30\%$ to $70\%$.

\Cref{tab:training_data} lists the geometry and track source of each dataset (VGGT-$\Omega$~\cite{wang26vggt-omega:}, CoWTracker~\cite{lai26cowtracker:}, DepthAnything~3~\cite{lin26depth:}). The $30\%$ second-stage MultiCamVideo weight comprises $21\%$ dense and $9\%$ sparse query-frame data.

For static scenes, we form source--target pairs from different chunks of the same monocular video, similar to 4Real-Video-V2~\cite{wang26fused:}; temporal synchronization is unnecessary since the scene is static. Dynamic training pairs come from synchronized multiview videos; we do not use dynamic monocular videos for training.

\section{Camera-Adherence Evaluation on DAVIS}
\label{app:davis_eval}
\begin{wraptable}[9]{r}{0.45\textwidth}
\vspace{-\intextsep}
\centering
\captionsetup{font=footnotesize,skip=3pt}
\caption{\textbf{Backbones and native resolutions.}}%
\label{tab:baseline_settings}
\fontsize{7.5}{9}\selectfont
\renewcommand{\arraystretch}{1.0}
\setlength{\tabcolsep}{2pt}
\begin{tabular*}{\linewidth}{@{\extracolsep{\fill}}lll@{}}
\toprule
\textbf{Method} & \textbf{Backbone} & \textbf{Res.} \\
\midrule
ReCamMaster & Wan2.1-T2V-1.3B & 480$\times$832 \\
TrajectoryCrafter & CogVideoX-5B & 384$\times$672 \\
EX-4D & Wan2.1-I2V-14B & 512$\times$512 \\
Gen3C & Cosmos-7B & 704$\times$1280 \\
Vista4D & Wan2.1-T2V-14B & 384$\times$672 \\
\rowcolor{gray!10}[0pt][0pt]\oursrowshade \nickname-14B & Wan2.1-T2V-14B & 384$\times$672 \\
\bottomrule
\end{tabular*}
\vspace{-0.5\baselineskip}
\end{wraptable}

We construct our DAVIS evaluation set by selecting $50$ videos~\cite{pont-tuset17davis:} and randomly sampling two target trajectories per video from the ReCamMaster~\cite{bai25recammaster:} and Vista4D~\cite{lin26vista4d:} trajectory pools. Baselines use released checkpoints and native generation resolutions; Vista4D uses 4RC-derived source geometry for this DAVIS comparison. We evaluate the first $49$ output frames. Camera recovery with VGGT-$\Omega$ and Pi3 uses $384\times672$ frames; VBench uses the same resolution at $16$\,FPS with each clip's caption. The main \nickname result uses 4RC source depth, cameras, and dense tracks, with $50$ denoising steps, CFG $5$, shift $5$, and seed $42$.

\begin{table}[!b]
\centering
\captionsetup{font=small}
\caption{\textbf{VBench across reconstruction backends on DAVIS.} $\alpha$: 4RC; $\beta$: two-stage backend with GMOS. 5B rows use Wan2.2-TI2V-5B at $704\times1280$. Columns follow \cref{tab:video_quality}.}%
\label{tab:vbench_registered}
\small
\renewcommand{\arraystretch}{1.0}
\setlength{\tabcolsep}{3pt}
\begin{tabular*}{\linewidth}{@{\extracolsep{\fill}}lcccccc@{}}
\toprule
\textbf{Method} & \textbf{Steps}
& \textbf{Aest.} $\uparrow$
& \textbf{Imag.} $\uparrow$
& \textbf{Subj.} $\uparrow$
& \textbf{Backg.} $\uparrow$
& \textbf{Temp.} $\uparrow$ \\
\midrule
Vista4D$^\alpha$ & 50
& 0.478
& 0.534
& \underline{0.904}
& 0.926
& \textbf{0.256} \\

\nickname-14B$^\alpha$ & 50
& \underline{0.493}
& \textbf{0.601}
& \textbf{0.910}
& \textbf{0.931}
& \underline{0.253} \\

\nickname-14B$^\beta$ & 50
& 0.478
& \underline{0.589}
& 0.891
& 0.919
& 0.252 \\

\nickname-14B$^\beta$ & 25
& 0.471
& 0.580
& 0.883
& 0.910
& 0.249 \\
\midrule
\nickname-5B$^\alpha$ & 50
& \textbf{0.501}
& 0.559
& 0.894
& \underline{0.928}
& 0.253 \\

\nickname-5B$^\beta$ & 50
& 0.489
& 0.554
& 0.881
& 0.920
& 0.252 \\
\bottomrule
\end{tabular*}
\end{table}

VGGT-$\Omega$ recovers cameras from generated videos. Translation metrics use the $54$ translation-bearing trajectories; rotation uses all $100$. DAVIS has no captured target-view reference for these paths, so we report camera control and VBench scores. All methods are evaluated with the same VGGT-$\Omega$ evaluator revision on all $100$ videos; baseline videos are the stored outputs of their released models, not resampled.
For ATE and RPE-T, we anchor both trajectories at their first camera and rescale the predicted camera centers to match the centered Frobenius norm of the requested centers. We average per-video RMSEs; relative errors use adjacent frames. No subsequent global rotation alignment is applied to ATE, so orientation errors after the first frame contribute to it. Its magnitude also depends on the requested trajectories and is not directly comparable across DAVIS and Vista4D trajectory sets.

\begin{table}[!t]
\centering
\captionsetup{font=small}
\caption{\textbf{Source geometry and sampling steps on DAVIS.} $\alpha$: 4RC; $\beta$: two-stage backend with GMOS. 5B rows use Wan2.2-TI2V-5B at $704\times1280$ ($384\times672$ for $\dagger$). $^\dagger$Trained without Syn4D, MVS-Synth, and DL3DV, with a dense track condition without persistent identities. ATE is scaled by $10^2$; RPE-T and TransErr$^2$ by $10^3$. Rotations are in degrees.}%
\label{tab:davis_backend}
\renewcommand{\arraystretch}{1.0}
\setlength{\tabcolsep}{3pt}
\small
\begin{tabular*}{\linewidth}{@{\extracolsep{\fill}}lcccccc@{}}
\toprule
\multirow{2}{*}{\textbf{Method}} & \multirow{2}{*}{\textbf{Steps}}
& \multicolumn{3}{c}{\textbf{VGGT-$\Omega$}}
& \multicolumn{2}{c}{\textbf{Pi3}} \\
\cmidrule(lr){3-5}\cmidrule(lr){6-7}
& & \textbf{ATE} $\downarrow$ & \textbf{RPE-T} $\downarrow$ & \textbf{RPE-R} $\downarrow$
& \textbf{TransErr$^2$} $\downarrow$ & \textbf{RotErr} $\downarrow$ \\
\midrule
Vista4D$^\alpha$ & 50
& 3.576
& 3.608
& 0.132
& 6.201
& \underline{2.070} \\

\nickname-14B$^\alpha$ & 50
& 1.657
& 2.415
& 0.102
& \underline{3.206}
& 2.332 \\

\nickname-14B$^\beta$ & 50
& \underline{0.719}
& \underline{1.299}
& \underline{0.101}
& 3.359
& 2.284 \\

\nickname-14B$^\beta$ & 25
& 0.729
& \textbf{1.095}
& \textbf{0.077}
& 3.625
& \textbf{1.785} \\
\midrule
\nickname-5B$^\alpha$ & 50
& 1.464
& 2.952
& 0.116
& \textbf{3.195}
& 2.275 \\
\nickname-5B$^\beta$ & 50
& \textbf{0.608}
& 1.302
& 0.105
& 3.425
& 2.199 \\
\nickname-5B$^{\beta,\dagger}$ & 25
& 0.914
& 1.363
& 0.106
& 3.604
& 2.270 \\
\bottomrule
\end{tabular*}
\end{table}

\noindent
\Cref{tab:davis_backend} evaluates the same checkpoint with the two-stage backend, where GMOS static/dynamic fusion fixes background identities in 3D. At $50$ steps, this condition lowers ATE from $1.657\times10^{-2}$ to $0.719\times10^{-2}$, while its VBench scores are lower (\cref{tab:vbench_registered}). With this condition, $25$ denoising steps give a lower rotation error than $50$ ($0.077^\circ$ versus $0.101^\circ$). The 4RC condition uses its reconstructed trajectories directly, without GMOS preprocessing.
\noindent A 5B model with a dense identity-target track condition (Wan2.2-TI2V-5B, generated at $704\times1280$ with $50$ steps) behaves like the 14B model: with 4RC it has ATE $1.464\times10^{-2}$ and rotation error $0.116^\circ$, and the two-stage backend with GMOS lowers these to $0.608\times10^{-2}$ and $0.105^\circ$ (\cref{tab:davis_backend}).
\noindent The \nickname-5B$^{\dagger}$ row in \cref{tab:davis_backend} is a 5B variant trained on OmniWorld-DROID, MultiCamVideo, TartanAir, and Kubric, excluding Syn4D, MVS-Synth, and DL3DV. With the same source condition and $25$ steps, its errors (ATE $0.914\times10^{-2}$, rotation $0.106^\circ$) remain below those of Vista4D. It uses step $29{,}000$ and a dense track condition without persistent identities. It also differs from \nickname-14B in backbone and track condition, so this comparison is not a controlled data ablation.
\noindent The Pi3 columns use source-aligned camera recovery. All \nickname variants have lower Pi3 translation error than Vista4D; only the 14B variant with VGGT-$\Omega$ at $25$ steps also has lower Pi3 rotation error.

\paragraph{Compatibility with a smaller backbone.}
The geometric interface also applies to Wan2.2-TI2V-5B with a variant of the track condition (\cref{tab:davis_backend}).
At $704\times1280$ with $50$ denoising steps on DAVIS, replacing 4RC with the two-stage backend and GMOS reduces ATE from $1.464\times10^{-2}$ to $0.608\times10^{-2}$, and on the nine-clip Syn4D comparison its source-aligned rotation error rises from $0.515^\circ$ with ground-truth geometry to $0.885^\circ$ with 4RC estimates (\cref{tab:rotation_attribution}).
Both trends match the 14B model.

\paragraph{Sampling time of the 5B model.} On one H200 with $50$ denoising steps and $49$ frames, \nickname-5B generates a $704\times1280$ video in $202.0$\,s on average, versus $417.4$\,s for \nickname-14B at $384\times672$ and $495.0$\,s for Vista4D at $384\times672$ (plus $6.2$\,s of point-cloud rendering), averaged over the same $100$ Vista4D trajectory pairs; peak GPU memory is $76.3$, $43.2$, and $49.9$\,GiB, respectively. For these two configurations, the measured sampling time of the 5B model is about $2.1\times$ lower than that of the 14B model; because backbone, resolution, and track condition all differ, this is not a speedup under matched conditions. These are complete sampling times with precomputed source reconstruction from the $100$-pair evaluation runs, not the matched two-trajectory profiling in \cref{tab:vista4d_runtime}, so the 14B totals differ between the two measurements.
\paragraph{GMOS within the same reconstruction backend.}
The shared trajectory representation supports foreground/background fusion without modifying the generator: for the two-stage backend, background points are held fixed in 3D while foreground points keep their tracked motion, and a foreground mask decides which is which. Our default 4RC condition uses no explicit fusion.
\begin{table}[tb]
\centering
\begin{minipage}{\linewidth}
\captionsetup{font=small}
\caption{\textbf{Segmentation for static/dynamic track fusion on DAVIS.} Two-stage backend with the same first-frame tracks, $384\times672$, $50$ steps; only the foreground mask differs. ATE: $\times10^2$; RPE-T and TransErr$^2$: $\times10^3$; rotations: degrees.}
\label{tab:gmos_ablation}
\renewcommand{\arraystretch}{1.0}
\small
\setlength{\tabcolsep}{6pt}
\begin{tabular*}{\linewidth}{@{\extracolsep{\fill}}lccccc@{}}
\toprule
\textbf{Mask} & \textbf{ATE} $\downarrow$ & \textbf{RPE-T} $\downarrow$ & \textbf{RPE-R} $\downarrow$ & \textbf{TransErr$^2$} $\downarrow$ & \textbf{RotErr} $\downarrow$ \\
\midrule
Motion-derived & 0.984 & 1.527 & 0.114 & \textbf{3.351} & 2.364 \\
\rowcolor{gray!10}[0pt][0pt]\oursrowshade GMOS & \textbf{0.719} & \textbf{1.299} & \textbf{0.101} & 3.359 & \textbf{2.284} \\
\bottomrule
\end{tabular*}
\end{minipage}
\end{table}

\begin{figure}[t]
\centering
\includegraphics[width=\textwidth]{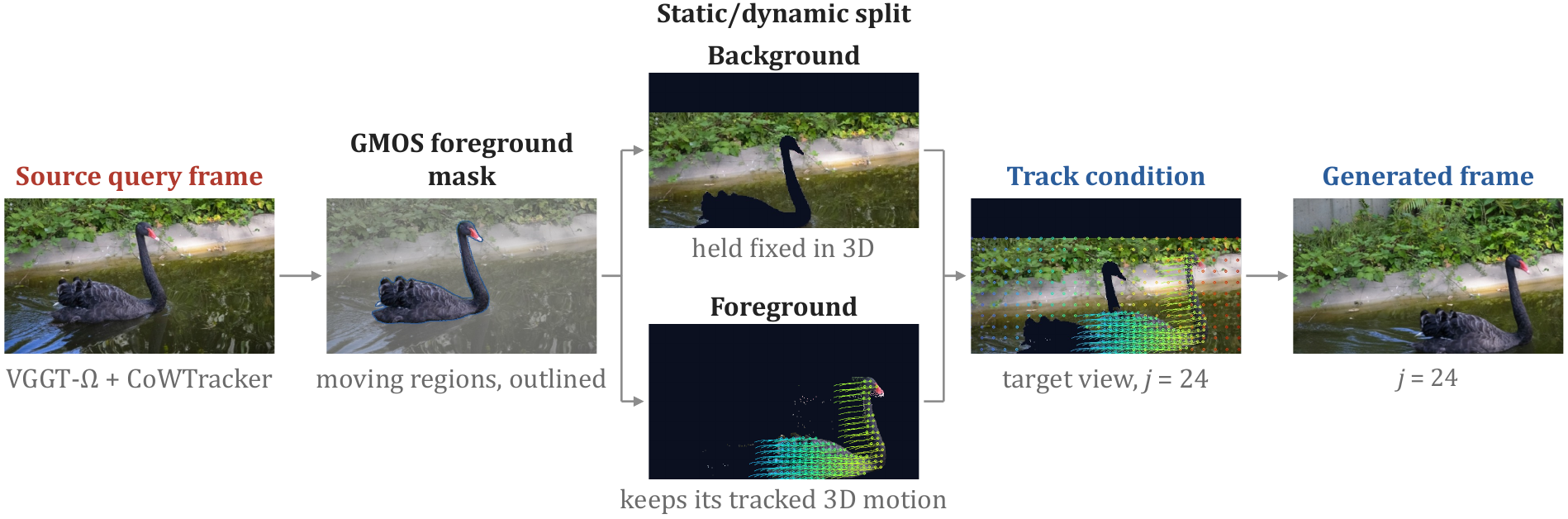}
\caption{\textbf{GMOS in the two-stage backend.} VGGT-$\Omega$ supplies depth and cameras and CoWTracker supplies tracks. GMOS segments the moving regions in the query frame; trajectories outside the mask are held at their query-frame 3D positions, and trajectories inside keep their tracked motion. Both sets form one track condition, shown projected into the target view at $j=24$. Background regions occluded by the foreground in the query frame have no trajectories. DAVIS \emph{blackswan}, final checkpoint, $50$ denoising steps.}
\label{fig:gmos_fusion}
\end{figure}

\Cref{tab:gmos_ablation} compares two foreground masks within the same static/dynamic fusion of the two-stage backend: GMOS masks and masks derived from track motion. Both rows use the step-$19{,}500$ checkpoint, the same VGGT-$\Omega$ and CoWTracker caches with first-frame queries, $50$ denoising steps at $384\times672$, CFG $5$, shift $5$, and seed $42$; the saved sampling settings are otherwise identical. In both rows, background identities are fixed at their query-frame positions and foreground identities keep their tracked motion, so the comparison isolates the segmentation, not fusion versus no fusion.
GMOS lowers the VGGT-$\Omega$ errors (ATE by $27\%$, RPE-T by $15\%$, RPE-R by $12\%$) and the Pi3 rotation error by $3\%$, while the Pi3 translation error is essentially unchanged. The GMOS row is the $50$-step VGGT-$\Omega$ row of \cref{tab:davis_backend}. \Cref{fig:gmos_fusion} illustrates the fusion on one DAVIS sequence.

\section{Camera Adherence on Vista4D Trajectories}
\label{app:vista4d_native}
We additionally evaluate on the $110$ native Vista4D trajectories with both generated-only VGGT-$\Omega$ recovery and Vista4D's source-aligned Pi3 protocol (\cref{tab:vista4d_native110}).
Vista4D uses the benchmark's released Pi3 reconstruction, masks, and rendered inputs, whereas \nickname uses 4RC or the two-stage backend.
This preserves Vista4D's native preprocessing and compares complete pipelines; it does not isolate the video generator as the source of performance differences.
For the qualitative comparisons in \cref{fig:qualitative,fig:baseline_comparison}, baselines use their released sampling settings and the benchmark's released Pi3 geometry where a point cloud is required; Vista4D videos are its official outputs.
The VGGT-$\Omega$ and Pi3 evaluators recover cameras for scoring and are separate from the source reconstruction used for generation.
With 4RC source geometry, \nickname has lower RPE-T under VGGT-$\Omega$ and lower translation error under Pi3, but higher rotation error under both evaluators.
VGGT-$\Omega$ evaluation follows the normalization in \cref{app:davis_eval}. ATE and RPE-T have $110$ valid values for \nickname with 4RC and $109$ each for Vista4D and the two-stage condition; RPE-R uses all $110$. These are per-method valid-metric counts, not a common-subset evaluation. Scale normalization requires non-degenerate requested and recovered camera paths; a completed video can therefore lack a valid translation metric. Vista4D's Pi3 protocol reconstructs source and generated frames jointly, then aligns the source cameras by $\mathrm{Sim}(3)$.

The 4RC condition in \cref{tab:vista4d_native110} uses its reconstructed trajectories without mask-guided fusion and aligns the geometry to the released camera coordinate system with a per-video scalar depth factor. An alternative pointmap $\mathrm{Sim}(3)$ alignment of the same checkpoint and 4RC geometry reduces VGGT-$\Omega$ RPE-R from $1.142^\circ$ to $0.348^\circ$, but increases Pi3 RotErr from $14.98^\circ$ to $17.19^\circ$. Thus the source-condition comparison includes alignment effects, and the two evaluators disagree on this preprocessing change.

\begin{table}[!t]
\centering
\begin{minipage}{\linewidth}
\captionsetup{font=small}
\caption{\textbf{Camera control with GT and estimated source geometry on Syn4D.} Nine held-out clips; Pi3 camera recovery. ATE: $\times10^2$; RPE-T and TransErr$^2$: $\times10^3$; rotations: degrees. Bold/underline: best/second best per geometry group. Protocols: \cref{app:vista4d_native}.}%
\label{tab:rotation_attribution}
\small
\renewcommand{\arraystretch}{1.0}
\setlength{\tabcolsep}{3pt}
\begin{tabular*}{\linewidth}{@{\extracolsep{\fill}}llccccc@{}}
\toprule
\textbf{Method} & \textbf{Geometry}
& \textbf{ATE} $\downarrow$
& \textbf{RPE-T} $\downarrow$
& \textbf{RPE-R} $\downarrow$
& \textbf{TransErr$^2$} $\downarrow$
& \textbf{RotErr} $\downarrow$ \\
\midrule
Vista4D & GT
& 2.233
& 12.12
& \underline{0.079}
& 13.36
& \underline{0.372} \\
\rowcolor{gray!10}[0pt][0pt]\oursrowshade \nickname-14B & GT
& \textbf{1.299}
& \textbf{8.121}
& \textbf{0.051}
& \textbf{1.881}
& \textbf{0.242} \\
\rowcolor{gray!10}[0pt][0pt]\oursrowshade \nickname-5B & GT & \underline{1.957} & \underline{11.19} & 0.080 & \underline{5.023} & 0.515 \\
\midrule
Vista4D & Pi3
& 2.348
& \underline{10.15}
& \underline{0.067}
& 19.26
& \underline{0.651} \\
\rowcolor{gray!10}[0pt][0pt]\oursrowshade \nickname-14B & Pi3 + GT 2D
& \textbf{1.330}
& \textbf{8.158}
& \textbf{0.052}
& \textbf{5.715}
& \textbf{0.534} \\
\rowcolor{gray!10}[0pt][0pt]\oursrowshade \nickname-5B & Pi3 + GT 2D & \underline{1.996} & 10.40 & 0.084 & \underline{13.50} & 0.740 \\
\midrule
\rowcolor{gray!10}[0pt][0pt]\oursrowshade \nickname-14B & 4RC
& \textbf{1.203}
& \textbf{8.280}
& \textbf{0.055}
& \textbf{24.15}
& \underline{0.931} \\
\rowcolor{gray!10}[0pt][0pt]\oursrowshade \nickname-5B & 4RC & \underline{2.263} & \underline{12.63} & \underline{0.087} & \underline{34.46} & \textbf{0.885} \\
\bottomrule
\end{tabular*}
\end{minipage}
\end{table}

\paragraph{Additional geometry comparisons.}
\Cref{sec:ablation} analyzes the ground-truth and 4RC comparisons in \cref{tab:rotation_attribution}.
All rows use nine held-out Syn4D \texttt{flying\_group} clips, with source camera $0$ and target camera $1$.
Pi3~\cite{wang26pi:} recovers cameras jointly from source and generated frames.
In the additional Pi3-source group, \nickname constructs tracks from estimated depth and cameras using ground-truth 2D correspondences and visibility, whereas Vista4D uses its native Pi3 reconstruction.
This partially oracle condition is distinct from both the full ground-truth and 4RC conditions.
The 5B model is Wan2.2-TI2V-5B with a dense identity-target track condition, generated at $384\times672$; its source-aligned rotation error with Pi3 geometry is $0.740^\circ$.

On Vista4D's native trajectories, source replacement changes camera estimates, geometry, and coverage together.
Separating generator sensitivity from reconstruction error would require running both methods with the same estimated geometry on these trajectories.

\paragraph{Conditioning signals in early checkpoints.}
To ask whether richer geometric conditions repair rotation, we compare four separately trained 1.3B checkpoints that add rendered RGB-D, tracks, and multi-frame rendering to Pl\"ucker camera rays (\cref{tab:conditioning_1p3b}). The Pi3 translation error decreases monotonically from $0.987$ to $0.685$ as conditions are added, whereas rotation error does not decrease; rendered RGB-D raises it by $2.33^\circ$ (95\% confidence interval $[0.18^\circ, 4.73^\circ]$ over the $20$ pairs). These early checkpoints barely follow the requested rotation, so the experiment shows only that added geometric conditions improve translation more readily than rotation; it does not measure the rotation accuracy of the final model.
All four Wan2.1-T2V-1.3B checkpoints are evaluated at step $3{,}000$, using the same 4RC reconstructions, $288\times512\times49$ outputs, and $50$ denoising steps. The $20$ pairs come from ten sources, five selected for rotation failures, with metrics as in \cref{tab:vista4d_native110}.
\begin{table}[t]
\centering
\captionsetup{font=small}
\caption{\textbf{Conditioning in early 1.3B checkpoints.} Rotation gain is the fitted ratio of generated to requested rotation. ATE is scaled by $10^2$; RPE-T by $10^3$. Rotations are in degrees.}%
\label{tab:conditioning_1p3b}
\renewcommand{\arraystretch}{1.0}
\small
\setlength{\tabcolsep}{3pt}
\begin{tabular*}{\linewidth}{@{\extracolsep{\fill}}lcccccc@{}}
\toprule
\multirow{2}{*}{\textbf{Condition}}
& \multicolumn{3}{c}{\textbf{VGGT-$\Omega$}}
& \multicolumn{2}{c}{\textbf{Pi3, source-aligned}}
& \multirow{2}{*}{\textbf{Rot.\ gain} $\uparrow$} \\
\cmidrule(lr){2-4}\cmidrule(lr){5-6}
& \textbf{ATE} $\downarrow$ & \textbf{RPE-T} $\downarrow$ & \textbf{RPE-R} $\downarrow$
& \textbf{TransErr$^2$} $\downarrow$ & \textbf{RotErr} $\downarrow$ & \\
\midrule
Pl\"ucker only & \textbf{15.03} & 60.92 & \textbf{1.423} & 0.9866 & \textbf{31.26} & \textbf{0.0218} \\
+ rendered RGB-D & 16.21 & \underline{41.69} & \underline{1.565} & 0.7625 & 33.59 & 0.0027 \\
+ rendered RGB-D + tracks & \underline{16.14} & 45.22 & 1.642 & \underline{0.7117} & \underline{33.25} & 0.0096 \\
+ multi-frame RGB-D + tracks & 16.39 & \textbf{34.69} & 1.635 & \textbf{0.6854} & 33.29 & \underline{0.0149} \\
\bottomrule
\end{tabular*}
\end{table}

\section{Novel-View Evaluation on iPhone/DyCheck}
\label{app:iphone_eval}
We use the five iPhone/DyCheck scenes and source--target split of Vista4D~\cite{lin26vista4d:}. One synchronized camera supplies the source video and a held-out camera supplies the target trajectory and reference RGB. Target references and masks are center-cropped to $16{:}9$ and resized to $288\times512$. Predictions are resized to the same canvas and evaluated over $49$ frames; EX-4D's square outputs are resized directly, without an additional center crop. Masked metrics use the intersection of valid and co-visible pixels, not a foreground-object mask. The two \nickname rows share the same checkpoint and use $50$ denoising steps, CFG $5$, shift $5$, and seed $42$. For both source conditions, source depth, cameras, and tracks are calibrated per scene to the source iPhone LiDAR scale.

EX-4D generates at its native $512{\times}512$ resolution with $25$ denoising steps and uses the dataset's source-view depth. The VGGT-$\Omega$ source condition uses CoWTracker tracks. Neither iPhone source condition uses GMOS.
All RGB rows, including EX-4D, are rescored from stored outputs with one evaluator. The iPhone columns of \cref{tab:independent_depth} report depth reconstructed by 4RC or VGGT-$\Omega$ from both the generated and the held-out target RGB; the target reconstruction is a pseudo-reference, not sensor ground truth. All iPhone depth rows, baselines included, use the same scorer, crop, valid co-visible mask, and confidence threshold, and are equal-weight means over the five scenes.

\section{Conditioning Comparisons}
\label{app:conditioning}
\paragraph{Stage-1 design study.}
The variants in \cref{tab:stage1_design} share the backbone initialization, training data, optimizer, and $15{,}000$ steps at $144\times256$. We evaluate them on DAVIS at $288\times512$ with 4RC source geometry, $25$ denoising steps, and VGGT-$\Omega$ camera recovery. The rendered-feature variant encodes rendered RGB together with validity and source/target motion masks, then sums the resulting features with the track features before injection, without adding attention tokens.

\paragraph{Additional conditioning variants.}
\Cref{tab:ablation_progressive} compares camera-ray, rendered-video, and trajectory conditioning. The upper block reports variants with different training schedules and evaluation sets; the lower block compares rendered streams on a common evaluation subset. In the lower block, adding rendered streams reduces translation errors but increases rotation error. This trade-off complements the controlled stage-1 comparison in \cref{tab:stage1_design}.
All variants use a 14B backbone with source-video and camera conditioning; cameras are recovered with VGGT-$\Omega$. In the upper block, the camera-only model uses $10$k training steps, track-payload variants use $15$k, and rendered variants use a different evaluation suite. The reported translation/rotation counts are $54/100$ for the upper block and $36/63$ for the lower block, so magnitudes should not be compared across blocks.
\begin{table}[htbp]
\centering
\begin{minipage}{\linewidth}
\captionsetup{font=small}
\caption{\textbf{Track and rendered conditioning on DAVIS.} PCD denotes rendered point clouds. The unmatched upper block is not ranked; lower-block bold and underline mark \textbf{best} and \underline{second-best} results. ATE and RPE-T are scaled by $10^2$ and $10^3$; rotations are in degrees.}%
\label{tab:ablation_progressive}
\small
\renewcommand{\arraystretch}{1.0}
\setlength{\tabcolsep}{3pt}
\begin{tabular*}{\linewidth}{@{\extracolsep{\fill}}lccc@{}}
\toprule
\textbf{Condition}
& \textbf{ATE} $\downarrow$
& \textbf{RPE-T} $\downarrow$
& \textbf{RPE-R} $\downarrow$
\\
\midrule
\multicolumn{4}{@{}l}{\emph{Conditioning variants (training not matched)}} \\
Pl\"ucker rays only
& 10.65
& 7.32
& 0.373
\\

PCD(RGB-D) + valid mask, no tracks
& 10.36
& 9.16
& 0.236
\\

PCD(RGB-D) + 3D tracks
& 2.35
& 2.38
& 0.086
\\

Tracks without per-track color
& 2.33
& 2.13
& 0.094
\\

\rowcolor{gray!10}[0pt][0pt]\oursrowshade \textbf{3D tracks only} (\nickname)
& 1.62
& 2.08
& 0.065
\\
\midrule
\multicolumn{4}{@{}l}{\emph{Rendered-stream variants (common-success camera subset)}} \\
\rowcolor{gray!10}[0pt][0pt]\oursrowshade \textbf{3D tracks only} (\nickname)
& 2.04
& 3.27
& \textbf{0.059}
\\

+ PCD render stream, mask off
& \textbf{1.32}
& 2.43
& 0.076
\\

+ PCD render stream, mask on
& \underline{1.50}
& \textbf{1.96}
& 0.076
\\

PCD render streams only, no tracks
& \textbf{1.32}
& \underline{2.37}
& \underline{0.071}
\\
\bottomrule
\end{tabular*}
\end{minipage}
\end{table}

\section{Additional Novel-View Synthesis Evaluation}
\label{app:additional_nvs}
\paragraph{Rendered and trajectory conditioning.}
\Cref{tab:ablation_nvs} compares five conditioning variants on five DyCheck sequences. These checkpoints use an earlier architecture with a rendered-conditioning branch, not the model of the main paper. Both source-backend panels use $49$ frames at $288\times512$, $50$ denoising steps, CFG $1$, shift $5$, and seed $42$. Adding tracks improves on the RGB-D tier under both backends; the relative ordering of single- and multi-frame variants depends on the metric and backend. Because neither panel contains a track-only checkpoint, this study does not test removal of the rendered branch.
\begin{table}[!htbp]
\centering
\captionsetup{font=small}
\caption{\textbf{Earlier conditioning variants on DyCheck.} Source geometry: (a) 4RC; (b) two-stage backend. ``m'' denotes masked metrics; bold and underline mark \textbf{best} and \underline{second-best} results per panel.}%
\label{tab:ablation_nvs}
\renewcommand{\arraystretch}{1.05}
\small
\setlength{\tabcolsep}{3pt}

\textbf{(a) Default 4RC source geometry}\par\vspace{0.25em}
\begin{tabular*}{\linewidth}{@{\extracolsep{\fill}}lcccccc@{}}
\toprule
\textbf{Configuration}
& \textbf{mPSNR} $\uparrow$
& \textbf{mSSIM} $\uparrow$
& \textbf{mLPIPS} $\downarrow$
& \textbf{PSNR} $\uparrow$
& \textbf{SSIM} $\uparrow$
& \textbf{LPIPS} $\downarrow$ \\
\midrule
Pl\"ucker rays only          & 11.23 & 0.442 & 0.630 & 11.36 & 0.281 & 0.717 \\
$+$ RGB render               & 11.91 & 0.466 & 0.611 & 11.95 & 0.314 & 0.725 \\
$+$ depth + valid mask       & 11.93 & 0.453 & 0.640 & 11.93 & 0.296 & 0.758 \\
$+$ 3D tracks (single-frame) & \underline{13.58} & \underline{0.477} & \textbf{0.490} & \underline{13.20} & \underline{0.319} & \textbf{0.568} \\
$+$ multi-frame static cache & \textbf{13.65} & \textbf{0.479} & \underline{0.497} & \textbf{13.28} & \textbf{0.322} & \underline{0.581} \\
\bottomrule
\end{tabular*}

\vspace{0.7em}
\textbf{(b) Two-stage source geometry}\par\vspace{0.25em}
\begin{tabular*}{\linewidth}{@{\extracolsep{\fill}}lcccccc@{}}
\toprule
\textbf{Configuration}
& \textbf{mPSNR} $\uparrow$
& \textbf{mSSIM} $\uparrow$
& \textbf{mLPIPS} $\downarrow$
& \textbf{PSNR} $\uparrow$
& \textbf{SSIM} $\uparrow$
& \textbf{LPIPS} $\downarrow$ \\
\midrule
Pl\"ucker rays only          & 11.25 & 0.443 & 0.631 & 11.36 & 0.282 & 0.720 \\
$+$ RGB render               & 11.92 & 0.466 & 0.613 & 11.94 & 0.314 & 0.726 \\
$+$ depth + valid mask       & 12.02 & 0.454 & 0.641 & 11.97 & 0.296 & 0.759 \\
$+$ 3D tracks (single-frame) & \textbf{13.81} & \underline{0.480} & \textbf{0.482} & \textbf{13.40} & \underline{0.325} & \textbf{0.554} \\
$+$ multi-frame static cache & \underline{13.31} & \textbf{0.485} & \underline{0.510} & \underline{13.18} & \textbf{0.330} & \underline{0.592} \\
\bottomrule
\end{tabular*}
\vspace{-0.5em}
\end{table}

\begin{table}[!t]
\centering
\captionsetup{font=small}
\caption{\textbf{Novel-view synthesis compared with ViDAR.} Under Vista4D's iPhone protocol, ViDAR is optimized per scene on the first $49$ source frames, center-cropped and resized to $288\times512$. We evaluate $49$ target frames, unlike ViDAR's full-sequence benchmark. $\alpha$: 4RC; $\beta$: two-stage backend. ``m'' denotes masked metrics. Bold and underline mark \textbf{best} and \underline{second-best} results.}%
\label{tab:vidar_comparison}
\small
\renewcommand{\arraystretch}{1.0}
\setlength{\tabcolsep}{3pt}
\begin{tabular*}{\linewidth}{@{\extracolsep{\fill}}lcccccc@{}}
\toprule
\textbf{Method}
& \textbf{PSNR} $\uparrow$
& \textbf{SSIM} $\uparrow$
& \textbf{LPIPS} $\downarrow$
& \textbf{mPSNR} $\uparrow$
& \textbf{mSSIM} $\uparrow$
& \textbf{mLPIPS} $\downarrow$ \\
\midrule
ViDAR~\cite{nazarczuk25vidar:}
& \underline{12.90}
& \textbf{0.385}
& 0.734
& 13.01
& \textbf{0.525}
& 0.653 \\
\rowcolor{gray!10}[0pt][0pt]\oursrowshade \nickname-14B\smash{$^{\alpha}$}
& 12.68
& 0.326
& \underline{0.509}
& \underline{13.22}
& 0.484
& \underline{0.435} \\
\rowcolor{gray!10}[0pt][0pt]\oursrowshade \nickname-14B\smash{$^{\beta}$}
& \textbf{13.20}
& \underline{0.348}
& \textbf{0.465}
& \textbf{14.03}
& \underline{0.507}
& \textbf{0.379} \\
\bottomrule
\end{tabular*}
\end{table}

\paragraph{Optimization-based reconstruction.}
\Cref{tab:vidar_comparison} compares with ViDAR~\cite{nazarczuk25vidar:} on the same five scenes. ViDAR is optimized per scene using only the first $49$ source frames, with outputs evaluated at $288\times512$. With the two-stage backend, \nickname has better PSNR and LPIPS, while ViDAR has better SSIM; the same ordering holds for masked metrics. ViDAR takes on average $5$ h $18$ min per scene on an A6000. This distinguishes scene-specific optimization from amortized generation, but is not a hardware-matched speed comparison.

Each scene is optimized once, and the first $49$ target frames are evaluated after center cropping and resizing to $288\times512$. This short-window comparison has less temporal and viewpoint coverage than ViDAR's full-sequence benchmark. We do not compare with CAT4D~\cite{wu25cat4d:}, whose code and checkpoints were not released at the time of our experiments.

\section{Synthetic Evaluation and Upstream Robustness}
\label{app:robustness}
\paragraph{Ground-truth source geometry.}
The Syn4D evaluation with ground-truth source geometry uses five held-out source clips and $32$ source--target pairs with exact target RGB and depth. Rendered-geometry baselines use ground-truth source depth and cameras to reproject source RGB; \nickname uses ground-truth 3D tracks, and ReCamMaster uses source RGB and cameras. No method receives target RGB, depth, or masks. Target-view image metrics compare directly against rendered target RGB. Syn4D also contributes training data for \nickname, whereas baseline checkpoints are not adapted to it. The held-out real-capture results provide a complementary test.
\nickname uses $25$ denoising steps at $384\times672$ for the ground-truth-source comparison and the reconstruction-backend and track-corruption studies (\cref{tab:syn4d_gt_comparison,tab:upstream_sensitivity,tab:track_corruption}).
Cameras are recovered with VGGT-$\Omega$ in both Syn4D comparisons. For these comparisons and the scaffold studies, ATE is the mean of per-pair trajectory RMSEs after $\mathrm{Sim}(3)$ alignment, including rotation, translation, and scale. This differs from the first-frame-anchored, scale-normalized ATE used on DAVIS and the native Vista4D trajectories (\cref{app:davis_eval,app:vista4d_native}); ATE values should therefore be compared within each evaluation setting. Image metrics and FVD use ground-truth target videos as references, including the upstream-replacement and track-corruption studies on the same $32$ ground-truth-source pairs. FVD in \cref{tab:syn4d_gt_comparison,tab:syn4d_4rc_nvs} uses the same I3D scorer at $288\times512\times49$; the two tables still evaluate different pairs, so their values should not be compared with each other. The FVD column of \cref{tab:upstream_sensitivity} comes from an earlier scoring at $256\times256$ and is comparable only within that table.

\begin{figure}[t]
\centering
\includegraphics[width=\textwidth]{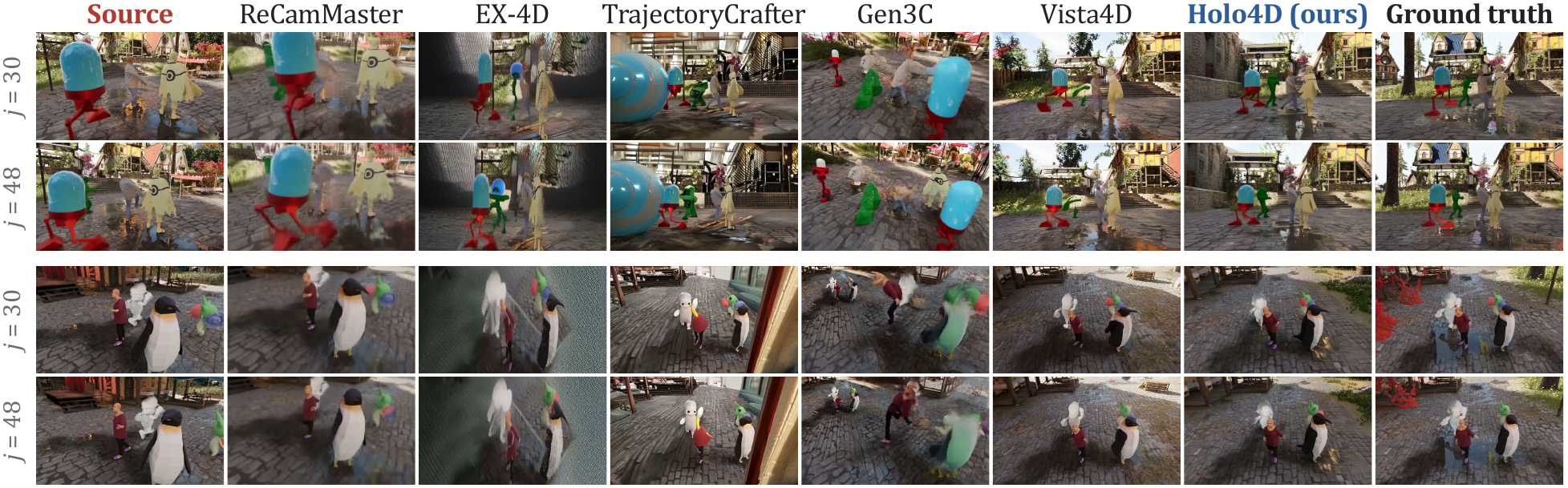}
\caption{\textbf{Syn4D with ground-truth source geometry.} Two held-out pairs with large viewpoint changes; the right column is the rendered target view. Rendered-geometry baselines reproject ground-truth source depth, \nickname receives ground-truth 3D trajectories, and ReCamMaster uses only the cameras (\cref{tab:syn4d_gt_comparison}). EX-4D generates square frames, shown as a central crop. \nickname places the moving objects as in the ground truth; all methods differ from it in unobserved regions.}
\label{fig:syn4d_qual}
\end{figure}

\Cref{fig:syn4d_qual} shows two of these pairs.

\paragraph{FVD reference videos.}
Using source videos as FVD references, as in Vista4D~\cite{lin26vista4d:}, measures distributional similarity to the inputs and does not establish adherence to the requested target views.
A model that returns the source videos unchanged would obtain zero FVD, up to numerical precision under identical preprocessing, even when the requested cameras differ.
We therefore report FVD only on Syn4D against ground-truth target videos.
FVD remains a distribution-level video-quality metric; camera adherence is evaluated separately using recovered camera trajectories. We assess video quality with VBench (\cref{tab:video_quality}) and target-referenced FVD; CLIP-based scores are not reported.

\paragraph{Estimated source geometry.}
\Cref{tab:syn4d_4rc_nvs} uses $32$ pairs from $16$ held-out \texttt{flying\_group} scenes, with camera pairs $0\!\rightarrow\!7$ and $2\!\rightarrow\!6$. Geometry-consuming methods receive 4RC source estimates. \nickname uses $50$ denoising steps at $384\times672$ and has the lowest camera errors and LPIPS; ReCamMaster has the highest PSNR and SSIM, while Vista4D has the lowest FVD. FVD compares all methods against the same $32$ target videos at $288\times512$ over $49$ frames; the small sample size makes it a supplementary distributional measure. This test set differs from the ground-truth-source experiment, so differences in their means do not isolate reconstruction error. All methods use the same neutral prompt.
\begin{table}[!htbp]
\centering
\captionsetup{font=small}
\caption{\textbf{Novel-view synthesis on Syn4D with 4RC source geometry.} Geometry-conditioned methods receive 4RC source geometry in their native form; ReCamMaster uses no geometry. FVD uses the paired target videos. ATE and RPE-T are scaled by $10^2$ and $10^3$; rotations are in degrees. Bold/underline: best/second best.}%
\label{tab:syn4d_4rc_nvs}
\renewcommand{\arraystretch}{1.0}
\small
\setlength{\tabcolsep}{3pt}
\begin{tabular*}{\linewidth}{@{\extracolsep{\fill}}lccccccc@{}}
\toprule
\textbf{Method}
& \textbf{PSNR} $\uparrow$
& \textbf{SSIM} $\uparrow$
& \textbf{LPIPS} $\downarrow$
& \textbf{ATE} $\downarrow$
& \textbf{RPE-T} $\downarrow$
& \textbf{RPE-R} $\downarrow$
& \textbf{FVD} $\downarrow$ \\
\midrule
ReCamMaster
& \textbf{12.40}
& \textbf{0.277}
& 0.761
& 11.69
& 63.79
& 0.776
& 1908 \\
TrajectoryCrafter
& 9.418
& 0.176
& 0.725
& 9.636
& \underline{39.98}
& 0.522
& 3073 \\
EX-4D
& 9.865
& 0.163
& 0.863
& 18.76
& 89.83
& 2.438
& 2625 \\
Gen3C
& \underline{11.72}
& \underline{0.216}
& 0.686
& \underline{9.336}
& 52.28
& \underline{0.395}
& \underline{1693} \\
Vista4D
& 10.98
& 0.158
& \underline{0.668}
& 9.972
& 41.56
& 0.419
& \textbf{1525} \\
\midrule
\rowcolor{gray!10}[0pt][0pt]\oursrowshade \nickname-14B
& 11.70
& 0.180
& \textbf{0.667}
& \textbf{3.267}
& \textbf{32.09}
& \textbf{0.272}
& 1711 \\
\bottomrule
\end{tabular*}
\end{table}

\paragraph{Output depth.}
On the Syn4D test with ground-truth source geometry, depth reconstructed from generated RGB with either VGGT-$\Omega$ or 4RC is compared against exact target depth. On iPhone/DyCheck, each reconstructor is applied to both generated and held-out target RGB; the target reconstruction serves as a pseudo-depth reference. For both Syn4D depth evaluators, each $49$-frame prediction is aligned to exact target depth by one robust median scale over the full video, with no per-frame alignment. Metrics are weighted by the number of valid pixels when aggregated over source--target pairs. Valid pixels have finite positive predicted and reference depths and pass the evaluator's confidence threshold at the fifth percentile.
In \cref{tab:independent_depth}, columns identify the output-depth evaluator; the iPhone \nickname row labels identify the source reconstructor used for generation. Both source conditions are evaluated with both output reconstructors. We define $\delta_1$ as the fraction of pixels with $\max(\hat d/d,d/\hat d)<1.25$.
\begin{table}[!t]
\centering
\captionsetup{font=small}
\caption{\textbf{Depth reconstructed from generated videos.} Syn4D uses exact target depth; iPhone uses reconstructed target-video depth as a pseudo-reference. $\alpha$: 4RC; $\beta$: two-stage source backend, without GMOS. Evaluation and aggregation are specified in \cref{app:iphone_eval,app:robustness}.}%
\label{tab:independent_depth}
\label{tab:iphone_depth}
\small
\renewcommand{\arraystretch}{1.0}
\setlength{\tabcolsep}{3pt}
\begin{tabular*}{\linewidth}{@{\extracolsep{\fill}}lcccccccc@{}}
\toprule
& \multicolumn{4}{c}{\textbf{Syn4D (exact depth)}} & \multicolumn{4}{c}{\textbf{iPhone (pseudo-reference)}} \\
\cmidrule(lr){2-5}\cmidrule(lr){6-9}
& \multicolumn{2}{c}{\textbf{VGGT-$\Omega$}} & \multicolumn{2}{c}{\textbf{4RC}} & \multicolumn{2}{c}{\textbf{VGGT-$\Omega$}} & \multicolumn{2}{c}{\textbf{4RC}} \\
\cmidrule(lr){2-3}\cmidrule(lr){4-5}\cmidrule(lr){6-7}\cmidrule(lr){8-9}
\textbf{Method} & \textbf{AbsRel} $\downarrow$ & \textbf{$\delta_1$} $\uparrow$ & \textbf{AbsRel} $\downarrow$ & \textbf{$\delta_1$} $\uparrow$ & \textbf{AbsRel} $\downarrow$ & \textbf{$\delta_1$} $\uparrow$ & \textbf{AbsRel} $\downarrow$ & \textbf{$\delta_1$} $\uparrow$ \\
\midrule
ReCamMaster & 0.762 & 0.311 & 0.738 & 0.298 & 0.152 & 0.765 & 0.121 & 0.835 \\
TrajectoryCrafter & 0.673 & 0.367 & 0.713 & 0.337 & 0.109 & 0.848 & 0.098 & 0.881 \\
EX-4D & 0.673 & 0.298 & 0.766 & 0.232 & 0.109 & 0.823 & 0.112 & 0.877 \\
Gen3C & 0.740 & 0.337 & 0.736 & 0.326 & 0.098 & \underline{0.862} & 0.112 & 0.895 \\
Vista4D & \underline{0.628} & \underline{0.509} & \underline{0.655} & \underline{0.455} & \underline{0.092} & 0.845 & 0.081 & 0.888 \\
\midrule
\nickname-14B (GT source) & \textbf{0.423} & \textbf{0.683} & \textbf{0.413} & \textbf{0.635} & -- & -- & -- & -- \\
\nickname-14B$^\alpha$ & -- & -- & -- & -- & 0.094 & 0.843 & \underline{0.077} & \underline{0.895} \\
\nickname-14B$^\beta$ & -- & -- & -- & -- & \textbf{0.076} & \textbf{0.914} & \textbf{0.064} & \textbf{0.933} \\
\bottomrule
\end{tabular*}
\end{table}

\paragraph{Reconstruction backends.}
We hold the trained checkpoint, target trajectories, sampler, and seed fixed and replace the source backend with 4RC, V-DPM~\cite{sucar26v-dpm:}, VGGT-$\Omega$ + CoWTracker, or VGGT-$\Omega$ + Open-D4RT~\cite{zhang26efficiently:}. In the last two cases, VGGT-$\Omega$ supplies cameras and depth and the tracker supplies point trajectories; these Syn4D conditions use no GMOS fusion. \Cref{tab:upstream_sensitivity} shows that the shared interface accepts all four backends without retraining, with different quality trade-offs.
\begin{table}[!t]
\centering
\captionsetup{font=small}
\caption{\textbf{Source geometry and track noise on Syn4D.} The same generator receives estimated source geometry from four reconstruction backends (top; no GMOS), ground-truth tracks (middle), or ground-truth tracks with Gaussian noise whose standard deviation is $\alpha$ times that of the source tracks (bottom). $32$ held-out pairs, $25$ denoising steps. ATE and RPE-T are scaled by $10^2$ and $10^3$; rotations are in degrees. FVD uses an earlier $256\times256$ scoring and is comparable only within this table. Bold/underline rank estimated backends.}%
\label{tab:upstream_sensitivity}
\label{tab:track_corruption}
\small
\renewcommand{\arraystretch}{1.0}
\setlength{\tabcolsep}{3pt}
\begin{tabular*}{\linewidth}{@{\extracolsep{\fill}}lccccccc@{}}
\toprule
\textbf{Source geometry}
& \textbf{ATE} $\downarrow$
& \textbf{RPE-T} $\downarrow$
& \textbf{RPE-R} $\downarrow$
& \textbf{PSNR} $\uparrow$
& \textbf{SSIM} $\uparrow$
& \textbf{LPIPS} $\downarrow$
& \textbf{FVD} $\downarrow$ \\
\midrule
VGGT-$\Omega$ + Open-D4RT
& 1.858
& 11.22
& 0.151
& 13.10
& 0.158
& 0.169
& 1337 \\
V-DPM
& 1.660
& 8.904
& \textbf{0.079}
& 14.96
& 0.268
& 0.151
& 1252 \\
4RC
& \underline{1.353}
& \textbf{7.148}
& \underline{0.087}
& \underline{14.98}
& \underline{0.278}
& \textbf{0.131}
& \textbf{1168} \\
VGGT-$\Omega$ + CoWTracker
& \textbf{1.205}
& \underline{7.688}
& 0.107
& \textbf{15.09}
& \textbf{0.294}
& \underline{0.137}
& \underline{1216} \\
\midrule
Ground-truth tracks
& 1.065
& 6.154
& 0.073
& 16.52
& 0.441
& 0.119
& 1098 \\
\midrule
GT tracks, $\alpha=0.1$
& 3.033
& 22.53
& 0.221
& 13.44
& 0.171
& 0.172
& 1375 \\
GT tracks, $\alpha=0.5$
& 2.568
& 33.71
& 0.540
& 13.49
& 0.111
& 0.238
& 1739 \\
GT tracks, $\alpha=1.0$
& 2.228
& 31.88
& 0.574
& 13.37
& 0.100
& 0.241
& 1838 \\
GT tracks, $\alpha=4.0$
& 2.163
& 33.02
& 0.604
& 13.26
& 0.093
& 0.242
& 1783 \\
\bottomrule
\end{tabular*}
\end{table}

\begin{figure}[t]
\centering
\includegraphics[width=\textwidth]{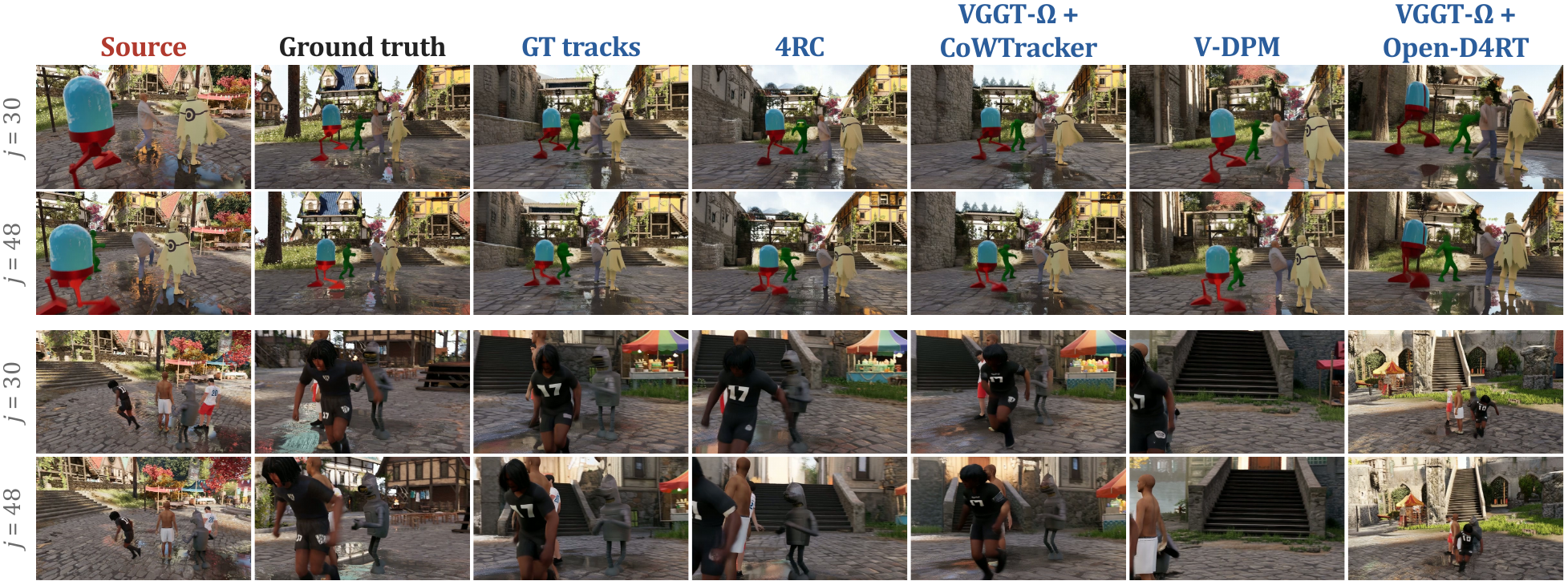}
\caption{\textbf{Changing the source reconstructor without retraining.} The same checkpoint receives ground-truth trajectories or trajectories from four reconstruction backends, with target cameras, sampler, and seed fixed (\cref{tab:upstream_sensitivity}). Geometric errors change the generated layout and appearance, as seen with V-DPM and Open-D4RT.}
\label{fig:backend_swap}
\end{figure}

\Cref{fig:backend_swap} shows two pairs under each backend.

\paragraph{Track corruption.}
We add Gaussian noise to ground-truth tracks while keeping the other inputs fixed. Per-axis noise standard deviation is the source-track standard deviation multiplied by $\alpha$. Rotation error, SSIM, and LPIPS degrade as corruption increases (\cref{tab:track_corruption}). ATE jumps at $\alpha=0.1$ but does not grow further under independent $\mathrm{Sim}(3)$ trajectory alignment, so it should not be used alone to characterize this sensitivity.

\section{Conditioning Cost, Runtime, and Memory}
\label{app:efficiency}

Source geometry and query-frame appearance are reusable across target trajectories. Target projection, depth, validity, scattering, and camera features must be recomputed. Training uses offline geometry annotations but still evaluates the learned adapters.

\paragraph{Channel-wise geometric conditioning.}
For hidden features $\mathbf h\in\mathbb R^{L\times D}$ and track features $\mathbf T\in\mathbb R^{L\times156}$, each injector projects $[\mathbf h;\mathbf T]\in\mathbb R^{L\times(D+156)}$ back to $\mathbb R^{L\times D}$.
Self-attention therefore keeps the same sequence length, avoiding Vista4D's additional rendered-video tokens~\cite{lin26vista4d:}.
Source-video tokens, adapters, and per-block projections still incur cost; fewer tokens do not imply a proportional reduction in total memory.

\begin{table}[htb]
\centering
\captionsetup{font=small}
\caption{\textbf{Warm inference runtime on one H200.} Times are mean $\pm$ standard deviation over four videos, including reconstruction, proxy rendering, and generation but excluding model loading and video writing.}%
\label{tab:runtime_breakdown}
\renewcommand{\arraystretch}{1.05}
\small
\setlength{\tabcolsep}{2pt}
\begin{tabular*}{\linewidth}{@{\extracolsep{\fill}}lcc@{}}
\toprule
\textbf{Stage} & \textbf{Time (s)} & \textbf{Share} \\
\midrule
4RC feed-forward       & $1.46 \pm 0.02$  & $1.5\%$ \\
Static cache + rendering & $7.54 \pm 0.30$  & $7.8\%$ \\
Dynamic render + fusion & $1.43 \pm 0.31$  & $1.5\%$ \\
3D-track encoding                     & $0.06 \pm 0.00$  & $0.1\%$ \\
Denoising + decoding                          & $86.50 \pm 0.26$ & $89.2\%$ \\
\midrule
\textbf{Warm-pipeline total} & $\mathbf{96.99 \pm 0.67}$ & $\mathbf{100\%}$ \\
\bottomrule
\end{tabular*}
\end{table}

\noindent\textbf{H200 stage-wise profile.}
\Cref{tab:runtime_breakdown} profiles step $19{,}500$ on one H200 NVL at $288\times512\times49$, with $25$ steps and CFG $5$.
Four videos after one warm-up average $96.99\pm0.67$ s, excluding model loading and MP4 writing.
The $1.46$ s 4RC entry is fresh reconstruction timed by its internal profiler, not a cache read.
The harness also performs proxy rendering ($8.97$\,s for the static cache and dynamic render rows), although the generator uses only track conditioning.
Preprocessing totals $10.49$ s; denoising and VAE decoding take $86.50$ s.

\begin{table}[htbp]
\centering
\begin{minipage}[t]{0.485\linewidth}
\captionsetup{font=footnotesize,skip=3pt,justification=raggedright,singlelinecheck=false}
\caption{\textbf{Reconstruction cost per $49$-frame source.} Warm/cold exclude/include model loading. Open-D4RT supplies tracks only.}
\label{tab:backend_cost}
\footnotesize
\renewcommand{\arraystretch}{1.0}
\setlength{\tabcolsep}{2pt}
\begin{tabular*}{\linewidth}{@{\extracolsep{\fill}}lcccc@{}}
\toprule
\textbf{Backend} & \textbf{Mode} & \textbf{GPU} & \textbf{Time (s)} & \textbf{GiB} \\
\midrule
4RC & warm & A6000 & $8.8$ & $16.0$ \\
VGGT-$\Omega$ + CoWTracker & warm & A6000 & $11.8$ & $25.2$ \\
\addlinespace[2pt]
VGGT-$\Omega$ + CoWTracker & cold & A6000 & $66.6$ & $25.1$ \\
Open-D4RT & cold & A6000 & $168.4$ & $5.9$ \\
V-DPM & cold & H100 & $139.4$ & $53.5$ \\
\bottomrule
\end{tabular*}

\end{minipage}\hfill
\begin{minipage}[t]{0.485\linewidth}
\captionsetup{font=footnotesize,skip=3pt,justification=raggedright,singlelinecheck=false}
\caption{\textbf{Sampling cost on one H200.} Both 14B models use $384\times672\times49$ and $50$ denoising steps. $N=13{,}104$ tokens per video.}
\label{tab:vista4d_runtime}
\footnotesize
\renewcommand{\arraystretch}{1.24}
\setlength{\tabcolsep}{2pt}
\begin{tabular*}{\linewidth}{@{\extracolsep{\fill}}lcc@{}}
\toprule
\textbf{Metric} & \textbf{Vista4D} & \textbf{\nickname} \\
\midrule
Tokens per DiT layer $\downarrow$ & $3N$ & $\boldsymbol{2N}$ \\
Step time (s) $\downarrow$ & $9.89$ & \textbf{7.39} \\
Total time (s) $\downarrow$ & $581.9$ & \textbf{570.9} \\
Peak memory (GiB) $\downarrow$ & $49.86$ & \textbf{49.14} \\
\bottomrule
\end{tabular*}

\end{minipage}
\end{table}

\Needspace{8\baselineskip}
\noindent\textbf{Source reconstruction cost.}
\Cref{tab:backend_cost} uses separate backend profiles.
The warm A6000 runs average ten $49$-frame ADT clips after one warm-up, excluding model loading.
Unlike the internally timed H200 stage, the $8.8$\,s 4RC measurement includes video decoding, native preprocessing, and inference.
Cold runs include loading and query dense tracks at the first and last frames.
Warm memory is PyTorch peak allocation; cold memory is \texttt{nvidia-smi} device occupancy.
The different inputs, hardware, and timing boundaries preclude a matched ranking.

\noindent\textbf{Matched sampling time.}
\Cref{tab:vista4d_runtime} compares two native Vista4D trajectories at $384\times672\times49$, using $50$ steps on one H200. Each video has $N=13{,}104$ latent tokens ($13$ latent frames of $24\times42$ patches); Vista4D attends over source, rendered, and target videos, \nickname over source and target.
Sampling-loop averages include classifier-free guidance but exclude source reconstruction and point-cloud rendering: $7.39$ s per step for \nickname versus $9.89$ s for Vista4D.
Including loading, conditioning, decoding, and writing, full-run times are $570.9$ and $581.9$ s.
The remaining stages jointly account for approximately $201$ and $87$ s; their individual contributions were not profiled.
Peak memory is similar ($49.14$ and $49.86$ GiB). Training-step cost was not measured.

\section{Additional Qualitative Results}
\label{app:qualitative}
\begin{figure}[p]
\centering
\includegraphics[width=\textwidth]{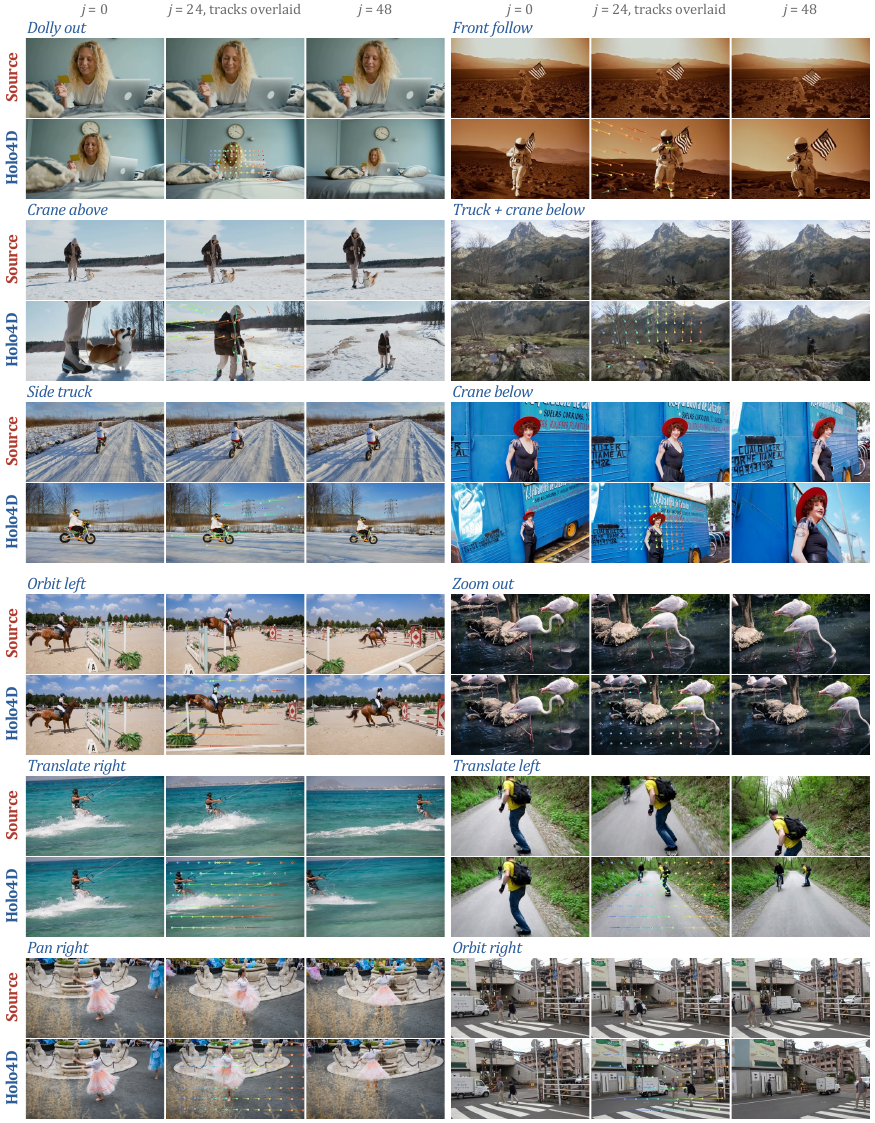}
\caption{\textbf{Additional video reshooting examples.} Source and generated frames for six trajectories from Vista4D's benchmark (top) and six from our DAVIS evaluation set (bottom), named above each example. The middle generated frame ($j=24$) overlays a subset of the projected trajectories, with trails covering the previous $12$ frames. All examples use the two-stage backend and $50$ denoising steps; none appears in the other figures.}
\label{fig:supp_vis}
\end{figure}

\Cref{fig:supp_vis} shows \nickname on six further trajectories from Vista4D's benchmark and six from DAVIS, covering lateral, orbit, crane, dolly, and zoom motions; overlaid trajectories indicate the requested image-space paths for visual comparison with the generated motion.
These examples complement the DAVIS and held-out iPhone comparisons in the main paper.

\begin{figure*}[!t]
\centering
\includegraphics[width=\textwidth]{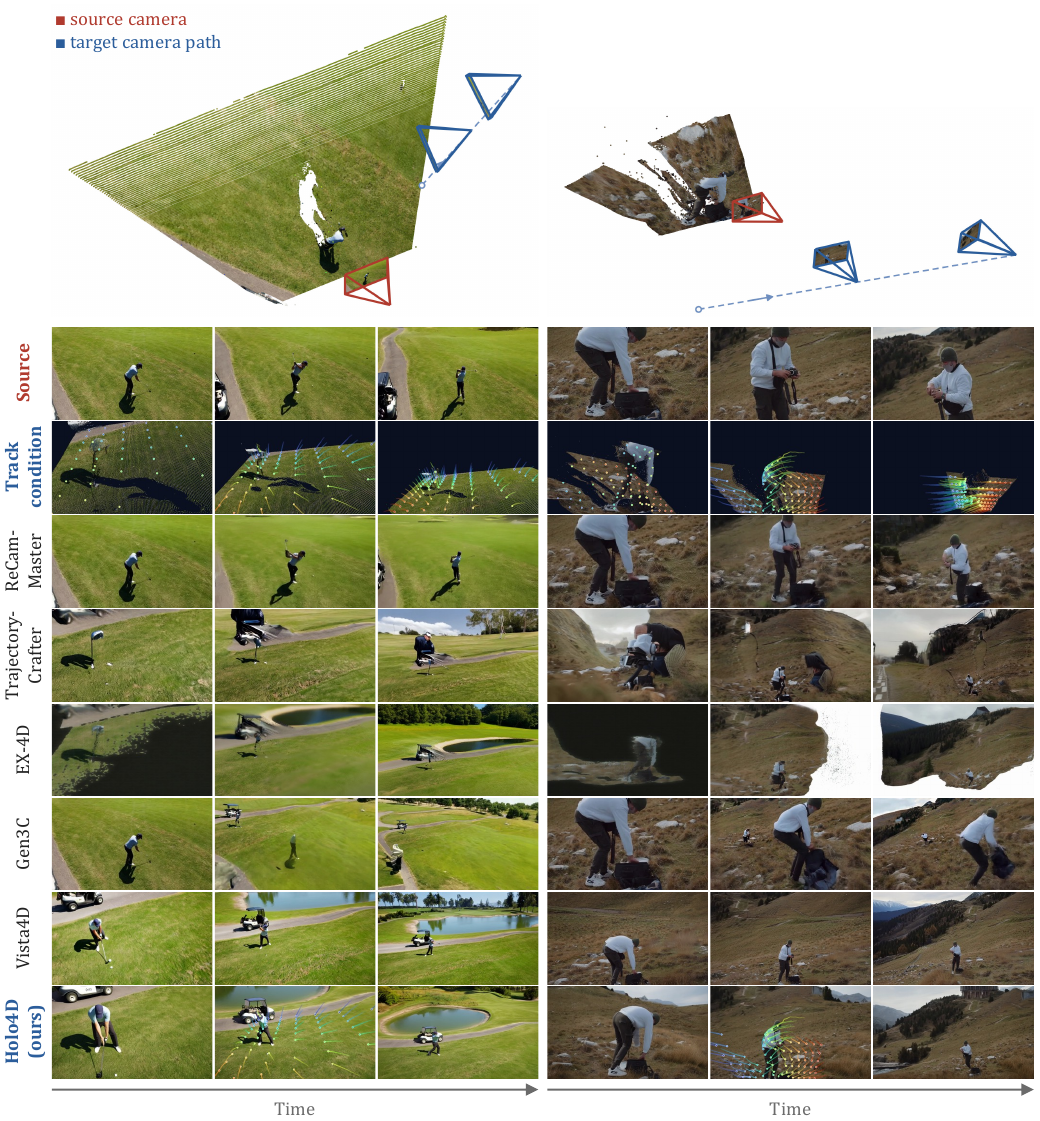}
\caption{\textbf{Comparison with all baselines on two further Vista4D trajectories.} Left: a dolly-out from a golfer that reveals a golf cart; right: a rising, zooming-out crane shot of a man on a hillside. Layout as in \cref{fig:qualitative}. TrajectoryCrafter, EX-4D, and Gen3C render the cart incomplete or distort the people, EX-4D leaves large holes, and ReCamMaster stays close to the source framing; Vista4D and \nickname both follow the requested paths. \nickname-14B uses 4RC source reconstruction and $50$ denoising steps; baselines use the same settings as in \cref{fig:qualitative}.}
\label{fig:baseline_comparison}
\end{figure*}

\Cref{fig:baseline_comparison} compares all methods on two further trajectories in the layout of \cref{fig:qualitative}.

\begin{figure}[!t]
\centering
\includegraphics[width=\textwidth]{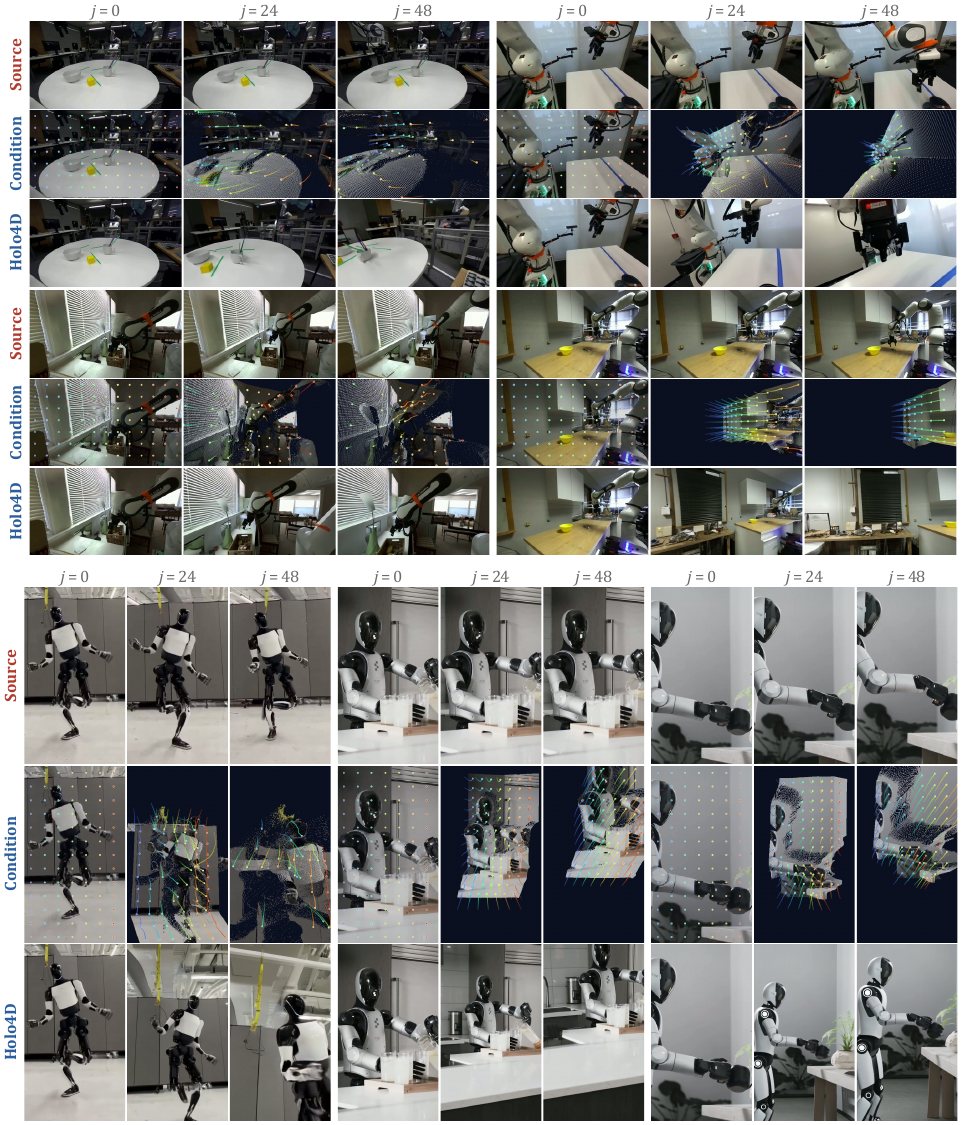}
\caption{\textbf{Zero-shot novel-view synthesis of robot videos.} Each example shows the source video (top), projected 3D trajectories with source colors (middle; dark regions have no support, and trails show a subset over the preceding $12$ frames), and the generated view (bottom). Top: DROID~\cite{khazatsky24droid:} episodes with 4RC reconstruction; bottom: humanoid clips with the two-stage backend. \nickname-14B uses no robot training data and no fine-tuning; $50$ denoising steps.}
\label{fig:robot}
\end{figure}

\paragraph{Robot videos.}
\Cref{fig:robot} applies \nickname-14B, trained without robot videos, to DROID manipulation episodes and humanoid-robot clips without fine-tuning, using a scaffold reconstructed from each source video.

\begin{figure}[p]
\centering
\includegraphics[width=\textwidth]{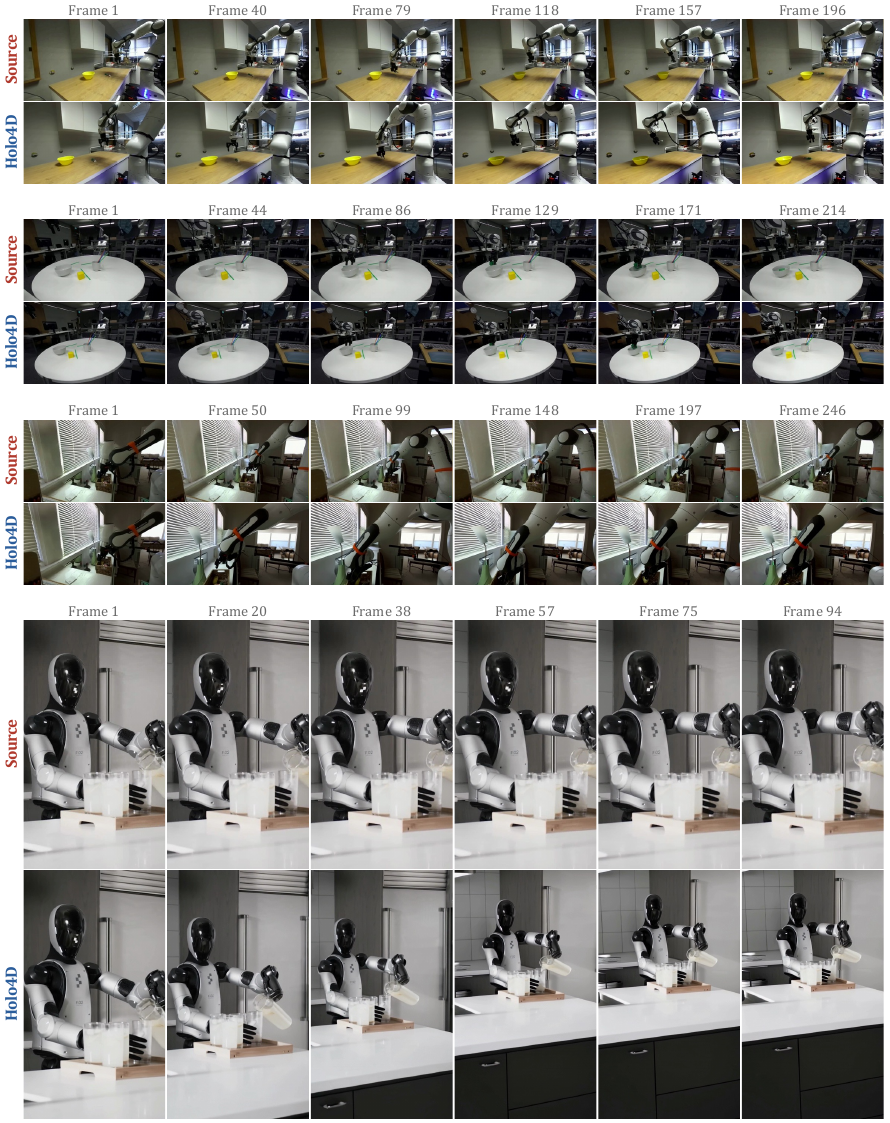}
\caption{\textbf{More long-video rollouts.} From top: three DROID~\cite{khazatsky24droid:} episodes of $196$, $214$, and $246$ frames ($5$, $5$, and $6$ windows) reshot to static target views, and a $94$-frame humanoid clip ($2$ windows) reshot along a close crane-above path. Six evenly spaced frames per video. Window~1 uses \nickname-14B with 4RC (DROID) or the two-stage backend (humanoid); later windows use the I2V continuation model, trained with OmniWorld-DROID, and 4RC; $50$ steps (\cref{app:impl}).}
\label{fig:long_rollout_more}
\end{figure}

\Cref{fig:long_rollout_more} shows four further robot-video rollouts using a continuation model trained with OmniWorld-DROID (\cref{app:impl}).

\paragraph{Remaining failure modes.}
Fine dynamic structures such as hands can remain inconsistent across views, and large disoccluded regions are completed by the generative prior without geometric support, so their appearance can be implausible. The conditioning does not remove these failures; we expect stronger base models to reduce them while keeping the same interface.

\begin{figure}[t]
\centering
\includegraphics[width=\textwidth]{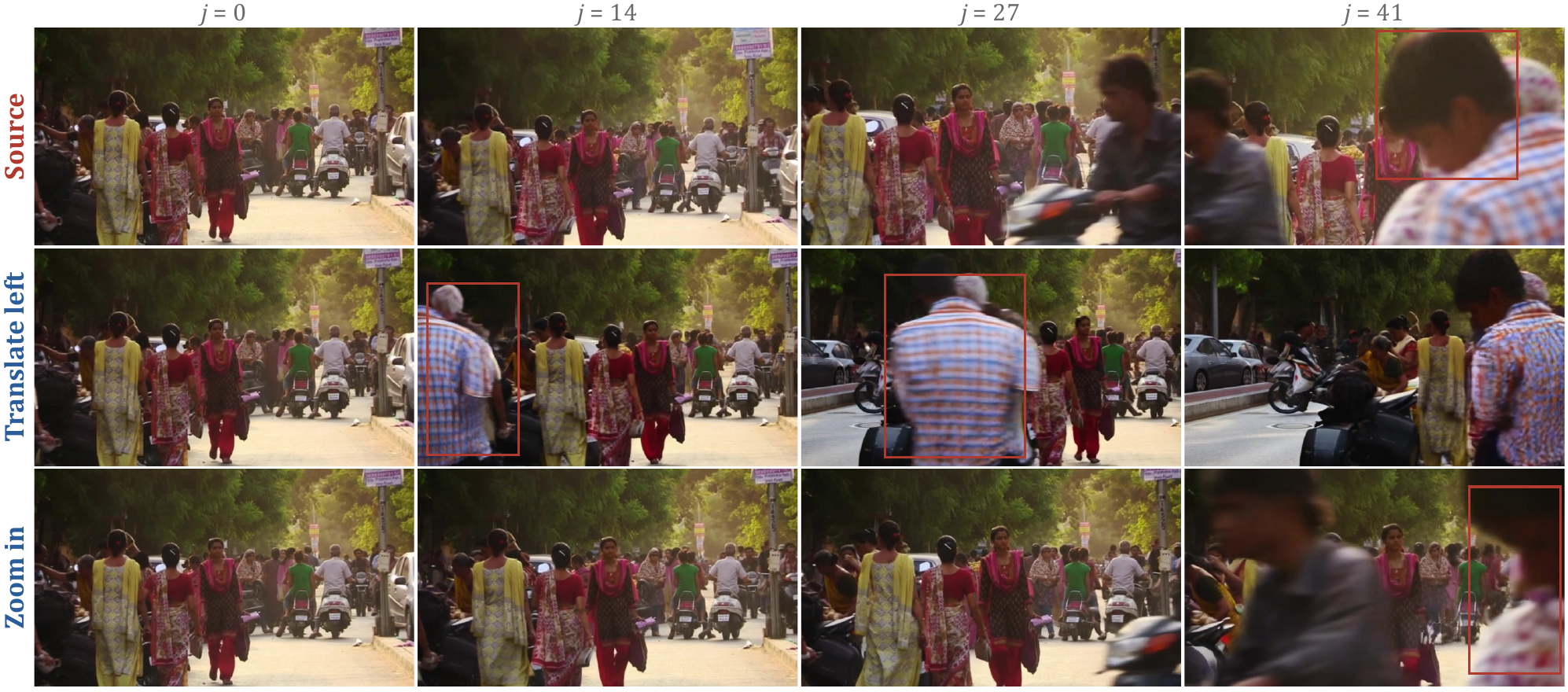}
\caption{\textbf{Content outside the initial scaffold.} The marked person enters the source view late and has no first-frame query points, so generated views must complete them without trajectory support (no target-view ground truth is available).}
\label{fig:limitation}
\end{figure}

\paragraph{Coverage of the initial scaffold.}
With first-frame queries, objects that enter the source video later may have no corresponding trajectories.
The generator still receives their appearance in the source video, but must infer their target-view placement without direct geometric support.
\Cref{fig:limitation} illustrates this case for a person entering the DAVIS \emph{india} sequence.
Because changing the camera can change when a person becomes visible, this example illustrates missing scaffold coverage rather than a measured timing error.

\section{Robot Policy Learning with Fast-WAM}
\label{app:fastwam}
Fast-WAM~\cite{yuan26fast-wam:} couples a Wan2.2-TI2V-5B video expert with an action expert. Both arms share the Wan2.2-TI2V-5B weights, VAE, action-expert initialization, and cached text features. The \nickname arm replaces $30$ video self-attention blocks ($300$ tensors) with those from a 5B image-to-video variant at step $29{,}500$, warm-started from a video-to-video \nickname-5B model. This is distinct from the step-$29{,}000$ no-identity variant in \cref{tab:davis_backend}; its pretraining mixture includes OmniWorld-DROID robot videos. Track, camera, and domain modules are discarded because Fast-WAM lacks them. Both arms train the full model, including the transferred blocks, for $20{,}000$ steps on the four LIBERO~\cite{liu23libero:} suites with identical data order, normalization, seed, and optimization settings. The official release is a separate reference.
All checkpoints are evaluated with the Fast-WAM closed-loop evaluator: LIBERO uses $10$ tasks per suite and $50$ trials per task ($2{,}000$ rollouts per checkpoint); LIBERO-plus~\cite{fei25libero-plus:} uses all $10{,}030$ perturbed tasks with one trial each.
Per-suite results are in \cref{tab:fastwam_detail} and are discussed in \cref{sec:robot}. On LIBERO-plus, \nickname improves the goal suite ($+6.2$) but drops $4.8$ points on the spatial suite. All runs use a single seed.
\begin{table}[htbp]
\centering
\captionsetup{font=footnotesize}
\caption{\textbf{Fast-WAM details.} Closed-loop success rates (\%) per suite on LIBERO ($500$ trials per suite) and LIBERO-plus (one trial per task); \emph{All}: suite mean (LIBERO), task-weighted mean (LIBERO-plus). Official: released Fast-WAM checkpoint; Ctrl.: Wan2.2-initialized control; Ours: \nickname self-attention initialization.}%
\label{tab:fastwam_detail}
\footnotesize
\renewcommand{\arraystretch}{1.0}
\setlength{\tabcolsep}{4pt}
\begin{tabular*}{\linewidth}{@{\extracolsep{\fill}}lccccccc@{}}
\toprule
& \multicolumn{3}{c}{\textbf{LIBERO}} & \multicolumn{4}{c}{\textbf{LIBERO-plus}} \\
\cmidrule(lr){2-4}\cmidrule(l){5-8}
\textbf{Suite} & \textbf{Official} & \textbf{Ctrl.} & \textbf{Ours} & \textbf{Tasks} & \textbf{Official} & \textbf{Ctrl.} & \textbf{Ours} \\
\midrule
Spatial   & 97.40 & 97.00 & \textbf{97.60} & 2,402 & 54.58 & \textbf{59.87} & 55.08 \\
Object    & \textbf{99.40} & 99.20 & 99.20 & 2,518 & \textbf{69.46} & 62.43 & 65.45 \\
Goal      & \textbf{96.40} & 95.20 & 96.00 & 2,591 & 37.32 & 38.67 & \textbf{44.89} \\
LIBERO-10 & \textbf{94.00} & 91.00 & 93.00 & 2,519 & 42.83 & 43.03 & \textbf{43.35} \\
\midrule
All       & \textbf{96.80} & 95.60 & 96.45 & 10,030 & 50.91 & 50.81 & \textbf{52.10} \\
\bottomrule
\end{tabular*}

\end{table}

\Needspace{6\baselineskip}
\section{Broader Impacts}
\label{app:impacts}
Controllable novel-view video synthesis can support content creation, immersive media, simulation, and the construction of multi-view robot data. It can also fabricate plausible but incorrect views of real events. Source-video and geometric conditioning constrain generation but do not guarantee factual fidelity, especially in disoccluded regions. Track validity is an observation-support signal, not a forensic test. Deployment should distinguish generated views from recorded footage and respect the base model's license and safety requirements.

\section{Existing Assets and Licenses}%
\label{app:licenses}

We will release the code, pretrained models, the predictions of all evaluated methods, and the evaluation code at \url{https://github.com/NIRVANALAN/Holo4D}.
External assets retain their original licenses and terms.
\Cref{tab:asset_licenses} lists the licenses stated by their official repositories and release cards, distinguishing code, data, and model weights. Third-party footage and component assets retain their own terms.
Training datasets and baseline backbones are listed in \cref{tab:training_data,tab:baseline_settings}. Camera-adherence evaluation uses DAVIS~\cite{pont-tuset17davis:}; real novel-view evaluation uses iPhone/DyCheck~\cite{gao22monocular:}.
Pretrained components include the Wan~2.1 backbone~\cite{wan25wan:}, the umT5-XXL text encoder~\cite{chung23unimax:}, DepthAnything~3~\cite{lin26depth:}, and 4RC~\cite{luo264rc:}.
Reconstruction and preprocessing also use VGGT-$\Omega$~\cite{wang26vggt-omega:}, CoWTracker~\cite{lai26cowtracker:}, GMOS~\cite{xie26gmos:}, V-DPM~\cite{sucar26v-dpm:}, and Open-D4RT~\cite{zhang26efficiently:}. Evaluation uses Pi3~\cite{wang26pi:}, VBench~\cite{huang24vbench:}, held-out Syn4D scenes, and Vista4D's released trajectory benchmark~\cite{lin26vista4d:}. The 5B models use Wan2.2-TI2V-5B, the long-video continuation model uses Wan2.1-I2V-14B, and both additionally train on OmniWorld-DROID~\cite{zhou26omniworld:,khazatsky24droid:}.
Robot reshooting uses DROID~\cite{khazatsky24droid:} and publicly available online videos of Figure and Tesla Optimus humanoid robots, shown for qualitative illustration only; policy experiments use Fast-WAM~\cite{yuan26fast-wam:}, LIBERO~\cite{liu23libero:}, and LIBERO-plus~\cite{fei25libero-plus:}. SAM3~\cite{carion25sam3:} supplies precomputed MultiCamVideo foreground masks.
\begin{table}[p]
\centering
\fontsize{9}{10.5}\selectfont
\captionsetup{font=small}
\caption{\textbf{External asset licenses.} Asset names link to official license notices or release cards. Code licenses do not automatically cover separate weights or source footage.}
\label{tab:asset_licenses}
\setlength{\tabcolsep}{6pt}
\renewcommand{\arraystretch}{1.08}
\begin{tabular}{@{}>{\raggedright\arraybackslash}p{0.31\linewidth}>{\raggedright\arraybackslash}p{\dimexpr0.69\linewidth-12pt\relax}@{}}
\toprule
\textbf{Asset} & \textbf{License and scope} \\
\midrule
\multicolumn{2}{@{}l}{\textit{Training and evaluation data}} \\
\href{https://huggingface.co/datasets/Syn4D/Syn4D}{Syn4D} & CC BY 4.0; underlying assets retain their individual terms. \\
\href{https://huggingface.co/datasets/KwaiVGI/MultiCamVideo-Dataset}{MultiCamVideo}, \href{https://github.com/KAIR-BAIR/dycheck}{DyCheck iPhone} & Apache 2.0 (dataset releases). \\
\href{https://github.com/DL3DV-10K/Dataset/blob/main/License.md}{DL3DV-10K} & CC BY-NC 4.0 and DL3DV-10K Terms of Use. \\
\href{https://phuang17.github.io/DeepMVS/mvs-synth.html}{MVS-Synth} & Custom terms: research and educational use only. \\
\href{https://github.com/google-research/kubric/blob/main/LICENSE}{Kubric}, \href{https://github.com/google/realestate10k/blob/master/LICENSE}{RealEstate10K} & Apache 2.0 (code); rendered assets/source videos have separate terms. \\
\href{https://theairlab.org/tartanair-dataset/}{TartanAir} & CC BY 4.0 (official dataset terms). \\
\href{https://www.roboticsproceedings.org/rss20/p120.pdf}{DROID} & CC BY 4.0 (dataset). \\
\href{https://huggingface.co/datasets/InternRobotics/OmniWorld}{OmniWorld-DROID} & CC BY-NC-SA 4.0 (OmniWorld release); DROID attribution also applies. \\
\href{https://huggingface.co/datasets/Eyeline-Labs/Vista4D-Eval-Data}{Vista4D trajectory benchmark} & Apache 2.0 (release); source-video terms remain separate. \\
\href{https://davischallenge.org/}{DAVIS}; Figure/Optimus videos & Original source terms; no blanket footage license asserted here. \\
\href{https://github.com/Lifelong-Robot-Learning/LIBERO/blob/master/LICENSE}{LIBERO}; \href{https://huggingface.co/datasets/Sylvest/LIBERO-plus}{LIBERO-plus} & MIT (LIBERO code; LIBERO-plus asset card). \\
\midrule
\multicolumn{2}{@{}l}{\textit{Generator, reconstruction, and preprocessing}} \\
\href{https://github.com/Wan-Video/Wan2.1/blob/main/LICENSE.txt}{Wan2.1}, \href{https://huggingface.co/Wan-AI/Wan2.2-TI2V-5B}{Wan2.2-TI2V-5B}, \href{https://huggingface.co/google/umt5-xxl}{\mbox{umT5-XXL}} & Apache 2.0. \\
\href{https://github.com/Luo-Yihang/4RC/blob/main/LICENSE}{4RC} & S-Lab License 1.0 (noncommercial). \\
\href{https://github.com/facebookresearch/vggt-omega/blob/main/LICENSE}{VGGT-$\Omega$}, \href{https://github.com/facebookresearch/cowtracker/blob/main/LICENSE}{CoWTracker} & FAIR Noncommercial Research License. \\
\href{https://github.com/Jyxarthur/gmos/blob/main/LICENSE}{GMOS}, \href{https://github.com/facebookresearch/sam2/blob/main/LICENSE}{SAM2}, \href{https://huggingface.co/Lijiaxin0111/OpenD4RT}{Open-D4RT} & Apache 2.0; dependencies retain their own terms. \\
\href{https://github.com/yyfz/Pi3/blob/main/LICENSE}{Pi3 code}; \href{https://huggingface.co/yyfz233/Pi3}{Pi3 weights} & BSD 3-Clause (code); BSD 2-Clause (model card). \\
\href{https://github.com/eldar/vdpm/blob/main/LICENSE}{V-DPM} & MIT (code); separate checkpoint terms not verified. \\
\href{https://huggingface.co/depth-anything/DA3-GIANT-1.1}{Depth Anything 3 Giant 1.1} & CC BY-NC 4.0 (weights). \\
\href{https://github.com/facebookresearch/sam3/blob/main/LICENSE}{SAM3} & SAM License (custom code and model terms). \\
\midrule
\multicolumn{2}{@{}l}{\textit{Baselines and evaluation software}} \\
\href{https://github.com/KlingAIResearch/ReCamMaster/blob/main/LICENSE}{ReCamMaster}, \href{https://github.com/vidar-4d/ViDAR/blob/main/LICENSE}{ViDAR}, \href{https://github.com/yuantianyuan01/FastWAM/blob/main/LICENSE}{\mbox{Fast-WAM}} & MIT (code); separate model/component terms apply. \\
\href{https://github.com/TrajectoryCrafter/TrajectoryCrafter/blob/main/License}{TrajectoryCrafter} / \href{https://github.com/TrajectoryCrafter/TrajectoryCrafter/blob/main/License_Model}{CogVideoX} & Custom academic/research/education terms and CogVideoX License. \\
\href{https://github.com/tau-yihouxiang/EX-4D}{EX-4D}, \href{https://github.com/Eyeline-Labs/Vista4D/blob/main/LICENSE}{Vista4D} & Apache 2.0 (code); Wan/component terms apply. \\
\href{https://github.com/nv-tlabs/GEN3C}{GEN3C} & Apache 2.0 (code); NVIDIA Open Model License (weights). \\
\href{https://github.com/Vchitect/VBench/blob/master/LICENSE}{VBench}, \href{https://github.com/facebookresearch/dino}{DINO}, \href{https://github.com/OpenGVLab/InternVideo/blob/main/LICENSE}{\mbox{InternVideo/ViCLIP}} & Apache 2.0 (repositories); auxiliary model notices remain separate. \\
\href{https://github.com/openai/CLIP/blob/main/LICENSE}{CLIP}; \href{https://github.com/richzhang/PerceptualSimilarity/blob/master/LICENSE}{LPIPS} & MIT (CLIP); BSD 2-Clause (LPIPS). \\
\href{https://github.com/chaofengc/IQA-PyTorch/blob/main/LICENSE}{IQA-PyTorch / MUSIQ} & PolyForm Noncommercial 1.0.0 (current code); installed-version and weight terms require separate verification. \\
I3D and aesthetic-head weights & Exact cached-weight license notices not verified. \\
\bottomrule
\end{tabular}%

\end{table}

\clearpage
\section*{NeurIPS Paper Checklist}

\begin{enumerate}

\item {\bf Claims}
    \item[] Question: Do the main claims made in the abstract and introduction accurately reflect the paper's contributions and scope?
    \item[] Answer: \answerYes{}
    \item[] Justification: The abstract and the contribution list at the end of~\cref{sec:intro} claim a holistic scaffold of persistent 3D trajectories that carry source appearance, channel-wise conditioning of a pretrained video diffusion model on these trajectories, and evaluations against reshooting baselines, reconstructor substitutions, trajectory perturbations, and robot videos. These claims are supported by~\cref{tab:iphone_nvs,tab:your_result,tab:syn4d_gt_comparison,tab:stage1_design,tab:scaffold_analysis_main,tab:vista4d_native110} and~\cref{sec:robot}; the policy-learning result is reported as mixed (\cref{tab:fastwam_main}).

\item {\bf Limitations}
    \item[] Question: Does the paper discuss the limitations of the work performed by the authors?
    \item[] Answer: \answerYes{}
    \item[] Justification: \Cref{sec:conclusion} states that \nickname remains sensitive to geometric error, incomplete source coverage, and difficult camera trajectories. \Cref{tab:scaffold_analysis_main,tab:track_corruption} quantify the sensitivity to trajectory noise; \cref{app:qualitative} describes remaining failure modes (fine dynamic structures, large disoccluded regions) and missing coverage for content that enters after the query frame (\cref{fig:limitation}); \cref{sec:robot} states that the long-video continuation model is not zero-shot and that the policy-learning results are mixed and single-seed.

\item {\bf Theory assumptions and proofs}
    \item[] Question: For each theoretical result, does the paper provide the full set of assumptions and a complete (and correct) proof?
    \item[] Answer: \answerNA{}
    \item[] Justification: The paper presents an empirical method and does not introduce formal theorems or proofs.

    \item {\bf Experimental result reproducibility}
    \item[] Question: Does the paper fully disclose all the information needed to reproduce the main experimental results of the paper to the extent that it affects the main claims and/or conclusions of the paper (regardless of whether the code and data are provided or not)?
    \item[] Answer: \answerYes{}
    \item[] Justification: \Cref{sec:method} defines the trajectory scaffold, its projection and encoding, and the channel-wise conditioning of the generator; \cref{sec:training_details} gives the backbone, training schedule, learning rates, and batch sizes; \cref{app:impl,app:training_data} give module dimensions, augmentation, long-video inference, dataset weights, and annotation sources; \cref{app:davis_eval,app:vista4d_native,app:iphone_eval,app:robustness} give the evaluation sets, trajectories, and camera-recovery protocols.

\item {\bf Open access to data and code}
    \item[] Question: Does the paper provide open access to the data and code, with sufficient instructions to faithfully reproduce the main experimental results, as described in supplemental material?
    \item[] Answer: \answerYes{}
    \item[] Justification: We will publicly release the code, pretrained models, the predictions of all evaluated methods, and the evaluation code, with instructions to reproduce the main results, at \url{https://github.com/NIRVANALAN/Holo4D} (\cref{app:licenses}). All training datasets (\cref{tab:training_data}) and evaluation benchmarks are publicly available, and the data sources and evaluation protocols are described in \cref{app:training_data,app:davis_eval,app:vista4d_native,app:iphone_eval,app:robustness}.

\item {\bf Experimental setting/details}
    \item[] Question: Does the paper specify all the training and test details (e.g., data splits, hyperparameters, how they were chosen, type of optimizer) necessary to understand the results?
    \item[] Answer: \answerYes{}
    \item[] Justification: \Cref{sec:experiments} describes baselines, evaluation sets, reconstruction backends, and inference settings; \cref{sec:training_details,app:impl,app:training_data} give the optimizer (AdamW), learning rates, batch sizes, schedules, resolutions, data mixture, and augmentation; per-benchmark protocols are in~\cref{app:davis_eval,app:vista4d_native,app:iphone_eval,app:robustness}.

\item {\bf Experiment statistical significance}
    \item[] Question: Does the paper report error bars suitably and correctly defined or other appropriate information about the statistical significance of the experiments?
    \item[] Answer: \answerNo{}
    \item[] Justification: The main video-generation comparisons report point estimates, without repeated-training confidence intervals. Evaluation uses fixed benchmark pairs; valid camera-metric counts can differ by method (\cref{app:vista4d_native}). The early 1.3B conditioning study reports a confidence interval over pairs, runtime profiling reports variation across videos (\cref{tab:runtime_breakdown}), and policy learning uses a single seed (\cref{app:fastwam}). These do not quantify uncertainty for the main results.

\item {\bf Experiments compute resources}
    \item[] Question: For each experiment, does the paper provide sufficient information on the computer resources (type of compute workers, memory, time of execution) needed to reproduce the experiments?
    \item[] Answer: \answerYes{}
    \item[] Justification: \Cref{app:impl} reports the training compute: both stages run on NVIDIA GH200 GPUs for about $24$ hours each, with stage~2 on $32$ GPUs using FSDP. \Cref{app:efficiency} reports inference hardware, time, and memory per stage and per reconstruction backend (\cref{tab:runtime_breakdown,tab:backend_cost,tab:vista4d_runtime}).

\Needspace{6\baselineskip}
\item {\bf Code of ethics}
    \item[] Question: Does the research conducted in the paper conform, in every respect, with the NeurIPS Code of Ethics \url{https://neurips.cc/public/EthicsGuidelines}?
    \item[] Answer: \answerYes{}
    \item[] Justification: The work uses publicly released datasets and models and publicly available robot videos under their original terms (\cref{app:licenses}), involves no new human-subject data collection, and discusses misuse risks (\cref{app:impacts}).

\item {\bf Broader impacts}
    \item[] Question: Does the paper discuss both potential positive societal impacts and negative societal impacts of the work performed?
    \item[] Answer: \answerYes{}
    \item[] Justification: \Cref{app:impacts} discusses positive uses (content creation, immersive media, simulation, and robot data) and the risk that generated views are mistaken for recorded footage, together with mitigations.

\item {\bf Safeguards}
    \item[] Question: Does the paper describe safeguards that have been put in place for responsible release of data or models that have a high risk for misuse (e.g., pre-trained language models, image generators, or scraped datasets)?
    \item[] Answer: \answerNA{}
    \item[] Justification: \nickname fine-tunes a publicly released video diffusion model (Wan2.1) and conditions on a given source video; it does not introduce a new high-risk pretrained model or a scraped dataset. Released checkpoints will follow the base model's license and usage terms, and \cref{app:impacts} discusses the risk that generated views are mistaken for recorded footage.

\item {\bf Licenses for existing assets}
    \item[] Question: Are the creators or original owners of assets (e.g., code, data, models), used in the paper, properly credited and are the license and terms of use explicitly mentioned and properly respected?
    \item[] Answer: \answerYes{}
    \item[] Justification: Every dataset, pretrained model, reconstruction method, and benchmark is cited at first use; \cref{app:licenses,app:training_data} list them and state that they are used under their original licenses and terms. The humanoid-robot clips are publicly available online videos shown for qualitative illustration only.

\item {\bf New assets}
    \item[] Question: Are new assets introduced in the paper well documented and is the documentation provided alongside the assets?
    \item[] Answer: \answerYes{}
    \item[] Justification: The code, checkpoints, predictions, and evaluation code will be publicly released at \url{https://github.com/NIRVANALAN/Holo4D} (\cref{app:licenses}) with documentation of the environment, training and inference commands, and data layout, under the base model's license.

\item {\bf Crowdsourcing and research with human subjects}
    \item[] Question: For crowdsourcing experiments and research with human subjects, does the paper include the full text of instructions given to participants and screenshots, if applicable, as well as details about compensation (if any)?
    \item[] Answer: \answerNA{}
    \item[] Justification: The work involves no crowdsourcing or newly recruited human participants. It uses existing datasets and publicly available robot videos (\cref{app:training_data,app:licenses}).

\item {\bf Institutional review board (IRB) approvals or equivalent for research with human subjects}
    \item[] Question: Does the paper describe potential risks incurred by study participants, whether such risks were disclosed to the subjects, and whether Institutional Review Board (IRB) approvals (or an equivalent approval/review based on the requirements of your country or institution) were obtained?
    \item[] Answer: \answerNA{}
    \item[] Justification: The paper does not involve research with human subjects, so IRB approval is not applicable.

\item {\bf Declaration of LLM usage}
    \item[] Question: Does the paper describe the usage of LLMs if it is an important, original, or non-standard component of the core methods in this research? Note that if the LLM is used only for writing, editing, or formatting purposes and does \emph{not} impact the core methodology, scientific rigor, or originality of the research, declaration is not required.
    \item[] Answer: \answerNA{}
    \item[] Justification: LLMs are not part of the core methodology of \nickname. The text encoder used by the diffusion backbone (umT5-XXL) is a standard component inherited from the Wan~2.1 base model and serves only as a frozen text-conditioning module rather than as an original methodological contribution.

\end{enumerate}

\end{document}